\documentclass{worldbench}
\definecolor{w_blue}{RGB}{52,204,204}
\definecolor{w_yellow}{RGB}{255,192,0}
\definecolor{w_red}{RGB}{192,0,0}

\usepackage{algorithmic}
\usepackage{algorithm}
\usepackage{amsmath}
\usepackage{amssymb}
\usepackage{amsthm}

\usepackage{booktabs}
\usepackage{adjustbox}

\usepackage{caption}
\usepackage{color}
\usepackage{colortbl}

\usepackage{enumitem}

\usepackage{fontawesome}

\usepackage{graphicx}
\usepackage{booktabs}
\usepackage{array}
\usepackage{xltabular}
\usepackage{hhline}
\usepackage{longtable}

\usepackage{makecell}
\usepackage{mathtools}
\usepackage{microtype}
\usepackage{multicol}
\usepackage{multirow}

\usepackage{pifont}

\usepackage{subcaption}

\usepackage{tabularx}
\usepackage{tikz}

\usepackage{wrapfig}

\usepackage{xspace}
\newcommand*{\ie}{\emph{i.e.}\@\xspace}

\usepackage{titletoc}

\definecolor{crKITTI}{RGB}{0, 102, 102}
\definecolor{crNYU}{RGB}{83, 22, 136}
\definecolor{crCarla}{RGB}{153, 153, 0}
\definecolor{crSemanticKITTI}{RGB}{231, 76, 60}
\definecolor{crnuScenes}{RGB}{89, 74, 235}
\definecolor{crWaymo}{RGB}{106, 228, 163}
\definecolor{crSeeingThroughFog}{RGB}{96, 96, 96}
\definecolor{vKITTI}{RGB}{202, 200, 229}
\definecolor{crArgoverse}{RGB}{255, 128, 0}
\definecolor{crLyft-Level5}{RGB}{235, 11, 140}
\definecolor{crnuPlan}{RGB}{52, 204, 204}
\definecolor{crPandaSet}{RGB}{32, 32, 32}
\definecolor{crOpenCOOD}{RGB}{0, 0, 255}
\definecolor{crKITTI360}{RGB}{165, 105, 189}
\definecolor{crCarlaSC}{RGB}{0, 176, 240}
\definecolor{crRobo}{RGB}{69, 82, 104}
\definecolor{crOpenOcc}{RGB}{255, 0, 255}
\definecolor{crOcc3D-nuScenes}{RGB}{89, 74, 235}
\definecolor{crOpenDV}{RGB}{255, 51, 51}
\definecolor{crSSCBench}{RGB}{83, 22, 136}
\definecolor{crNAVSIM}{RGB}{153, 0, 76}
\definecolor{crDrivingDojo}{RGB}{152, 77, 16}
\definecolor{EUVS}{RGB}{187, 155, 209}
\definecolor{crOmniDrive}{RGB}{155, 200, 26}
\definecolor{crPi3DET}{RGB}{239, 99, 75}
\definecolor{crPrivate}{RGB}{90, 90, 90}
\definecolor{crNuplanOcc}{RGB}{214, 152, 60}
\definecolor{crLidarOpen}{RGB}{80, 160, 175}
\definecolor{crOccInteract}{RGB}{175, 120, 195}

\definecolor{crWorldLens}{RGB}{60, 130, 200}

\newcolumntype{C}[1]{>{\centering\arraybackslash}m{#1}}
\newcolumntype{L}[1]{>{\raggedright\arraybackslash}p{#1}}
\newcolumntype{Y}{>{\raggedright\arraybackslash}X}
\newcolumntype{Z}{>{\centering\arraybackslash}X}
\definecolor{VLAbg}{HTML}{FFFBE8}
\definecolor{WAMbg}{HTML}{EEF6FF}
\newcommand{\DataIcon}[1]{\raisebox{-0.18ex}{\includegraphics[height=1.12em]{#1}}}
\newcommand{\DataUMI}{\DataIcon{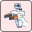}}
\newcommand{\DataReal}{\DataIcon{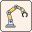}}
\newcommand{\DataEgo}{\DataIcon{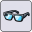}}
\newcommand{\DataSim}{\DataIcon{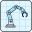}}
\newcommand{\DataGen}{\DataIcon{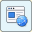}}
\newcommand{\DataIcons}[1]{\mbox{#1}}

\newtcolorbox{wbcard}[1]{%
  enhanced, breakable,
  colback=wblue!4, colframe=wblue,
  boxrule=0.7pt, arc=4pt,
  left=9pt, right=9pt, top=6pt, bottom=6pt,
  before skip=8pt, after skip=8pt,
  fonttitle=\sffamily\bfseries\large, coltitle=white,
  colbacktitle=wblue, title={#1}
}
\newtcolorbox{wbtakeaway}{%
  enhanced, breakable,
  colback=wblue!5, colframe=wblue,
  boxrule=0.5pt, arc=3pt,
  left=9pt, right=9pt, top=5pt, bottom=5pt,
  before skip=7pt, after skip=7pt,
  borderline west={2.5pt}{0pt}{wblue},
  fonttitle=\sffamily\bfseries, coltitle=wfg,
  title={Takeaway}
}
\definecolor{layerReal}{HTML}{C0392B}
\definecolor{layerEgo}{HTML}{8E44AD}
\definecolor{layerUMI}{HTML}{1F9E9E}
\definecolor{layerSim}{HTML}{2E7D32}
\definecolor{layerGen}{HTML}{34495E}
\definecolor{layerApp}{HTML}{F5B027}
\newtcbox{\layerbadge}[1][gray]{%
  on line, arc=2pt, colback=#1, colframe=#1, boxrule=0pt,
  left=3pt, right=3pt, top=1pt, bottom=1pt,
  fontupper=\scriptsize\sffamily\bfseries\color{white}
}
\newcommand{\bReal}{\layerbadge[layerReal]{Real-Robot}}
\newcommand{\bEgo}{\layerbadge[layerEgo]{Egocentric}}
\newcommand{\bUMI}{\layerbadge[layerUMI]{UMI}}
\newcommand{\bSim}{\layerbadge[layerSim]{Simulation}}
\newcommand{\bGen}{\layerbadge[layerGen]{General}}
\newcommand{\bApp}{\layerbadge[layerApp]{Data application}}
\DeclareRobustCommand{\DataLegendItem}[2]{%
  \begingroup
    \edef\SavedCaptionFont{\the\font}%
    \raisebox{-0.18\height}{%
      \resizebox{!}{1.15em}{{#1}}%
    }%
    \SavedCaptionFont\nobreakspace #2%
  \endgroup
}

\titlecontents{section}
  [0.45em]
  {\normalsize\sffamily\bfseries\vspace{4pt}}
  {\textcolor{wblue}{\contentslabel{1.65em}}}
  {}
  {\titlerule*[0.45pc]{.}\contentspage}
\titlecontents{subsection}
  [2.1em]
  {\normalsize\sffamily\vspace{1pt}}
  {\textcolor{wblue}{\contentslabel{2.35em}}}
  {}
  {\titlerule*[0.45pc]{.}\contentspage}

\project{\href{https://jasper-aaa.github.io/embodied-data-pyramid/}{\texttt{Embodied Data Pyramid}}}
\github{\href{https://github.com/worldbench/awesome-embodied-data-pyramid}{\texttt{Awesome Embodied Data Pyramid}}}
\data{\href{https://www.modelscope.cn/datasets/jasper0519/DataPyramid}{\texttt{ModelScope: DataPyramid}}}

\title{Data Pyramid for Embodied Manipulation: A Survey}
\data{\href{https://www.modelscope.cn/datasets/jasper0519/DataPyramid}{\texttt{10K+ Hours of Formatted Robot Data}}}
\project{\href{https://jasper-aaa.github.io/embodied-data-pyramid/}{\texttt{Embodied Data Pyramid}}}
\github{\href{https://github.com/worldbench/awesome-embodied-data-pyramid}{\texttt{Awesome Embodied Data Pyramid}}}
\date{\today}

\affiliationlogos{%
  \includegraphics[height=0.65cm]{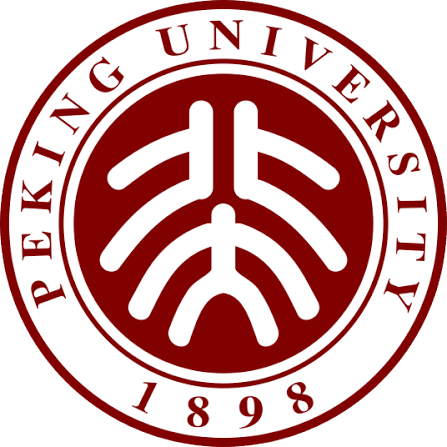}\hspace{0.18cm}%
  \includegraphics[height=0.65cm]{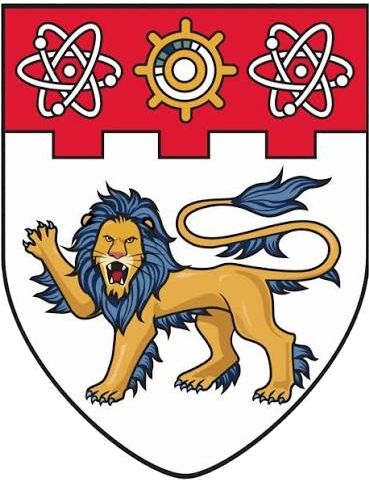}\hspace{0.18cm}%
  \includegraphics[height=0.65cm]{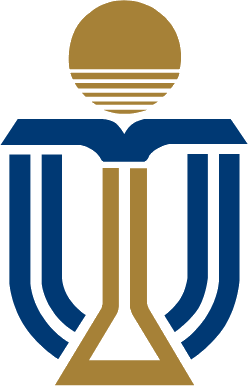}\hspace{0.18cm}%
  \includegraphics[height=0.65cm]{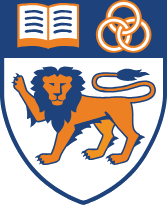}\hspace{0.18cm}%
  \includegraphics[height=0.65cm]{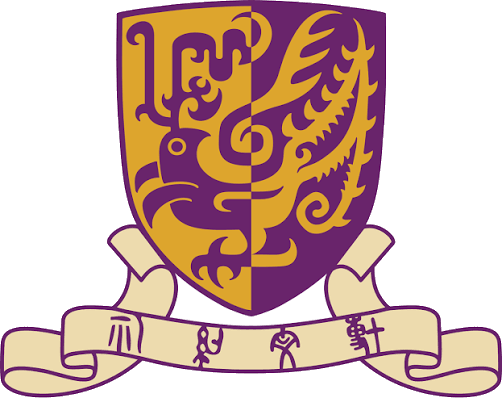}\hspace{0.18cm}%
  \includegraphics[height=0.65cm]{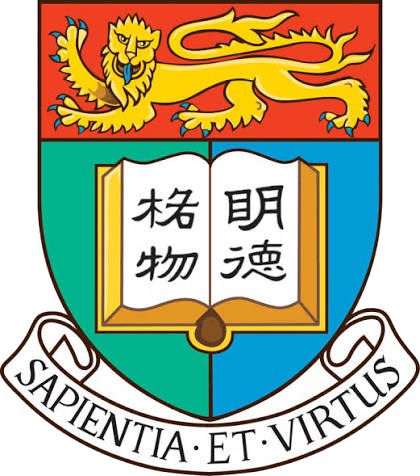}\hspace{0.18cm}%
  \includegraphics[height=0.65cm]{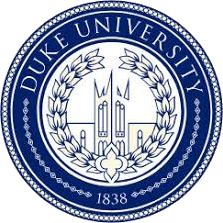}\hspace{0.18cm}%
  \includegraphics[height=0.65cm]{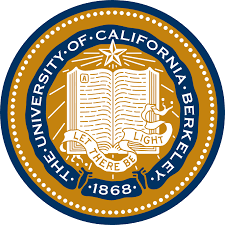}\hspace{0.18cm}%
  \includegraphics[height=0.65cm]{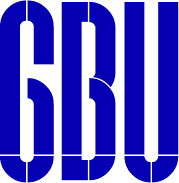}\hspace{0.18cm}%
  \includegraphics[height=0.65cm]{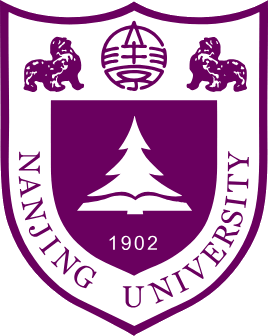}\hspace{0.18cm}%
  \includegraphics[height=0.65cm]{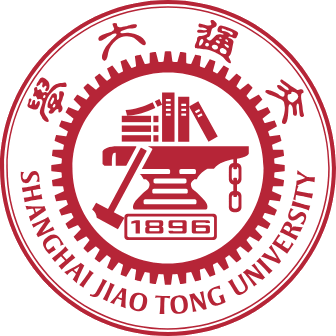}%
}

\author[1]{Yifan~Ye$^{*}$}
\author[1]{Yankai~Fu$^{*}$}
\author[1]{Yaoxu~Lv$^{*}$}
\author[2]{Bohan~Hou$^{*}$}
\author[3]{Jun~Cen$^{*, \dagger}$}
\author[4]{Lingdong~Kong$^{*}$}
\author[5]{Duo~Zheng}
\author[6]{Tianxing~Chen}
\author[1]{Jiaming~Liu}
\author[2]{Ziang~Cao}
\author[1]{Yunfan~Lou}
\author[4]{Wei~Chow}
\author[7]{Xian~Sun}
\author[8]{Yingshuo~Wang}
\author[3]{Kuangzhi~Ge}
\author[3]{Xiaowei~Chi}
\author[9]{Xidong~Zhang}
\author[1]{Zhibo~Pang}
\author[1]{Yiwu~Zhong}
\author[3]{Sirui~Han}
\author[10]{Zhihe~Lu}
\author[10]{Weihao~Yuan}
\author[3]{Qifeng~Chen}
\author[9]{Michael~Yu~Wang}
\author[11]{Yao~Mu}
\author[2]{Ziwei~Liu}
\author[2]{Jianfei~Yang}
\author[6]{Ping~Luo}
\author[1]{Shanghang~Zhang$^\ddagger$}

\affiliation[1]{PKU}
\affiliation[2]{NTU}
\affiliation[3]{HKUST}
\affiliation[4]{NUS}
\affiliation[5]{CUHK}
\affiliation[6]{HKU}
\affiliation[7]{Duke}
\affiliation[8]{UCB}
\affiliation[9]{GBU}
\affiliation[10]{NJU}
\affiliation[11]{SJTU}

\contribution[*]{Equal Contributions}
\contribution[\dagger]{Project Leader}
\contribution[\ddagger]{Corresponding Author}

\abstract{
Multimodal foundation models learned to see and to speak by consuming the whole internet. Embodied agents admit no such shortcut, since they require data that couple observations with physical states and actions. These signals can be provided, to varying degrees, by multiple data sources. In this work, we organize the embodied data ecosystem as a ``pyramid'' spanning five complementary sources: $^1$real-robot data, $^2$UMI-style data, $^3$egocentric and exocentric data, $^4$simulation data, and $^5$general vision-language data. We organize the pyramid around the tension between scalability and robot alignment, and further characterize each source in terms of data quality, diversity, reusability, and physical fidelity. We then analyze recent embodied foundation models through the lens of data recipes, examining how different sources are selected, aligned, and mixed during pretraining. For embodied brain models, vision-language-action models, and world-action models alike, we relate data composition to capabilities in perception, reasoning, planning, action generation, and world prediction. To facilitate community use, we additionally consolidate over 10,000 hours of trajectories from widely used robot datasets into a unified LeRobot format, with transparent frame-level quality annotations for filtering. We close by discussing six open challenges: building large-scale tactile datasets, collecting failure and recovery data, developing scalable data-collection pipelines, aligning actions across embodiments, leveraging egocentric data for dexterous manipulation, and designing principled data recipes for robot learning. We hope this work lays the foundation for the design of next-generation embodied systems.

}

\begin{document}

\maketitle
\thispagestyle{empty}

\begin{center}
    \centering
    \includegraphics[width=\linewidth]{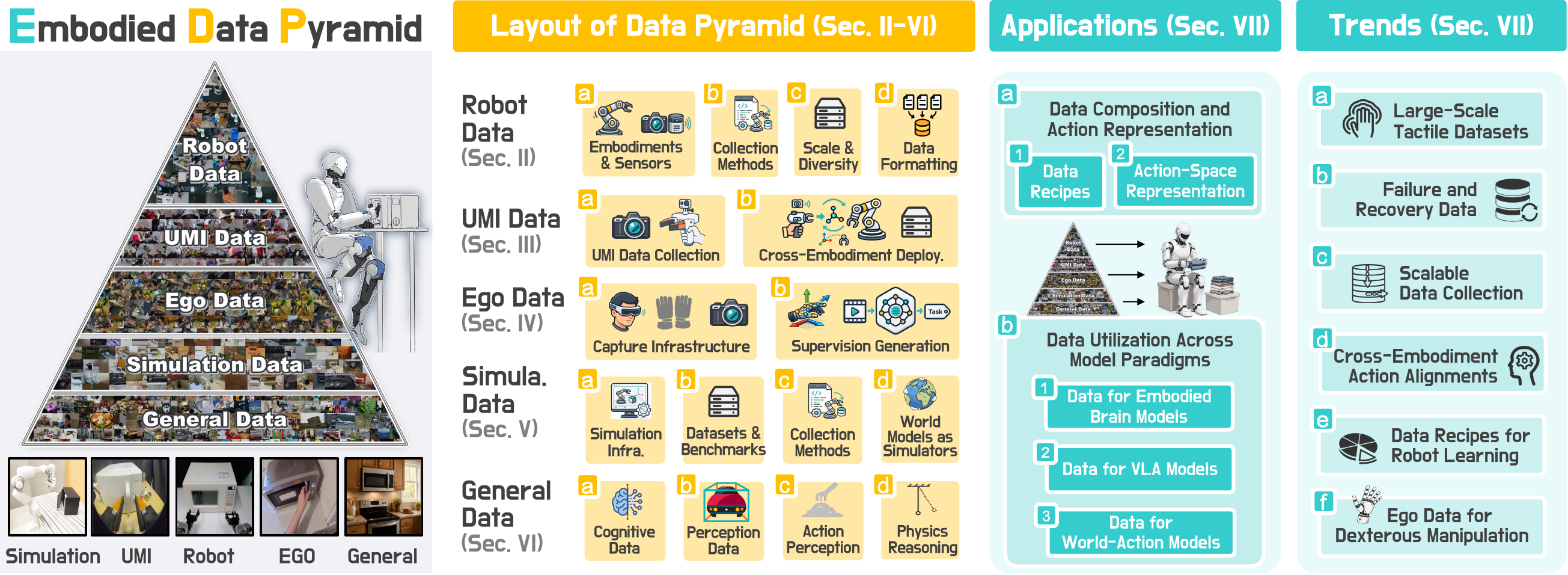}
    \vspace{-0.55cm}
    \captionof{figure}{\textbf{Overview of Organization and Scope.} This work is organized into three parts. We first introduce the embodied data pyramid, which categorizes data sources into real-robot, UMI-style, egocentric-exocentric, simulation, and general data according to their scalability and robot alignment, and reviews their datasets, collection pipelines, embodiments, sensing configurations, supervision, scale, diversity, and transferability. We then examine how these heterogeneous sources support embodied foundation models. Finally, we discuss future directions in embodied data.}
    \label{fig:pyramid}
    \vspace{-1.5cm}
\end{center}

\clearpage\clearpage
\thispagestyle{empty}
{\hypersetup{linkcolor=wfg}\vspace*{-1.7cm}\tableofcontents\hypersetup{linkcolor=wfg}\vspace*{-1cm}}

\clearpage\clearpage
\section{Introduction}

Multimodal foundation models have acquired broad perceptual, linguistic, and reasoning capabilities largely by scaling pretraining over abundant visual and linguistic data~\cite{brohan2022rt,driess2023palmeembodiedmultimodallanguage}. Extending this paradigm to embodied foundation models requires more than the ability to process visual and linguistic inputs. Embodied agents must additionally understand physical states and dynamics, reason about how actions transform the environment, and execute appropriate behaviors in the physical world~\cite{mu2023embodiedgpt,brohan2023rt2visionlanguageactionmodelstransfer,kim2024openvla,bjorck2025gr00t}. These requirements fundamentally change the nature of supervision needed for pretraining and raise a central question:

\begin{wbcard}{}
\centering
\emph{What data should be used to train embodied foundation models with such capabilities?}
\end{wbcard}

The community has explored several complementary sources of supervision for embodied foundation models. These include observation-action trajectories collected from physical robots and simulation, egocentric and exocentric recordings of human interaction, general image, video, language, and vision-language data, and, more recently, UMI-style demonstrations that capture object- and end-effector-centric manipulation without requiring a robot during collection~\cite{brohan2022rt,bu2025agibot,robogen2024,grauman2024egoexo4d,brohan2023rt2visionlanguageactionmodelstransfer,chi2024universal}. Each source contributes different forms of semantic, temporal, physical, and action-related supervision. However, their respective roles, trade-offs, relationships, and integration strategies remain insufficiently systematized. 

Several recent embodied foundation models, including Motus~\cite{bi2025motusunifiedlatentaction} and GR00T~\cite{bjorck2025gr00t}, have introduced hierarchical or pyramid-like views of training data to motivate the use of heterogeneous sources. These formulations are typically designed around the training recipe of a particular model and provide limited analysis of how data categories differ in collection scalability, embodiment dependence, physical fidelity, transferability, and downstream utility. Existing surveys likewise tend to focus on model architectures or specific subproblems, including Vision-Language-Action models~\cite{vlasurvey}, World-Action models~\cite{hou2026world}, and world modeling~\cite{kong2025worldmodeling}. Consequently, the embodied data ecosystem has not yet been systematically organized at the category level, nor has the use of heterogeneous data by embodied foundation models been comprehensively analyzed from a data-centric perspective.

To answer this question systematically, we therefore frame the embodied data problem around two questions:
\begin{itemize}
\item \emph{``How can heterogeneous embodied data sources be collected, organized, compared, and integrated despite their differences in scalability, robot alignment, physical fidelity, and transferability?''}

\item \emph{``How can these heterogeneous data sources be effectively selected, combined, and utilized by embodied foundation models?''}

\end{itemize}

To examine the first question, we organize the existing embodied data ecosystem into a \textbf{data pyramid for robotics and embodied intelligence}. The pyramid is guided by two complementary design principles. The first is \emph{Scalability}, which concerns how efficiently a data source can be expanded with respect to hardware dependence, human labor, environment reset, safety supervision, and marginal generation cost. The second is \emph{Robot Alignment}, which concerns how directly its observations, representations, and supervision signals can support learning and execution on physical robots. 
\begin{figure}[t]
    \centering
    \includegraphics[width=\textwidth]{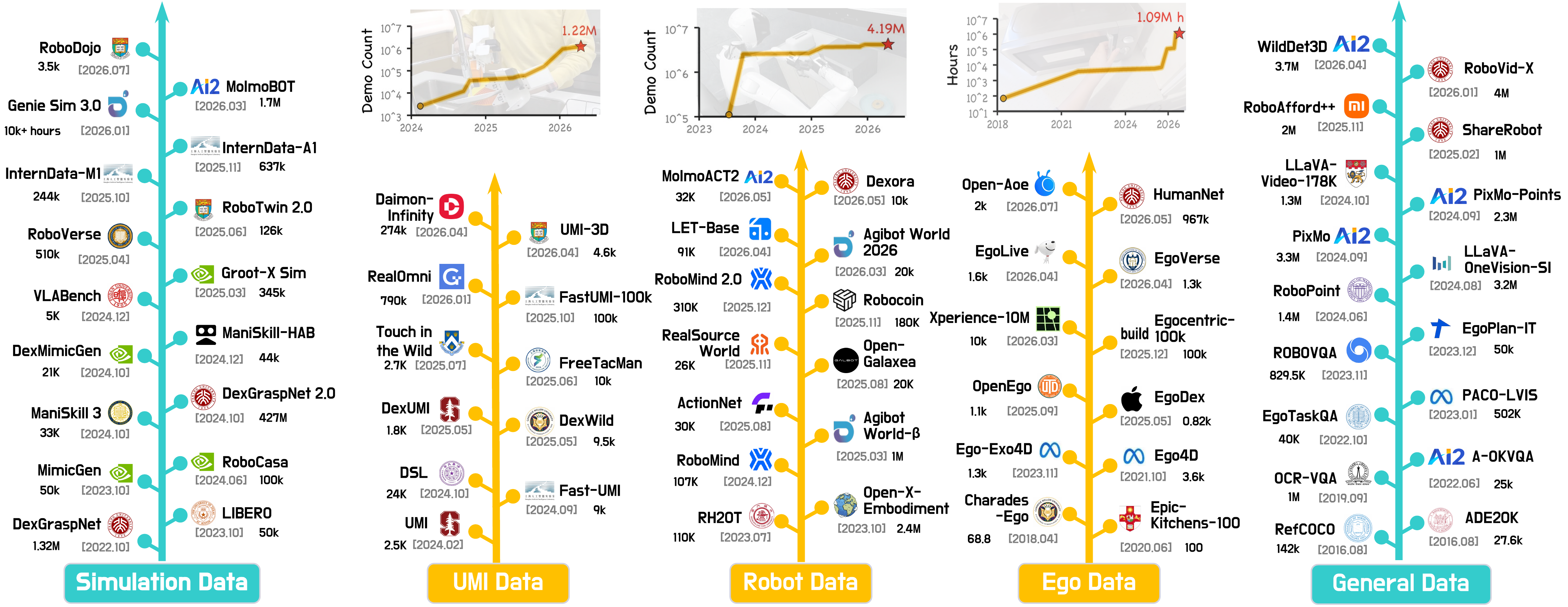}
    \vspace{-0.6cm}
    \caption{\textbf{Evolution of Data Scale.} Evolution of data scale across the embodied data pyramid. From left to right, we select a subset of representative datasets from simulation data, UMI data, robot data, ego data, and general data to visualize their chronological scale growth. Simulation, UMI, and robot datasets are measured by the number of demonstrations, ego datasets by the number of collected hours, and general datasets by the number of question-answer pairs. The inset curves summarize the scale evolution of selected datasets within the corresponding data categories.}
    \label{fig:timeline}
\end{figure}


These two principles are often in tension: data that is closely aligned with real-robot execution is typically expensive and difficult to scale, whereas highly scalable data may provide only indirect or imperfect supervision for physical interaction. Simulation, for example, can continuously generate robot-oriented trajectories once the required assets, tasks, and generation pipelines have been constructed, but its utility remains constrained by discrepancies between simulated and real physical systems~\cite{robogen2024,gensim22024,maniskill32025,robocasa2024,da2025surveysimtorealmethodsrl,simplerenv2024}.

\begin{figure}[t]
    \centering
    \includegraphics[width=\textwidth]{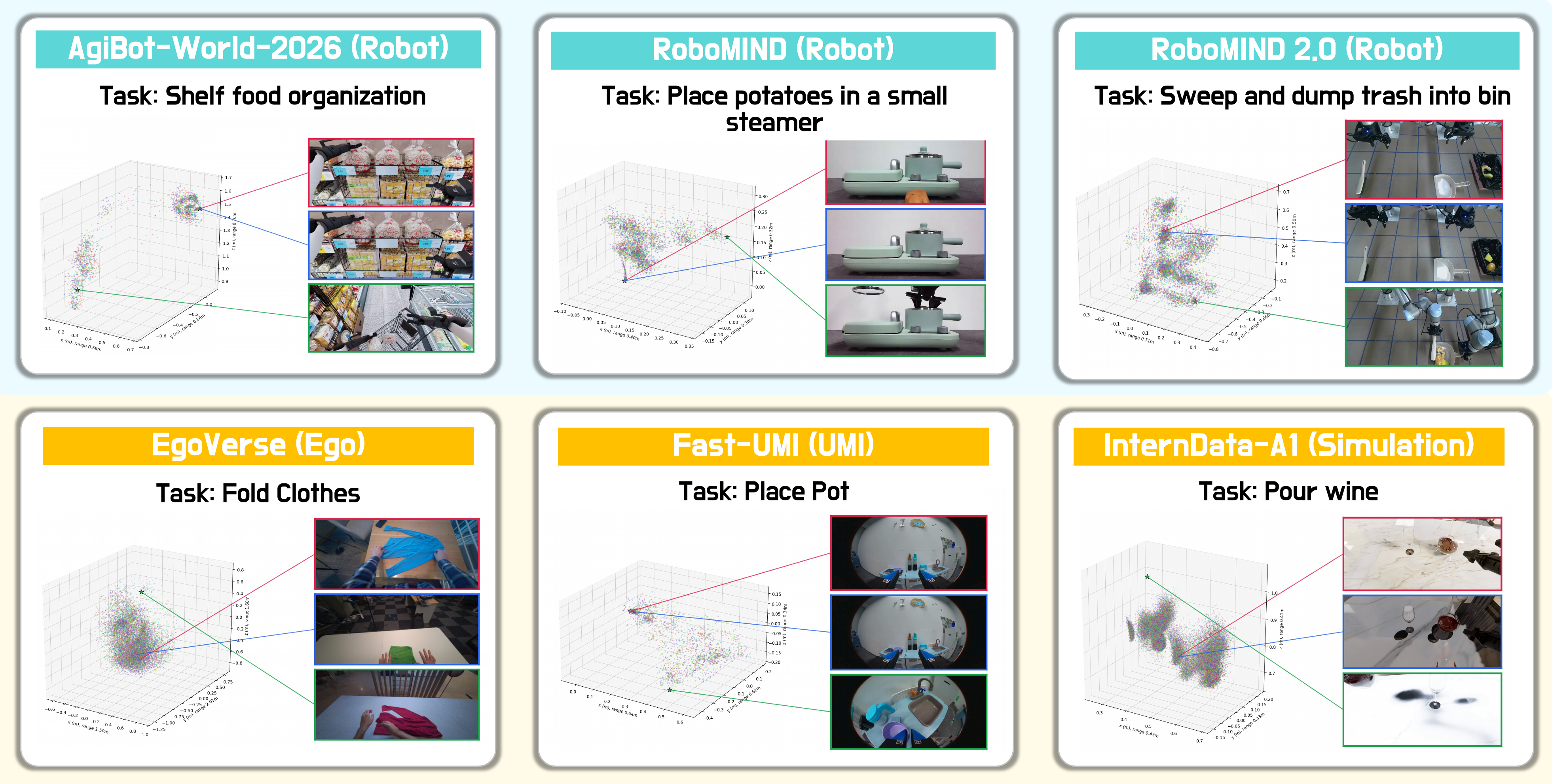}
    \vspace{-0.55cm}
    \caption{\textbf{Visualization of action-trajectory diversity across representative datasets.} Key action frames are extracted using a heuristic procedure following PerAct~\cite{shridhar2022peract}. Under comparable initial configurations, the spatial distribution of grasping and interaction keyframes provides a proxy for the diversity of demonstrated trajectories. Human-hand trajectories are processed with an analogous heuristic to extract corresponding hand keyframes.}
    \label{fig:diversity}
\end{figure}

Beyond these two primary organizing principles, we introduce four complementary category-level dimensions to characterize the broader utility of different data sources. \emph{Quality} concerns the validity, consistency, informativeness, and task relevance of the collected experience; simply increasing data volume may provide limited benefit when trajectories are repetitive, annotations are inaccurate, signals are poorly synchronized, or collection conditions are systematically biased~\cite{lin2024data}. \emph{Diversity} reflects coverage across tasks, objects, scenes, viewpoints, instructions, embodiments, sensing configurations, behaviors, and outcomes, and is important for robustness under distribution shift~\cite{o2024open,roboverse2025}. As shown in Figure~\ref{fig:diversity}, representative datasets exhibit substantially different distributions of interaction trajectories, observation viewpoints, task stages, and operating environments.  \emph{Reusability} characterizes how readily data can be transferred across tasks, environments, embodiments, sensing systems, and model families. For example, object- or end-effector-centric demonstrations can be retargeted to more dexterous embodiments~\cite{chi2024universal,seo2024legato}, while egocentric videos provide reusable supervision for hand-object interaction, affordance understanding, and task decomposition~\cite{grauman2022ego4d,grauman2024egoexo4d,kareer2025egomimic,yang2025egovla,srirama2024hrp}. Finally, \emph{physical fidelity} describes how faithfully a data category captures real interaction dynamics and feedback, including contact, friction, compliance, sensing noise, actuation delay, and object motion. These four dimensions complement scalability and robot alignment by explaining whether a data source is not only large and robot-relevant, but also informative, diverse, transferable, and physically meaningful.

Taken together, these six dimensions motivate our category-level organization of the embodied data ecosystem. From the apex to the base, we order the five categories as \emph{real-robot data}, \emph{UMI-style data} (Universal Manipulation Interface), \emph{egocentric and exocentric data}, \emph{simulation data}, and \emph{general data}. This ordering reflects an overall transition from stronger physical grounding and more directly actionable robot supervision toward greater accessibility and scalability.

Descending the pyramid is thus a systematic trade. \emph{Real-robot data} sits at the apex because its recorded actions are directly executable on the platform that produced them, a fidelity paid for in hardware, human effort, and resets~\cite{brohan2022rt,walke2023bridgedata,o2024open,bu2025agibot}. Each layer below relaxes some part of that coupling in exchange for reach: UMI-style rigs drop the robot from the collection loop while retaining end-effector supervision~\cite{chi2024universal,wu2024fast,liu2025vitamin,seo2024legato,liu2025fastumi100k}; human recordings drop the gripper but keep real physics and everyday diversity~\cite{grauman2022ego4d,grauman2024egoexo4d,huang2024egoexolearn,kareer2025egomimic,yang2025egovla,luo2025beingh0}; simulation restores executable actions and privileged labels but approximates the physics itself~\cite{robogen2024,gensim22024,maniskill32025,robocasa2024,da2025surveysimtorealmethodsrl,simplerenv2024}; and general data abandons robot grounding altogether for web-scale semantic and reasoning coverage~\cite{brohan2023rt2visionlanguageactionmodelstransfer,rynnbrain2026}. The cards below summarize each layer, and the ordering represents an overall synthesis of the six dimensions rather than a strictly monotonic progression along every individual property.

\begin{wbcard}{The Data Pyramid at a Glance}
Two primary axes organize the pyramid. Scalability captures how efficiently a data source can be expanded, while robot alignment captures how directly it supports learning and execution on physical robots. These axes are often in tension. Quality, diversity, reusability, and physical fidelity further characterize the utility and limitations of each layer.
\end{wbcard}

\begin{wbcard}{}
    \textbf{Type 1: Real-Robot Data}:\hfill\bReal\\[1ex]
    Trajectories teleoperated or scripted on the target robot, recording observations, states, actions, and outcomes in one closed loop. Nothing is more directly executable, or more expensive: every hour needs hardware, an operator, and a reset.
\end{wbcard}

\begin{wbcard}{}
    \textbf{Type 2: UMI-Style Data}:\hfill\bUMI\\[1ex]
    Demonstrations captured with handheld grippers that record end-effector motion without a robot in the loop, freeing collection to move into everyday environments. The gripper supervision survives that move; joint-level proprioception does not, and retargeting remains.
\end{wbcard}

\begin{wbcard}{}
    \textbf{Type 3: Egocentric / Exocentric Data}:\hfill\bEgo\\[1ex]
    First- and third-person recordings of people acting, offering real physics, dexterous hands, and everyday variety at the scale wearable capture allows. No robot is involved, so actions must be reconstructed and retargeted across the human-robot gap.
\end{wbcard}

\begin{wbcard}{}
    \textbf{Type 4: Simulation Data}:\hfill\bSim\\[1ex]
    Interaction generated in physics engines, returning executable actions alongside privileged labels such as contacts, poses, and success signals, in parallel and at negligible marginal cost. What it cannot return is the physics it approximates.
\end{wbcard}

\begin{wbcard}{}
    \textbf{Type 5: General Data}:\hfill\bGen\\[1ex]
    Web-scale images, video, language, and vision-language corpora carrying semantics, spatial structure, and commonsense far beyond any robot dataset. They ground perception and reasoning rather than action, and say nothing of contact or consequence.
\end{wbcard}
\vspace{0.2cm}
The data pyramid provides a static, category-level organization of the embodied data ecosystem. Beyond this taxonomy, we further examine how the scale of each data category has evolved over time. As shown in Figure~\ref{fig:timeline}, real-robot, UMI-style, egocentric and exocentric, simulation, and general datasets have all expanded over time, with each category exhibiting a strong growth trend. 

\begin{figure}[t]
    \centering
    \includegraphics[width=\textwidth]{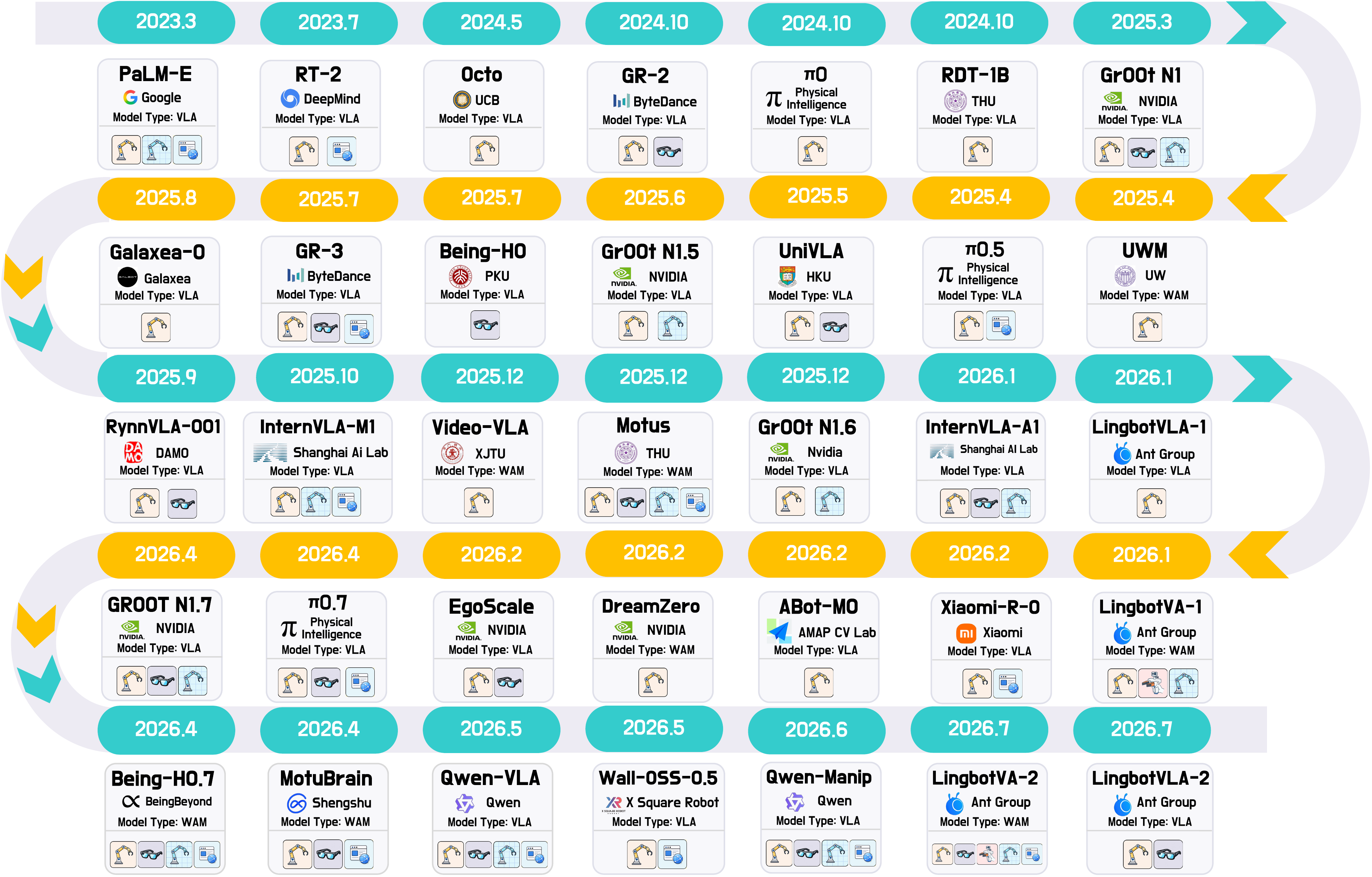}
    \vspace{-0.55cm}
    \caption{\textbf{Evolution of Data Utilization.} Evolution of data utilization in representative embodied foundation models. Early systems primarily rely on real-robot trajectories, whereas recent systems increasingly explore co-training with real-robot, egocentric, simulation, general, and UMI-style data. The architectural landscape also expands from action-centric VLA systems toward world-model-augmented and unified world-action models. Data sources are indicated by \DataLegendItem{\DataReal}{real-robot data}, \DataLegendItem{\DataEgo}{egocentric data}, \DataLegendItem{\DataUMI}{UMI data}, \DataLegendItem{\DataSim}{simulation data}, and \DataLegendItem{\DataGen}{general data}.}
    \label{fig:data-application-evolution}
\end{figure}

We further examine embodied foundation models from a data-centric perspective. As shown in Figure~\ref{fig:data-application-evolution}, we examine the data sources used during the pretraining of representative embodied foundation models. The reviewed systems indicate a transition from early training recipes dominated by real-robot trajectories toward increasingly heterogeneous mixtures of real-robot, simulation, general, egocentric, and, less frequently, UMI-style data. A more comprehensive summary of the models covered and their data compositions is provided in Table~\ref{tab:vla-wam-methods}.

Beyond documenting these data recipes, we analyze how embodied foundation models accommodate heterogeneous datasets from two perspectives: action-space alignment and geometric alignment. The former concerns how models reconcile differences in action dimensionality, control interfaces, and action semantics across robot embodiments, whereas the latter concerns how observations and demonstrations collected from different viewpoints, coordinate systems, and embodiments are mapped into robot-compatible representations. We then examine how three representative model families draw on different layers of the data pyramid. Embodied brain models primarily use broad multimodal and embodied data to support perception, spatial and temporal reasoning, affordance understanding, memory, and high-level planning~\cite{mu2023embodiedgpt,robobrain2025,rynnbrain2026}. Vision-language-action models require robot-compatible action supervision to map observations and instructions to executable behaviors~\cite{brohan2023rt2visionlanguageactionmodelstransfer,kim2024openvla,bjorck2025gr00t}. World action models additionally exploit action-free temporal data, action-conditioned interaction, and synthetic experience to model environment evolution and action consequences~\cite{vjepa2,dreamdojo,CosmosPredict,cen2025worldvla,hou2026world}. This analysis connects the composition and alignment of heterogeneous training data with their roles across different embodied foundation-model architectures.

Finally, we organize the key challenges and future directions for embodied data around three questions: \emph{what to collect, how to collect it, and how to use it}. First, future datasets should better represent interaction signals that remain scarce, particularly failure and recovery trajectories~\cite{black2026pi0visionlanguageactionflowmodel,dai2025racer,lu2025robofac} and tactile feedback that captures contact states, forces, material properties, and fine-grained interaction dynamics~\cite{zheng2026omnivta,dream-tac,tac2real2026}. Second, data-collection pipelines should reduce their dependence on costly manual teleoperation, broaden task and environmental coverage, and adaptively prioritize underrepresented or informative interactions~\cite{brohan2022rt,bu2025agibot,chi2024universal}. Third, heterogeneous data should be used more systematically by aligning action representations across embodiments, determining effective mixtures of data from different pyramid layers, and transferring human interaction experience to robot manipulation~\cite{o2024open,bjorck2025gr00t,seo2024legato,kareer2025egomimic,yang2025egovla}. 

To complement this analysis with a reusable community resource, we construct DataPyramid by converting selected subsets from six robot-dataset families into a common LeRobot format and applying embodiment-aware quality checks. The resulting collection contains 10,207.57 hours of robot trajectories across 1,721 repositories; a transparent frame-level validity mask identifies 135.12 hours for exclusion and retains 10,072.46 hours for training-time selection. Physical-robot and simulation sources remain explicitly distinguished. This resource addresses the practical cost of integrating heterogeneous data without conflating storage standardization with action-space equivalence (Section~\ref{sec:opendataset}).

Overall, this work organizes the embodied data ecosystem through a data-pyramid taxonomy, reviews the construction and collection of each data category, and examines how heterogeneous sources are utilized by embodied foundation models.

The main contributions of this work are summarized as follows.

\begin{itemize}
    \item \textbf{A data-pyramid taxonomy guided by scalability and robot alignment.} 
    We organize embodied AI data into five complementary categories, ordered from apex to base: real-robot data, UMI-style data, egocentric and exocentric data, simulation data, and general data. We characterize each category in terms of quality, diversity, reusability, physical fidelity, collection scalability, and alignment with physical robot learning. To support the community, we also release and maintain an open-source repository that curates representative datasets and related resources under this taxonomy.

    \item \textbf{A data-centric analysis of embodied foundation models.}
    We examine how heterogeneous data sources support embodied vision-language models, vision-language-action models, and world action models. Our analysis connects dataset properties, including semantic coverage, temporal structure, action supervision, embodiment alignment, diversity, and physical fidelity, with model capabilities such as perception, reasoning, planning, action generation, and world prediction.

    \item \textbf{A discussion of open challenges and future directions.}
    We identify key challenges in embodied data, including quality and diversity assessment, cross-embodiment reuse, multimodal standardization, scalable real-world collection, human-to-robot and sim-to-real transfer, failure and recovery data curation, and the integration and evaluation of heterogeneous data sources. We further discuss opportunities for constructing more general, robust, reusable, and physically grounded embodied agents.

    \item \textbf{A unified and auditable 10k-hour robot-data resource.}
    We convert and quality-annotate 10,207.57 hours of robot trajectories from six dataset families in a common LeRobot format, covering 972,324 episodes. Four complementary checks address motion discontinuities, forward-kinematics inconsistency, extreme values, and state--action disagreement. Together with basic validity checks, they identify 135.12 hours for mask-based exclusion and retain 10,072.46 hours. The accompanying conversion and cleaning pipeline, versioned configuration, and manifests make heterogeneous physical-robot and simulation sources easier to load, inspect, and reuse across the community.

\end{itemize}

\section{Real-Robot Data}

\begin{figure}[t]
    \centering
    \includegraphics[width=\textwidth]{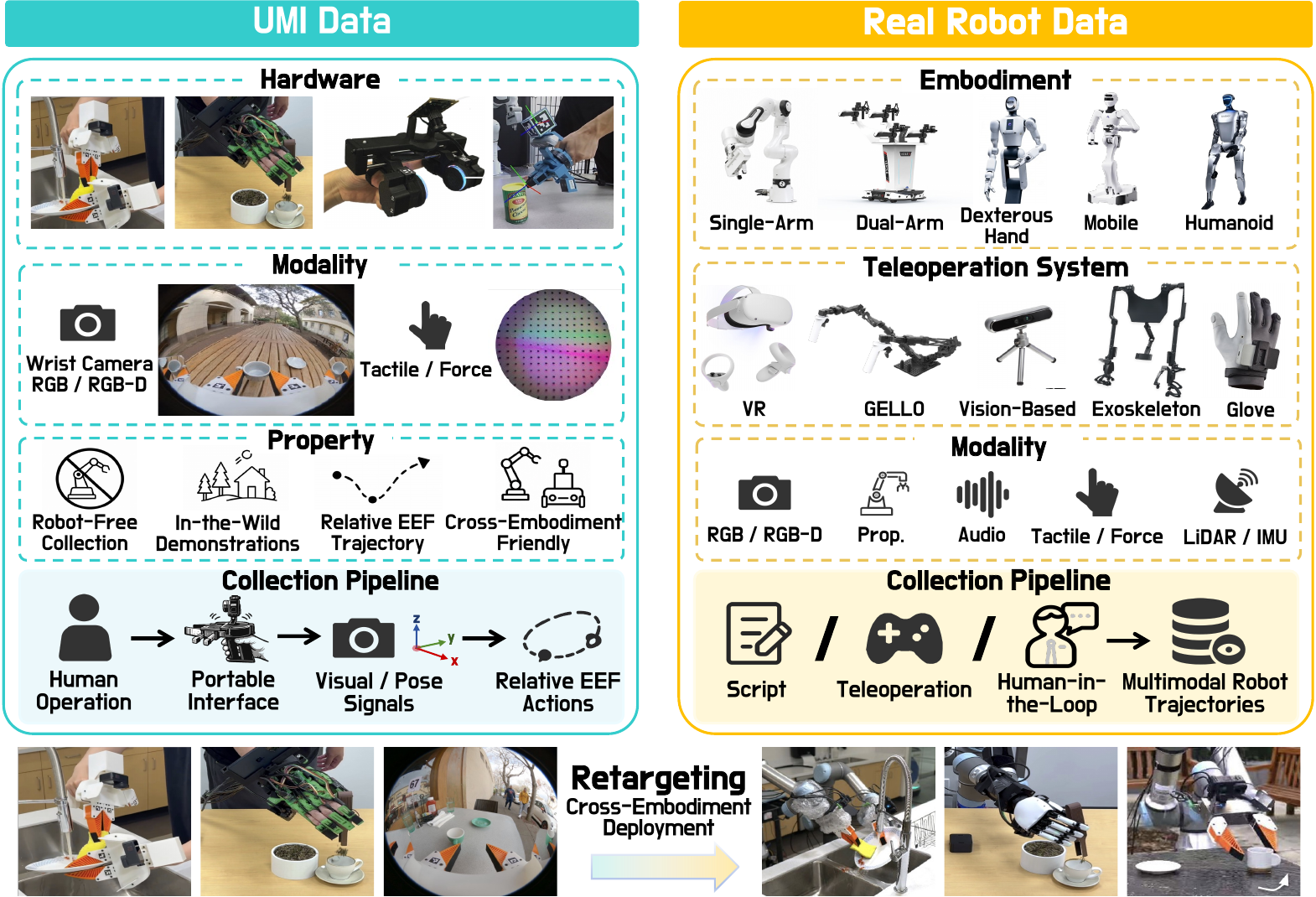}
    \vspace{-0.55cm}
    \caption{Overview of UMI data and real-robot data. \textbf{Left}: UMI data are collected with portable human-operated interfaces, emphasizing robot-free in-the-wild demonstrations, multimodal sensing, and relative end-effector trajectories. \textbf{Right}: real-robot data are collected on physical robot embodiments with diverse teleoperation systems, sensing modalities, and collection paradigms. Retargeting connects the two by transferring UMI demonstrations to executable robot actions for cross-embodiment deployment.}
    \label{fig:umi_real_robot}
\end{figure}

\subsection{Data Overview}

Real-robot data occupies the apex of the embodied-intelligence data pyramid, representing the layer closest to real-world execution. It is collected through the physical interaction between a robot and the environment, typically comprising sensory observations, robot states, and control actions. In contrast to human videos or embodiment-agnostic demonstration data, it directly captures the closed-loop relationship between perception, action, and physical consequences under a specific robotic embodiment. Consequently, it is characterized by high quality, physical fidelity, diversity, and reusability, making it the most direct source of supervision for learning executable manipulation policies and grounding model predictions in real-world dynamics.

The development of real-robot datasets has evolved from isolated, task-specific collections toward increasingly general and reusable data resources, as shown in Table~\ref{tab:real-robot-datasets}. Early efforts primarily focused on large numbers of trials for a small set of behaviors, such as robotic grasping, often collected using scripted controllers in controlled environments~\cite{pinto2016supersizing,kalashnikov2018scalable}. Subsequent datasets expanded toward human-operated demonstrations covering broader task sets and more diverse scenes~\cite{sharma2018multiple,mandlekar2018roboturk,ebert2021bridge,khazatsky2024droid}. More recent efforts further scale data collection across institutions, robot platforms, and application domains~\cite{o2024open,robomind2025,wu2025robocoin,hou2025robomind}, incorporating bimanual systems~\cite{agibotworld2026,fourier2025actionnet}, mobile manipulators~\cite{fu2024mobile}, humanoids~\cite{zhao2025humanoid,unitree2026unifolm}, and dexterous hands~\cite{zhang2026dexora,li2026deco}. This progression reflects a broader shift from collecting data for individual policies toward constructing large-scale data ecosystems for training generalist robot models.

Alongside this scaling trend, a complementary line of work specializes real-robot collection toward interaction regimes that generic household datasets underrepresent. PRISM~\cite{yu2026prism} targets high-precision, contact-rich industrial operations, recording more than 5,000 teleoperated trajectories over 25 manipulation tasks with synchronized multi-view RGB-D, force/torque, tactile, and robot-state measurements. SoftVTBench~\cite{jing2026softvtbench} instead focuses on deformable objects, pairing multi-view RGB, dual-finger tactile images and marker motion, proprioception, and language with evaluator-only finite-element states, so that physical interaction quality can be assessed independently of task success. A third direction adds structure rather than trajectories: RoboInter1.5~\cite{wang2026robointer15} re-annotates over 230K manipulation episodes with more than ten types of dense per-frame intermediate representations, including subtasks, primitive skills, object and gripper grounding, affordance, grasp poses, contact points, and motion traces. Together these efforts suggest that the marginal value of new real-robot data increasingly lies in covering underrepresented contact regimes and in enriching annotation depth, rather than in raw trajectory count alone.

Annotation enrichment can also improve the geometric value of existing demonstrations. DROID-3D~\cite{bardhan2026depthworld} augments over 70,000 DROID episodes with learned stereo depth and robot-grounded multi-view calibration, jointly optimizing shared robot parameters and scene-specific extrinsics. These annotations support RGB-depth prediction in DepthWorld, but the depth remains estimated and the held-out-trajectory evaluation uses the same robots and laboratories as training. The contribution therefore expands supervision quality rather than the number of independently collected physical interactions. Temporal enrichment can instead use the robot state stream: KiRoSe~\cite{xie2026kirose} learns transition-supervised sparse motion tokens and uses a language model to segment kinematic trajectories, retaining the need for annotated training transitions. FLORA~\cite{leng2026flora} complements this approach with models for timestamped captions and outcome or process reward labels, supplying semantic supervision for filtering and post-training rather than additional physical trajectories.

GazeAct~\cite{pani2026gazeact} enriches real-robot demonstrations with the teleoperator's gaze on the robot-camera display, synchronizing attention targets with images, proprioception, and actions. Its reported collection contains 12,000 demonstrations across 40 tasks; gaze is propagated between calibrated views to supervise attention during training, rather than required during deployment. This is robot-view teleoperation supervision, not an additional human egocentric corpus. Its collection scale also does not establish public availability of the full dataset, which was not verified in the checked release sources.

\begin{table}[!t]
\centering
\scriptsize
\setlength{\tabcolsep}{2.0pt}
\renewcommand{\arraystretch}{1.15}
\caption{\textbf{Statistics and key properties of real-robot manipulation datasets.} ``Embod.'' denotes the number of distinct embodiment types; ``Calib.'' denotes calibrated camera extrinsics. ``Dex.'' denotes dexterous-hand data. ``Anno.'' denotes subtask labels within episodes; ``Mobile'' denotes mobile manipulation data. In the ``Arm'' column, `S', `D', `H' denote single-arm, dual-arm, and Humanoid embodiments.}
\vspace{-0.2cm}
\label{tab:real-robot-datasets}
\rowcolors{2}{white}{w_blue!7}
\begin{tabular}{@{}
>{\raggedright\arraybackslash}m{0.21\textwidth}
>{\centering\arraybackslash}m{0.03\textwidth}
>{\centering\arraybackslash}m{0.035\textwidth}
>{\centering\arraybackslash}m{0.035\textwidth}
>{\centering\arraybackslash}m{0.113\textwidth}
>{\centering\arraybackslash}m{0.04\textwidth}
>{\centering\arraybackslash}m{0.33\textwidth}
>{\centering\arraybackslash}m{0.03\textwidth}
>{\centering\arraybackslash}m{0.03\textwidth}
>{\centering\arraybackslash}m{0.03\textwidth}
>{\centering\arraybackslash}m{0.04\textwidth}
@{}}
\toprule
\textbf{Dataset} & \textbf{Time} & \textbf{Arm} & \textbf{Embod.} & \textbf{Traj. / Hours} & \textbf{Tasks} & \textbf{Modality} &  \textbf{Calib.} & \textbf{Dex.} & \textbf{Anno.} &  \textbf{Mobile}  \\
\midrule\midrule
Pinto and Gupta~\cite{pinto2016supersizing} & 2015 & S & 1 & 50K / 700 & 1 & RGB; Proprio.; Act.; Grasp Succ. & \ding{55} & \ding{55} & \ding{55}  & \ding{55} \\
DAML~\cite{yu2018one} & 2018 & S & 2 & 2.9K / 4.08 & 3 & RGB; Proprio.; Act. &  \ding{55} & \ding{55} & \ding{55} & \ding{55}   \\
MIME~\cite{sharma2018multiple} & 2018 & S/D & 1 & 8.3K / 13.7 & 20 & RGB-D; Proprio.  &  \ding{55} & \ding{55} & \ding{55} & \ding{55}    \\
QT-Opt~\cite{kalashnikov2018scalable} & 2018 & S & 1 & 580k / 800 & 1 & RGB; Proprio.; Act.; Grasp Succ. & \ding{55} & \ding{55} & \ding{55} & \ding{55}  \\
RoboTurk~\cite{mandlekar2018roboturk} & 2018 & S & 3 & 2.1K / 111.3 & 3 & RGB-D; Proprio.; &  \ding{55} & \ding{55} & \ding{55} &  \ding{55}  \\
RoboNet~\cite{dasari2020robonet} & 2019 & S & 7 & 162k / 833.3 & - & RGB; Proprio.; Act. &  \ding{55} & \ding{55} & \ding{55} &  \ding{55}   \\
MT-Opt~\cite{kalashnikov2021mt} & 2021 & S & 1 & 800k / - & 12  & RGB; Proprio.; Act. &  \ding{55} & \ding{55} & \ding{55} &  \ding{55}   \\
BridgeData~\cite{ebert2021bridge} & 2021 & S & 1 & 7.2k / - & 71 &  RGB; Proprio.; Act. & \ding{55} & \ding{55} & \ding{55} &  \ding{55}   \\
BC-Z~\cite{jang2022bc} & 2022 & S & 1 & 26K / 125 & 100 & RGB; Proprio.; Act.; Lang. & \ding{55} & \ding{55} &  \ding{55} & \ding{55}  \\
RT-1~\cite{brohan2022rt} & 2022 & S & 1 & 130K / - & 700+ & RGB; Proprio.; Act.; Lang. & \ding{55} & \ding{55} &  \ding{55} & \ding{51}  \\
FurnitureBench~\cite{heo2025furniturebench} & 2023 & S & 1 & 5.1K / 219.6 & 9 & RGB-D; Proprio.; Act. & \ding{55} & \ding{55} & \ding{55} &  \ding{55}  \\
RH20T~\cite{fang2023rh20t} & 2023 & S & 4 & 110K / - & 147  & RGB-D; Proprio.; Act.; Lang.; Tactile; Force; Audio & \ding{51} & \ding{55} & \ding{55}  & \ding{55} \\
RoboSet~\cite{bharadhwaj2024roboagent} & 2023 & S & 1 & 98.5K / - & 48  & RGB-D; Proprio.; Act.; Lang.; Force; Torque & \ding{55} & \ding{51} & \ding{51} &  \ding{55}  \\
BridgeData V2~\cite{walke2023bridgedata} & 2023 & S & 1 & 60.1K / - & 13 & RGB; Proprio.; Act.; Lang. & \ding{55} & \ding{55} & \ding{55} & \ding{55}  \\
Open X-Embodiment~\cite{o2024open} & 2023 & S/D & 22 & 2.4M / - & 527 & RGB-D; Proprio.; Act.; Lang.; Force &  \ding{55} & \ding{55} & \ding{55} & \ding{55} \\
ALOHA Unleashed~\cite{zhao2024aloha} & 2024 & D & 1 & 26.2K / - & 5  & RGB; Proprio.; Act. & \ding{55} & \ding{55} & \ding{55} & \ding{55}  \\
DROID~\cite{khazatsky2024droid} & 2024 & S & 1 & 76K / 350 & 86 & RGB; Proprio.; Act.; Lang. & \ding{51} & \ding{55} & \ding{55} & \ding{55} \\
EgoMimic~\cite{kareer2025egomimic} & 2024 & D & 1 & 1K / 12 & 3 & RGB; Proprio.; Act. & \ding{51} & \ding{55} & \ding{55} & \ding{55}\\
FMB dataset~\cite{luo2025fmb} & 2024 & S & 1 & 22.6K / - & 2 & RGB-D; Proprio.; Act.; Force; Torque & \ding{55} & \ding{55} & \ding{55} & \ding{55} \\
RoboMIND~\cite{robomind2025} & 2024 & S/D & 4 & 107K / 305.5 & 479 & RGB; Proprio.; Act.; Lang.; Fail. labels & \ding{51} & \ding{51} & \ding{51} & \ding{55} \\
AgiBot World Beta~\cite{bu2025agibot} & 2025 & D & 1 & 1M / 2976.4 & 217 & RGB-D; Proprio.; Act.; Lang.; Force & \ding{51} & \ding{51} & \ding{51} & \ding{51} \\
REASSEMBLE~\cite{sliwowski2025reassemble} & 2025 & S & 1 & 4.6K / 13.02 & 121 & RGB; Proprio.; Act.; Audio; Force; Torque & \ding{51} & \ding{55} & \ding{51} & \ding{55} \\
PH2D~\cite{qiu2025humanoid} & 2025 & H & 1 & 1.6K /  -  & - & RGB; Proprio.; Act.; Lang. & \ding{55} & \ding{51} & \ding{55} & \ding{55} \\
AIST-Bimanual~\cite{aist2025bimanip} & 2025 & D & 1 & 10.7K / - & 119 & RGB; Proprio.; Act.; Lang. & \ding{55} & \ding{55} & \ding{55} & \ding{55} \\
ActionNet~\cite{fourier2025actionnet} & 2025 & H & 3 & 30K / 140 & - & RGB-D; Proprio.; Act.; Lang. & \ding{55} & \ding{51} & \ding{55} & \ding{55} \\
MotionTrans~\cite{yuan2025motiontrans} & 2025 & S & 1 & 1.5K / -  & 15 & RGB; Proprio.; Act.; Lang. & \ding{55} & \ding{51} & \ding{55} & \ding{55} \\
Open Galaxea~\cite{jiang2025galaxea} & 2025 & D & 1 & 20.5K / 500 & 150 & RGB-D; Proprio.; Act.; Lang. & \ding{55} & \ding{55} & \ding{51} & \ding{51} \\
RealSource World~\cite{realsourceworld} & 2025 & D & 1 & 26.7K / - & 35 & RGB; Proprio.; Act.; Lang.; Force; Torque &  \ding{51} & \ding{55} & \ding{51} & \ding{55} \\
RoboCOIN~\cite{wu2025robocoin} & 2025 & D & 15 & 180K / - & 421 & RGB-D; Proprio.; Act.; Lang. & \ding{55} & \ding{51} & \ding{51} & \ding{55} \\
Humanoid Everyday~\cite{zhao2025humanoid} & 2025 & H & 2 & 10.3k / 27.8+  & 260 & RGB-D; Proprio.; Lang.; Tactile; LiDAR; IMU; Odometry & \ding{55} & \ding{51} & \ding{55} & \ding{51} \\
RoboMIND 2.0~\cite{hou2025robomind} & 2026 & S/D & 6 & 310K / 1000+  & 739 & RGB-D; Proprio.; Act.; Lang.; Tactile; Force; Torque; Fail. labels &  \ding{51} & \ding{51} & \ding{51} & \ding{51} \\
DECO-50~\cite{li2026deco} & 2026 & D & 1 & 8K / 50+ & 28  & RGB; Proprio.; Act.; Tactile & \ding{55} & \ding{51} & \ding{55} & \ding{55} \\
Dexora~\cite{zhang2026dexora} & 2026 & D & 1 & 12.2K / 40.5 & 200 & RGB; Proprio.; Act.; Lang. & \ding{55} & \ding{51} & \ding{55} & \ding{55} \\
MolmoAct2-BimanualYAM~\cite{fang2026molmoact2} & 2026 & D & 1 & 34.5K / 720 & 736 & RGB; Proprio.; Act.; Lang. & \ding{55} & \ding{55} & \ding{55} & \ding{55} \\
MolmoAct2-SO100/101~\cite{fang2026molmoact2} & 2026 & S & 2 & 38K / 184 & 1315  & RGB; Proprio.; Act.; Lang. & \ding{55} & \ding{55} & \ding{55} & \ding{55} \\
AgiBot World 2026~\cite{agibotworld2026} & 2026 & D & 1 & 20.7K / - & - & RGB-D; Proprio.; Act.; Lang.; Tactile & \ding{51} & \ding{51} & \ding{51} & \ding{51}  \\
Unitree UnifoLM-WBT~\cite{unitree2026unifolm} & 2026 & H & 1 & 4.5K / 30.74 & 14 & RGB; Proprio.; Act.; Lang. & \ding{55} & \ding{51} & \ding{55} & \ding{51}  \\
Open-H-Embodiment~\cite{nelson2026open} & 2026 & S/D & 20 & 125.8K / 780 & 33 & RGB; Proprio.; Lang. &  \ding{55} & \ding{55} & \ding{55} & \ding{55} \\
Baihu-VTouch~\cite{hua2026vtouch++} & 2026 & S/D & 3 & - / 1000+ & 380+ & RGB-D; Proprio.; Act.; Lang.; Tactile & \ding{55} & \ding{51} & \ding{55} & \ding{51} \\
LET-Base-Dataset~\cite{lejurobotics2026letbase} & 2026 & H & 2 & 91.5K / 1000+ & 31 & RGB-D; Proprio.; Lang.; IMU & \ding{55} & \ding{51} & \ding{51} & \ding{55} \\
LET-Dex-Dataset & 2026 & H & 1 & 3K / 20 & 14 & RGB-D; Proprio.; Lang.; Tactile; Force; Torque; IMU & \ding{55} & \ding{51} & \ding{55} & \ding{55} \\
Haptile~\cite{alian2026haptile} & 2026 & S & 1 & 1.7K / 12.5 & 38 & RGB; Proprio.; Act.; Lang.; Tactile & \ding{55} & \ding{55} & \ding{55} & \ding{55} \\
OmniViTac~\cite{zheng2026omnivta} & 2026 & S & 1 & 21.9K / - & 86 & RGB-D; Proprio.; Act.; Lang.; Tactile & \ding{55} & \ding{55} & \ding{55} & \ding{55} \\
ABC-130K~\cite{abc2026} & 2026 & D & 1 & 130.7K / 3591 & 197 & RGB; Proprio.; Act.; Lang.; Torque & \ding{55} & \ding{55} & \ding{51} & \ding{55} \\
\bottomrule
\end{tabular}%
\end{table}

\subsection{Embodiment and Modalities}
Real-robot datasets vary substantially in the physical embodiments through which interaction data are collected. Early datasets were predominantly built around fixed-base, single-arm manipulators equipped with parallel grippers, as exemplified by large-scale grasping and manipulation datasets such as Pinto and Gupta, QT-Opt, and RoboNet~\cite{pinto2016supersizing,kalashnikov2018scalable,dasari2020robonet}. This configuration offers a relatively constrained action space and facilitates scalable data collection, but covers only a limited range of whole-body and coordinated manipulation behaviors. Later datasets increasingly incorporate dual-arm systems for coordinated and bimanual manipulation~\cite{zhao2023learning,zhao2024aloha,aist2025bimanip}, mobile manipulators that jointly control the base and upper body~\cite{fu2024mobile,jiang2025galaxea}, and humanoid platforms involving whole-body motion and locomotion~\cite{he2024omnih2o,fu2024humanplus,zhao2025humanoid}. More recent efforts further extend data collection to dexterous hands and heterogeneous robot fleets, enabling the study of fine-grained contact-rich manipulation and cross-embodiment learning~\cite{wu2025robocoin,hou2025robomind,zhang2026dexora}.

Another notable trend is the increasing richness of sensing modalities in real-robot datasets. Beyond the standard vision-proprioception-action interface, several datasets incorporate additional signals to capture aspects of physical interaction that cannot be reliably inferred from images alone. Tactile sensing has emerged as an important complementary modality for contact-rich and dexterous manipulation, as it provides local, surface-level information about contact location, pressure distribution, incipient slip, grasp stability, and object-hand interaction states~\cite{fang2023rh20t,zhao2025humanoid,hou2025robomind,li2026deco,zheng2026omnivta,alian2026haptile,yu2026prism}. Recent datasets increasingly treat tactile sensing as a first-class evaluation signal rather than an auxiliary channel: SoftVTBench~\cite{jing2026softvtbench} synchronizes dual-finger tactile images and marker motion at 20\,Hz with independent finite-element ground truth, allowing slip and excessive compression to be measured even when a task nominally succeeds. In contrast, force and torque measurements typically characterize the net interaction wrench at the wrist, gripper, or robot joints, offering a more global view of contact forces and interaction dynamics~\cite{fang2023rh20t,sliwowski2025reassemble,luo2025fmb,yu2026prism}. Audio can further provide complementary cues about collisions, friction, and task progression~\cite{fang2023rh20t,sliwowski2025reassemble}. Whole-body platforms may additionally include IMU, LiDAR, and odometry signals to represent balance, locomotion, and spatial navigation~\cite{zhao2025humanoid,lejurobotics2026letbase}. These richer multimodal observations not only enable a more comprehensive characterization of robot and environment interaction, but also increase heterogeneity across datasets.

Proprioceptive and action representations also differ considerably across embodiments. Fixed-base manipulators commonly record joint positions (qpos), represented as an ordered vector of joint coordinates, or end-effector (EEF) poses, represented by the Cartesian position and orientation of the wrist or tool, together with gripper states. Correspondingly, actions are typically expressed in either joint space or Cartesian EEF space~\cite{dasari2020robonet,ebert2021bridge,heo2025furniturebench,bharadhwaj2024roboagent,fang2023rh20t,o2024open}. Dexterous hands, due to their high-dimensional articulated structures, are typically represented through finger joint positions~\cite{robomind2025,wu2025robocoin,li2026deco,zhang2026dexora}, sometimes complemented by fingertip poses to facilitate alignment with human data~\cite{yang2025egovla,fu2025metis,zheng2026egoscale}. Mobile and humanoid platforms further incorporate base velocity, torso configuration, and whole-body joint states~\cite{fu2024mobile,zhao2025humanoid,hou2025robomind,unitree2026unifolm}.

These differences affect not only the dimensionality of the learning problem, but also the extent to which data can be shared across platforms. Joint-space representations provide direct and precise control for articulated hands, but are strongly tied to robot morphology. In contrast, end-effector-centric representations can offer a more transferable interface for arm motion across embodiments, although they still require consistent coordinate frames and control conventions. In policy learning, proprioceptive states and actions are often normalized to maintain suitable numerical scales and facilitate learning across embodiments. Common choices include MinMax, Q01-Q99, and MeanStd normalization~\cite{wu2026foundationapplicationimprovingvla}.

\subsection{Data Collection Paradigms}
\subsubsection{Script}
Scripted collection broadly refers to data-generation processes in which robot behavior is executed without continuous human control. Rather than manually controlling each episode, human effort is shifted toward designing the task logic, control pipeline, and execution conditions. This separation between system design and episode-level execution enables repeated and parallel data collection at relatively low marginal cost. Depending on how robot actions are produced, scripted collection can be broadly categorized into rule-based execution, trajectory playback, and autonomous policy rollouts.

\noindent\textbf{Rule-Based Execution.}
In rule-based collection, robot behavior is generated through manually specified task logic, often implemented as action sequences, finite-state machines, or heuristic controllers. A task is typically decomposed into predefined stages, such as detecting a target, selecting an interaction pose, approaching the target, executing the grasp, and transporting the object to a designated region. Within this pipeline, inverse kinematics, motion planning, and trajectory optimization can be used to convert stage-level goals into kinematically feasible and collision-free robot motions, while visual observations or sensor thresholds determine target poses and trigger transitions between stages. Early large-scale grasping datasets adopted such structured pipelines to execute repeated grasp attempts and automatically record both successful and failed interactions~\cite{pinto2016supersizing,dasari2020robonet,wang2025fieldgen}. Rule-based execution is particularly effective for tasks with clearly specified objectives, structured workspaces, and reliable reset mechanisms. However, because both the task decomposition and motion-generation process rely on predefined assumptions, the collected behaviors may exhibit limited diversity and may not generalize well to unanticipated states or complex contact interactions.

\noindent\textbf{Trajectory Playback.}
Trajectory playback collects data by re-executing previously recorded robot motions. The source trajectories may originate from human teleoperation, kinesthetic teaching, motion planning or earlier policy rollouts. During playback, the robot can reproduce demonstrations under different object configurations or environmental conditions, and perturbations may be introduced into the initial state, trajectory, or control parameters to increase variation. This approach reduces the need for repeated human operation while preserving the overall structure of successful demonstrations. For example, RoboSet combines human-collected demonstrations with playback-based execution to expand its real-robot experience~\cite{bharadhwaj2024roboagent}. Nevertheless, playback is sensitive to discrepancies between the original and current environment, since open-loop reproduction may fail when object locations or contact conditions change substantially. Closed-loop correction and replanning are therefore often required for reliable reuse.

\noindent\textbf{Autonomous Policy Rollouts.}
A more adaptive form of automated collection uses learned or heuristic policies to interact with the environment and generate new trajectories. Such policies may be obtained through reinforcement learning, imitation learning, or prior rounds of data collection, and can be iteratively improved as additional experience becomes available. Large-scale systems such as QT-Opt~\cite{kalashnikov2018scalable} and MT-Opt~\cite{kalashnikov2021mt} use autonomous rollouts to collect interaction data across repeated task executions, enabling the dataset and policy to improve through a continual data-generation cycle. RoboSet~\cite{bharadhwaj2024roboagent} similarly includes heuristic policy rollouts alongside human demonstrations and trajectory playback. Compared with fixed scripts, autonomous rollouts can respond to observations and cover a broader range of states, including policy failures and recovery behaviors. However, their data distribution is strongly shaped by the capabilities and exploration strategy of the current policy, potentially reinforcing existing biases or producing large quantities of low-quality trajectories.

\subsubsection{Teleoperation}
Teleoperation collects robot demonstrations by allowing a human operator to continuously control the robot during task execution. Relative to scripted collection, it can more naturally capture adaptive, goal-directed, and contact-rich behaviors, particularly in tasks where suitable control policies are not yet available. As illustrated in Figure~\ref{fig:umi_real_robot}, teleoperation systems exhibit diverse configurations.

\noindent\textbf{Kinesthetic Teaching.}
Kinesthetic teaching provides the most direct form of human control: the operator physically guides the robot arm or end effector through the desired motion while the system records joint states, end-effector poses, and gripper commands. This approach requires little additional interface hardware and allows operators to specify precise trajectories through direct physical interaction, as demonstrated in datasets such as MIME~\cite{sharma2018multiple}. It is particularly suitable for fixed-base manipulators and tasks requiring accurate geometric guidance. However, physical manipulation of the robot can impose substantial operator effort, and the approach is difficult to apply to high-payload systems, mobile platforms, or complex whole-body embodiments. HALO~\cite{yang2026halocollection} extends kinesthetic teaching to coordinated mobile manipulation by coupling direct guidance of the arms with a passive collection platform and aligning its recorded motion to the powered base used at deployment.

\noindent\textbf{Leader-Follower Teleoperation.}
Leader-follower systems use a dedicated input mechanism whose motion is mapped to a follower robot. When the leader and follower have identical or closely matched kinematic structures, joint-level correspondence provides an intuitive and precise control interface. This paradigm has been widely adopted for bimanual manipulation, as exemplified by ALOHA and its subsequent extensions~\cite{zhao2023learning,zhao2024aloha}, as well as larger bimanual datasets such as AIST-Bimanual~\cite{aist2025bimanip} and Open Galaxea~\cite{jiang2025galaxea}. GELLO~\cite{wu2024gello} further demonstrates that the leader can be implemented as a low-cost, scaled, and kinematically equivalent replica of the target arm, using 3D-printed components and inexpensive actuators while retaining direct joint-level correspondence. Leader-follower control is well suited to coordinated two-arm behaviors because it preserves temporal synchronization between both arms and reduces the complexity of Cartesian control. Its main limitation is the dependence on specialized, and often embodiment-specific, hardware, which can constrain portability across robot platforms.

\noindent\textbf{Integrated Leader-Follower Teleoperation.}
Integrated leader-follower teleoperation combines the input device and controlled robot within the same mechanism: the operator steers a gravity-compensated arm through an end-effector handle while the system records robot states, visual observations, and interaction signals. This method provides direct visual and force feedback, supports precise contact-rich manipulation, and avoids the kinematic-mapping errors of separate leader arms. Eliminating the dedicated leader device also reduces hardware, calibration, and workspace requirements. A representative practical example is the ARX AC-One, which has recently been used as a real-world platform for bimanual manipulation research~\cite{bi2025motusunifiedlatentaction,cai2026internvla}. The main drawback is that the operator's hand may appear in training images but be absent during autonomous deployment, creating a visual distribution shift. However, existing evidence from human-to-robot~\cite{yang2025egovla,zheng2026egoscale} transfer suggests that models can learn to accommodate such differences during pretraining.

\noindent\textbf{Device-Mediated Teleoperation.}
A more portable alternative uses generic input devices, including VR or XR controllers, smartphones, SpaceMouse devices and joysticks. Human inputs are commonly interpreted as Cartesian end-effector poses or velocities and converted into robot joint commands through inverse kinematics or low-level controllers. RoboTurk~\cite{mandlekar2018roboturk}, for example, uses a mobile device for six-degree-of-freedom control, while DROID~\cite{khazatsky2024droid} employs VR controllers to collect manipulation demonstrations across diverse environments. SpaceMouse and haptic devices have also been used in datasets such as FMB~\cite{luo2025fmb} and REASSEMBLE~\cite{sliwowski2025reassemble}. Touchscreens provide an even lower-cost channel for planar tasks: a recent learning-from-demonstration dataset records 3,142 handwriting demonstrations from 22 participants across all 52 Latin character-case combinations through a touchscreen teleoperation interface, capturing planar position, contact force, and timing~\cite{kenan2026robot}. For mobile manipulators, foot pedals can provide an additional hands-free control channel for commanding base motion, allowing the operator to navigate the platform while keeping both hands available for arm manipulation. Because these devices are largely independent of robot morphology, they can be deployed across different manipulators with relatively limited hardware modification. Yet their indirect control interfaces may introduce latency, scaling ambiguities, and reduced precision, especially for coordinated, high-dimensional, or contact-intensive tasks. A recent trend is to push this hardware independence from the input device to the whole collection rig: ModPack~\cite{citron2026modpack} packages computation, power, communication, and storage into a self-contained wearable backpack and exposes plug-and-play capability modules for joint-level teleoperation with haptic feedback, mobile manipulation, and active perception, so that the same interface can be reused across robot platforms and task domains.

\noindent\textbf{Vision-Based Teleoperation.}
Vision-based systems infer human body, arm, or hand motion directly from RGB or RGB-D observations, without requiring physical markers or dedicated wearable sensors. The estimated poses are then retargeted to the robot while accounting for kinematic feasibility and joint constraints. Such approaches can support both whole-body humanoid control and dexterous-hand teleoperation. HumanPlus~\cite{fu2024humanplus}, for example, transfers monocularly estimated human motion to a humanoid robot, while DexPilot~\cite{handa2020dexpilot} and Anyteleop~\cite{qin2023anyteleop} map visually estimated human hand motion to a dexterous robot hand. By reducing the need for specialized sensing hardware, vision-based teleoperation provides a low-cost and accessible interface, although its performance remains sensitive to occlusion, fast motion, and pose-estimation errors.

\noindent\textbf{Wearable and Motion-Capture Retargeting.}
Wearable systems capture human motion using devices such as inertial suits, optical trackers, data gloves, or arm and hand exoskeletons. These interfaces can provide higher-dimensional and more stable motion measurements than generic controllers, making them particularly suitable for bimanual, dexterous, and whole-body teleoperation. RoboMIND~\cite{robomind2025,hou2025robomind} combines multiple teleoperation interfaces, including motion-capture systems, to collect data across heterogeneous robot embodiments, while RoboCOIN~\cite{wu2025robocoin} extends wearable control to a large collection of dual-arm platforms. For dexterous manipulation, systems such as Dexora~\cite{zhang2026dexora} use arm exoskeletons and VR-based hand control to capture coordinated arm and finger motion. Whole-body teleoperation systems further map human upper-body, locomotion, and base motion to humanoid robots~\cite{unitree2026unifolm,zhao2025humanoid}, and comparable interfaces have been built for wheeled bimanual platforms, where a single VR interface coordinates base and dual-arm control to collect multimodal mobile-manipulation demonstrations~\cite{yang2026learning}. The required mapping depends on the sensing interface and the degree of correspondence between the human and robot embodiments. Exoskeleton-based systems often provide direct joint-level control, whereas data gloves and motion-capture devices typically require calibration and kinematic retargeting to account for differences in scale, joint ranges, and morphology. Glove-based sensing introduces a further, less obvious source of error: because the tactile elements must remain soft and flexible to preserve hand dexterity, they respond not only to contact force but also to hand pose, producing pose-related artifacts that raise the minimum detectable force and cause missed or late contact onsets in the low-force range. A glove-agnostic pose-aware framework~\cite{yu2026poseaware} characterizes these artifacts and mitigates them algorithmically by conditioning force estimation on hand pose, without modifying the glove, which is relevant to any pipeline that treats glove readings as contact ground truth.

\noindent\textbf{Feedback Mechanisms.} Feedback mechanisms constitute another important dimension of teleoperation-system design. Visual feedback provides the operator with observations of the environment, objects, and robot state through external or egocentric cameras. Immersive systems further employ stereoscopic displays and head-tracked active viewpoints to improve depth perception and spatial awareness~\cite{cheng2024open,ding2025bunny}. Force feedback transmits interaction forces or end-effector resistance from the robot to the operator through bilateral leader devices, haptic controllers, or exoskeletons, enabling more precise regulation of contact during insertion, assembly, and other force-sensitive tasks~\cite{sliwowski2025reassemble,zhang2025doglove}. Tactile feedback communicates localized contact events at the fingers or grasp interface, for example, through vibration, pressure, or fingertip stimulation, helping the operator perceive contact onset and adjust grasping behavior~\cite{ding2025bunny,zhang2025doglove,alian2026haptile}. The integration of these feedback channels improves teleoperation accuracy, stability, and operator awareness, enabling more natural control and higher-quality demonstrations, particularly in contact-rich tasks. A complementary direction moves in the opposite direction along the same loop, using the robot to assist the operator: goal-free intent-inference frameworks estimate which object or grasp region the operator is reaching for from partial motion, and can then bias assistance toward feasible grasps. GUIDER~\cite{kenny2026guider} evaluates such a framework on recorded real-robot manipulation data under the original temporal conditions, reporting a mean time to confident prediction of 3.7\,s and 96.4\% prediction stability across 20 manipulation steps. Shared-autonomy mechanisms of this kind can lower operator burden during long collection sessions, but they also shape the recorded trajectories, so the degree of assistance is itself a dataset property worth reporting.

\subsubsection{Human-in-the-Loop Enhancement}
Human-in-the-loop enhancement is an iterative real-robot data collection paradigm in which a human supervisor intervenes during policy execution and records corrective actions at difficult or failure-prone states. Rather than repeatedly collecting complete demonstrations, it selectively augments the dataset with policy-induced states, failure cases, and recovery trajectories that are underrepresented in the original training data.

Following the dataset-aggregation principle introduced by DAgger~\cite{ross2011reduction}, robot-learning methods have progressively improved the efficiency and scalability of this process. Intervention Weighted Regression~\cite{mandlekar2020human} collects corrective demonstrations through remote teleoperation and increases their contribution during training. ThriftyDAgger~\cite{hoque2021thriftydagger} selectively requests assistance according to state novelty and estimated failure risk, while Sirius~\cite{liu2025robot} continuously incorporates intervention data gathered during robot deployment. Fleet-DAgger~\cite{hoque2023fleet} extends this collection process to multiple concurrently operating robots. In dexterous manipulation, DexCap~\cite{wang2024dexcap} records human takeovers and residual corrections during policy rollouts. IntervenGen~\cite{hoque2024intervengen} subsequently improves data efficiency by generating additional corrective trajectories from a small number of human interventions. More recently, CR-DAgger~\cite{xu2026compliant} collects compliant delta-action corrections for contact-rich manipulation without fully interrupting autonomous execution. Overall, human-in-the-loop enhancement transforms errors encountered during real-robot deployment into targeted training data, improving the coverage of challenging states while reducing the need for full-trajectory demonstrations.

PrimeBot household data~\cite{xu2026scalingbimanual} illustrates the complementarity of expert demonstrations and deployment-time corrections. The collected corpus combines 531.7 hours of physical robot teleoperation with approximately 1,000 hours of bimanual UMI demonstrations, while the public release contains the real-robot data and staged UMI subsets. Clothes-folding experiments show diminishing returns from additional expert data and continued improvements from human interventions at policy-induced failure states, supporting targeted corrective collection rather than an unrestricted scaling claim.

SharpaDex v1.0~\cite{sharpadex_v1_2026} extends bimanual teleoperation toward fingertip-tactile supervision, pairing multi-view RGB with robot states, commands, tactile force/torque and tactile video, and temporally aligned language. Its official release reports 28,993 demonstrations across 59 tasks under CC BY-NC 4.0; the LeRobot v2.1 and v3.0 exports represent the same demonstrations and must not be counted twice.

\subsection{Scale and Diversity}
The scale of real-robot data has increased substantially over the past decade. Early datasets often accumulated large numbers of interactions within a narrowly defined task distribution. For example, Pinto and Gupta~\cite{pinto2016supersizing} and QT-Opt~\cite{kalashnikov2018scalable} collected tens or hundreds of thousands of grasping trials, but primarily focused on a single behavior. Subsequent efforts expanded both the number of trajectories and the breadth of tasks. MT-Opt~\cite{kalashnikov2021mt} contains approximately 800K episodes across 12 tasks, while RT-1~\cite{brohan2022rt} and RoboMIND~\cite{robomind2025} cover hundreds of manipulation skills with 130K and 107K trajectories, respectively. More recent datasets have reached million-scale collection or thousands of hours of interaction. AgiBot World Beta~\cite{bu2025agibot} reports one million trajectories and nearly 3,000 hours of data, while RoboMIND 2.0~\cite{hou2025robomind} contains more than 310K trajectories across 739 tasks and over 1,000 hours of execution. Cross-dataset initiatives such as Open X-Embodiment~\cite{o2024open} further aggregate more than two million trajectories from multiple institutions and robot platforms. The ArmnetBench rollout release~\cite{selvaraj2026rollout} illustrates a complementary source of diversity: it publishes 600 teleoperated demonstrations and 3,118 outcome-labelled policy rollouts on the same 12 tabletop tasks, including 2,288 failures.

For policy learning, data diversity is often more consequential than the raw number of trajectories, as it determines the range of states, behaviors, and interaction conditions represented during training. A clear trend in recent real-robot datasets is therefore the expansion of task coverage, from narrowly defined behaviors such as grasping or assembly to hundreds of household, industrial, mobile-manipulation, and dexterous tasks~\cite{wu2025robocoin,hou2025robomind,fang2026molmoact2,zhang2026dexora}. Task diversity alone, however, does not guarantee coverage of the interaction regimes a policy will face. Datasets remain concentrated on short-horizon, low-contact behaviors such as pick-and-place, which underrepresent the precision control and force or tactile regulation required by industrial assembly~\cite{yu2026prism}, and evaluation that reports only task success can hide whether a policy achieved that success while slipping or over-compressing the object~\cite{jing2026softvtbench}. Coverage should therefore be assessed along contact and precision axes as well as task counts. This task diversity is accompanied by increasing variation in objects, scene layouts, initial configurations and camera viewpoints, providing broader coverage of the state distributions encountered at deployment. At the same time, datasets increasingly span heterogeneous embodiments, including single-arm and bimanual manipulators, mobile platforms, humanoids, and dexterous hands, as well as richer sensing modalities such as RGB-D, force and torque, audio, IMU, LiDAR, and odometry~\cite{o2024open,wu2025robocoin,zhao2025humanoid,hou2025robomind}. Together, diversity across tasks, environments, embodiments, and modalities provides the foundation for learning policies that are more robust and transferable beyond a narrowly defined data-collection setting.

Beyond task, embodiment, and modality diversity, trajectory-level diversity captures variations in how the same nominal task is executed across demonstrations. Such variation may arise from differences in initial and terminal states, motion paths, contact sequences, and execution speeds, thereby broadening the state-action distribution covered within each task. The Cartesian distributions of keyframes in Figure~\ref{fig:diversity} provide an intuitive illustration of the resulting differences in spatial coverage and distributional structure. In the examples shown, AgiBot-World-2026~\cite{agibotworld2026} exhibits several well-separated spatial clusters, suggesting coverage of distinct workspace regions and interaction heights. RoboMIND~\cite{robomind2025} shows a more compact and anisotropic distribution, whereas RoboMIND 2.0~\cite{hou2025robomind} presents broader, partially connected, and vertically stratified spatial coverage.

\subsection{OpenDataset}
\label{sec:opendataset}

Beyond organizing existing resources, we contribute \textbf{DataPyramid}, a unified collection and processing pipeline that makes heterogeneous robot trajectories easier to reuse. We convert selected subsets from six dataset families---AgiBot World Beta, ABC-130k, AgiBot World 2026, Galaxea, InternData A1, and RoboMIND---into a common LeRobot v3 storage format and apply an auditable, embodiment-aware quality-filtering pipeline. The processed collection comprises \textbf{1,721 repositories, 972,324 episodes, and 10,207.57 hours} of robot trajectories. Its contribution is the consolidation of existing data into a common loading and quality-annotation interface, rather than the collection of additional demonstrations.

\paragraph{Scope and duration accounting.}
Table~\ref{tab:opendataset-hours} reports the exact subsets processed in this release, rather than the full scales advertised by the original datasets. Although presented here as a practical response to fragmentation in robot data, the collection also includes explicitly identified simulation subsets from InternData A1 and AgiBot World 2026. Consequently, the 10,207.57-hour total denotes \emph{robot-trajectory data}, not exclusively physical-robot interaction. The AgiBot World 2026 RL subset consists of real-robot trajectories and is therefore categorized as Real~\cite{agibotworld2026}. Source provenance remains available for constructing domain-specific training mixtures.

\begin{table}[t]
\centering
\small
\setlength{\tabcolsep}{4pt}
\renewcommand{\arraystretch}{1.12}
\caption{\textbf{DataPyramid conversion and filtering by processed source subset.} All durations are in hours. AgiBot World 2026 and InternData-A1 each aggregate their real-robot and simulation subsets. ``LeRobot'' is the measured converted duration entering filtering. ``Excluded'' is the duration marked invalid by the union of all checks, and ``Retained'' is the duration available when applying the validity mask. Values are rounded independently.}
\label{tab:opendataset-hours}
\begin{tabular}{@{}llrrr@{}}
\toprule
\textbf{Processed source} & \textbf{Type} & \textbf{LeRobot} & \textbf{Excluded} & \textbf{Retained} \\
\midrule
AgiBot World Beta~\cite{bu2025agibot} & Real & 2,539.82 & 14.25 & 2,525.57 \\
ABC-130k~\cite{abc2026} & Real & 2,716.32 & 5.26 & 2,711.07 \\
AgiBot World 2026~\cite{agibotworld2026} & Real / Simulation & 621.92 & 2.07 & 619.84 \\
Galaxea Lite / Pro~\cite{jiang2025galaxea} & Real & 462.33 & 11.00 & 451.34 \\
InternData-A1~\cite{interndataa12025} & Real / Simulation & 3,701.47 & 102.43 & 3,599.05 \\
RoboMIND~\cite{robomind2025} & Real & 165.70 & 0.11 & 165.59 \\
\midrule
\textbf{Total} & & \textbf{10,207.57} & \textbf{135.12} & \textbf{10,072.46} \\
\bottomrule
\end{tabular}
\end{table}

Durations are computed per repository using its recorded sampling rate, rather than a single rate for the entire collection. For repository $r$, let $N_r$ denote its frame count, $f_r$ its sampling rate in frames per second, and $m_{r,t}\in\{0,1\}$ the validity of frame $t$. We report
\begin{equation}
H_{\mathrm{converted}}=\sum_r\frac{N_r}{3600f_r},\qquad
H_{\mathrm{excluded}}=\sum_r\frac{\sum_t(1-m_{r,t})}{3600f_r},\qquad
H_{\mathrm{retained}}=H_{\mathrm{converted}}-H_{\mathrm{excluded}}.
\end{equation}
These are trajectory-equivalent hours: simultaneous camera streams are not counted as separate interaction time. Independent pre-conversion duration accounting is unavailable, so the table supports measured conversion-output and filtering statistics, not a claim of lossless recovery of every original source frame.

\paragraph{A common format with explicit embodiment semantics.}
The pipeline places state/action arrays, timestamps, episode indices, and task metadata into a consistent LeRobot organization, with Parquet shards for numerical streams and indexed video streams for visual observations. Dataset-specific adapters expose the relevant joint and end-effector fields while retaining their embodiment definitions, units, coordinate conventions, and availability masks. A unified file format therefore does not imply identical action dimensions or interchangeable control semantics. Original numerical columns and temporal indices are preserved during cleaning, and a Boolean \texttt{quality.valid} field supplies a common selection interface. This allows users to combine supported sources through the same data-loading workflow while retaining the information required for embodiment-specific processing.

\paragraph{Four complementary quality checks.}
We apply four frame-level checks on top of basic numerical, mask, and index validity checks. Thresholds are versioned in the \texttt{alpha10} configuration, and checks that cannot be evaluated are recorded as unavailable rather than treated as successful comparisons.
\begin{enumerate}
\item \textbf{Velocity and acceleration discontinuities.} We derive first- and second-order motion signals from timestamps within continuous episode segments. For each repository and channel, the upper threshold on absolute motion is $q_{0.99}+10(q_{0.99}-q_{0.01})$, with quantiles computed from finite, valid samples. Differences do not cross episode boundaries, index gaps, or invalid observations. Rotational motion uses shortest-arc rotation differences.
\item \textbf{Forward-kinematics consistency.} Where joint states, observed end-effector poses, and a supported kinematic model and frame mapping are available, we compare the predicted and recorded poses using position and orientation tolerances of 1\,cm and $2^\circ$. Missing observations or invalid end-effector masks cause the corresponding comparison to be skipped. Such checks diagnose consistency with the configured model and calibration; they do not establish task success.
\item \textbf{Extreme-value filtering.} For each eligible scalar channel, we retain values within $[q_{0.01}-10\Delta_q,\;q_{0.99}+10\Delta_q]$, where $\Delta_q=q_{0.99}-q_{0.01}$. Quantiles are fitted separately per repository and channel. Gripper, dexterous-hand, and quaternion components are not indiscriminately subjected to joint-position thresholds.
\item \textbf{State--action consistency.} For corresponding arm-joint position channels, we smooth state and action signals with a five-frame centered window and compare their velocity directions. An episode-level search over $\pm15$ frames compensates for temporal offset. A frame is marked inconsistent when directional agreement is below 0.5 among at least two jointly moving channels, each exceeding 0.01\,rad/s in both streams. Missing, stationary, and invalid-window comparisons are skipped.
\end{enumerate}

\begin{table}[t]
\centering
\small
\setlength{\tabcolsep}{5pt}
\renewcommand{\arraystretch}{1.12}
\caption{\textbf{Frame-level quality audit of the 10,207.57-hour collection.} Individual checks are measured independently against the same input; their hits overlap and must not be added. The final row is the union of all checks, including basic validity checks. Frame and duration percentages differ because repositories have different sampling rates.}
\label{tab:opendataset-filtering}
\begin{tabular}{@{}lrrrr@{}}
\toprule
\textbf{Check} & \textbf{Flagged frames} & \textbf{Frames (\%)} & \textbf{Flagged hours} & \textbf{Hours (\%)} \\
\midrule
Velocity / acceleration (union) & 2,935,547 & 0.2725 & 27.24 & 0.2669 \\
FK inconsistency & 9,992,235 & 0.9274 & 92.52 & 0.9064 \\
Extreme values & 483,277 & 0.0449 & 8.31 & 0.0814 \\
State--action inconsistency & 773,701 & 0.0718 & 8.83 & 0.0865 \\
\midrule
\textbf{All checks (union)} & \textbf{13,998,735} & \textbf{1.2992} & \textbf{135.12} & \textbf{1.3237} \\
\bottomrule
\end{tabular}
\end{table}

\paragraph{Release outcome and community use.}
The combined mask identifies 13,998,735 invalid frames, corresponding to 135.12 hours (1.3237\% of the input duration), and retains \textbf{10,072.46 hours} for mask-based selection. Table~\ref{tab:opendataset-filtering} separates the effects of individual checks from their deduplicated union. We retain all rows and video indices in the stored release rather than physically deleting flagged frames; users apply \texttt{quality.valid} together with the original modality masks when constructing training samples. This preserves alignment, supports alternative filtering policies, and makes exclusion decisions traceable through per-frame reports, fitted bounds, and coverage records. Temporal-window loaders should respect validity and episode boundaries rather than joining nonadjacent valid frames into artificial contiguous trajectories.

The conversion adapters, filtering configuration, repository manifest, and validation workflow form the reproducibility interface for this collection. Our practical contribution is to reduce repeated format-conversion and quality-inspection effort across laboratories, providing a common foundation for multi-source robot learning. The reported exclusion rate is an audit of the specified policy, not evidence of improved downstream policy performance or complete removal of all data defects. In particular, FK disagreement can reflect model or calibration mismatch and requires interpretation alongside the recorded coverage and source provenance.

\subsection{Advantages and Limitations}
Real-robot data provide physically grounded supervision by directly capturing the relationship between robot observations, actions, and their physical consequences. Compared with simulation or human videos, they naturally reflect real sensing noise, control latency, contact dynamics, and hardware constraints. Moreover, the recorded actions are directly executable on the corresponding platform, making such data particularly valuable for learning reliable manipulation policies and recovery behaviors.

The main limitation is the high cost of collection. Real-robot data require physical hardware, human operation or automated collection pipelines, repeated environment resets, maintenance, and safety supervision, which restrict large-scale and parallel data acquisition. In addition, many datasets remain concentrated on specific robots, tasks, and environments. Differences in embodiment, sensor configuration, coordinate systems, and action representation further complicate data aggregation and cross-platform transfer.

\begin{wbtakeaway}
\bReal~\textbf{Real-robot data} sits at the apex: it provides direct robot control signals together with observations of physical interactions between the robot and the real world. We believe that future real-robot data development should proceed along three directions. First, teleoperation systems should support simpler, smoother, and more intuitive control while preserving fine-grained trajectories and direct physical interaction feedback. In this respect, \emph{Integrated Leader-Follower Teleoperation} provides a promising direction because of its relatively low hardware cost, ease of use, and capacity for fluent demonstration collection. Second, real-robot datasets should capture broader in-the-wild variations in scenes, objects, and environmental conditions. AIGC may complement physical collection by synthesizing embodiment-specific data or augmenting scenes and observations, thereby expanding environmental diversity without requiring proportional increases in real-world deployment. Third, data collection should extend beyond human teleoperation to include policy rollouts, which reflect how an embodied agent explores the physical world under its own induced state distribution. Human-in-the-loop interventions during these rollouts can further transform failures into recovery trajectories, providing supervision for error correction, task continuation, and improved policy robustness.
\end{wbtakeaway}

\section{UMI Data}

\subsection{Data Overview}
Universal Manipulation Interface (UMI) data denotes a family of real-world manipulation datasets collected through portable, robot-independent interfaces rather than directly operating a target robot~\cite{chi2024universal}. A typical UMI system integrates one or more cameras, a hand-held gripper or dexterous interface, and a pose-tracking module to record visual observations, end-effector motion, gripper states, and, in some recent systems, force or tactile signals. These demonstrations can then be converted into robot-compatible trajectories through calibration, action transformation, and embodiment-specific retargeting. 

Compared with human egocentric videos, UMI data preserves natural, first-person observations of real-world manipulation while additionally providing synchronized end-effector poses, gripper states, and calibrated action trajectories. These structured signals yield high-quality, physically faithful demonstrations that are readily usable for robot policy learning and, with appropriate retargeting, cross-embodiment deployment. Since the original UMI system, this paradigm has expanded from single-arm gripper manipulation~\cite{liu2024maniwav,wu2024fast} to bimanual systems~\cite{nai2026humanoid}, dexterous hands~\cite{tao2025dexwild,xu2025dexumi,realdexumi}, mobile platforms~\cite{ha2024umi}, multi-view perception, 3D sensing, and tactile or force-aware data collection~\cite{li2025vitamin,zhu2025touch,wu2026tamen}. As summarized in Table~\ref{UMI dataset}, UMI-style datasets span a growing range of embodiments, sensing modalities, tasks, and collection scales.

\subsection{UMI Data Collection Pipeline}

\subsubsection{Evolution of UMI-Style Systems}
The evolution of UMI-style systems first manifests in the design of the physical collection interface. The original UMI established a robot-free collection paradigm based on a portable hand-held gripper that integrates wrist-mounted cameras, pose tracking, and gripper-state sensing to capture synchronized observations and actions during natural human demonstrations~\cite{chi2024universal}. Later systems improved the portability, modularity, and ease of deployment of this design. FastUMI~\cite{wu2024fast}, for example, simplifies the hardware and tracking pipeline to reduce setup effort and support more efficient large-scale collection, while LEGATO~\cite{seo2024legato} introduces a grasping-tool interface designed to provide a consistent demonstration representation across different robot platforms. More recent systems further reconsider how users physically interact with the collection device. In contrast to conventional UMI interfaces, where the operator holds a gripper-shaped tool and controls its opening through a trigger, FreeTacMan~\cite{wu2025freetacman} adopts a finger-worn interface that allows the operator to directly control the gripper using natural finger motion. This progression reflects a broader shift from mechanically mediated hand-held tools toward more natural, wearable, and contact-aware collection interfaces. UMI-Live~\cite{velamoor2026umilive} combines self-contained LiDAR--inertial sensing and asynchronous reconstruction, releasing 3,255 bimanual episodes across three tasks.

The data produced by UMI-style systems has also become increasingly diverse in terms of embodiment, sensing modality, and scale. Early UMI datasets primarily focused on single-arm manipulation with parallel grippers and wrist-view RGB observations. Subsequent work extended this paradigm to mobile manipulators and whole-body humanoid systems, as in UMI on Legs~\cite{ha2024umi} and HuMI~\cite{nai2026humanoid}, as well as to bimanual manipulation and dexterous hands~\cite{tao2025dexwild,xu2025dexumi}.

The observation space has similarly expanded beyond monocular wrist-view images. MV-UMI~\cite{rayyan2025mv} introduces additional camera viewpoints, while UMI-3D~\cite{wang2026umi} incorporates explicit 3D spatial observations. ViTaMIn~\cite{liu2025vitamin}, FreeTacMan~\cite{wu2025freetacman}, exUMI~\cite{xu2025exumi}, and ManipForce~\cite{lee2025manipforce} further augment visual trajectories with tactile or force-related measurements, enabling the collection of contact-rich manipulation data. Soft YUBI~\cite{kelemen2026softyubi} captures visible finger deformation as a proxy for grasp force, while RUMI~\cite{avdiu2026rumi} pairs interchangeable fingertip sensors with augmented-reality feedback about kinematic feasibility. At the same time, improvements in collection efficiency have increased dataset scale from several hundred task-specific demonstrations to collections containing tens of thousands or even hundreds of thousands of trajectories, such as FastUMI-100K~\cite{liu2025fastumi100k}. Overall, UMI-style data has evolved from single-arm visual trajectories into a broader class of large-scale, multimodal, and multi-embodiment manipulation data.

Large-scale releases also mix collection paradigms. PrimeBot household data~\cite{xu2026scalingbimanual} combines approximately 1,000 collected hours of bimanual UMI demonstrations with a separately reported 531.7-hour real-robot subset. The staged public UMI release should therefore be distinguished from both the collected UMI total and the combined headline scale. AIRoA-0~\cite{wu2026airoa0} reports 16,151 hours of bimanual YUBI demonstrations with task and subtask labels and initializes its policy from an existing $\pi_{0.5}$ checkpoint.

\subsubsection{Relative Trajectory Action Representation}
Since UMI demonstrations are collected without directly operating a target robot, they do not contain embodiment-specific joint states or joint-space actions. Instead, UMI combines a wrist-mounted camera with an onboard IMU and visual–inertial SLAM to estimate the six-degree-of-freedom pose trajectory of the handheld gripper, along with its open–close state. To reduce sensitivity to the global tracking frame, future end-effector targets are represented relative to the current end-effector pose. This relative trajectory representation preserves the spatial structure of the demonstrated motion while reducing error accumulation during execution. It also provides an embodiment-agnostic action space that can be mapped to different robot platforms through calibration, inverse kinematics, and low-level control.

\begin{table}[t]
\centering
\scriptsize
\setlength{\tabcolsep}{2.0pt}
\renewcommand{\arraystretch}{1.15}
\caption{\textbf{Statistics and key properties of UMI datasets.} These are real-world manipulation datasets collected with portable handheld interfaces that enable scalable, low-cost demonstration collection across diverse tasks and environments.}
\vspace{-0.2cm}
\label{UMI dataset}
\rowcolors{2}{white}{w_blue!7}
\begin{tabular}{lcccccc}
\toprule
\textbf{Dataset} & \textbf{Time} & \textbf{Arm Type} & \textbf{Tasks} & \textbf{End Effector} & \textbf{Demos} & \textbf{Tactile} \\
\midrule\midrule
UMI~\cite{chi2024universal} & 2024 & Single / Bimanual & 5 & Gripper  & 2.5K & \ding{55} \\
ManiWAV~\cite{liu2024maniwav} & 2024 & Single & 5 & Gripper  & 1.0K & \ding{55} \\
UMI on Legs~\cite{ha2024umi} & 2024 & Single & 2 & Gripper  & 514 & \ding{55} \\
FastUMI~\cite{wu2024fast} & 2024 & Single & 22 & Gripper  & 9.0K & \ding{55} \\
Data Scaling Laws~\cite{lin2024data} & 2024 & Single & 6 & Gripper  & 24.1K & \ding{55} \\
LEGATO~\cite{seo2024legato} & 2024 & Single & 6 & Gripper  & 900 & \ding{55} \\
ViTaMIn~\cite{liu2025vitamin} & 2025 & Single & 8 & Gripper  & 841 & \ding{51} \\
DexWild~\cite{tao2025dexwild} & 2025 & Single & 10 & Dexterous Hand  & 9.5K & \ding{55} \\
DexUMI~\cite{xu2025dexumi} & 2025 & Single & 5 & Dexterous Hand  & 1.8K & Force / Torque \\
FreeTacMan~\cite{wu2025freetacman} & 2025 & Single & 50 & Gripper  & 10K & \ding{51} \\
Touch in the Wild~\cite{zhu2025touch} & 2025 & Single & 20 & Gripper  & 2.7K & \ding{51} \\
exUMI~\cite{xu2025exumi} & 2025 & Single & 8 & Gripper  & 1.7K & \ding{51} \\
MV-UMI~\cite{rayyan2025mv} & 2025 & Single & 4 & Gripper  & 1.4K & \ding{55} \\
ManipForce~\cite{lee2025manipforce} & 2025 & Single & 6 & Gripper  & 597 & Force / Torque \\
FastUMI-100K~\cite{liu2025fastumi100k} & 2025 & Single & 32 & Gripper  & 100K & \ding{55} \\
ViTaMIn-B~\cite{li2025vitamin} & 2025 & Single & 4 & Gripper  & 884 & \ding{51} \\
HuMI~\cite{nai2026humanoid} & 2026 & Bimanual & 5 & Gripper  & 827 & \ding{51} \\
RealOmni & 2026 & Bimanual & Diverse & Gripper  & 789.8K & \ding{51} \\
UMI-3D~\cite{wang2026umi} & 2026 & Single & 3 & Gripper  & 4.6K & \ding{55} \\
TAMEn~\cite{wu2026tamen} & 2026 & Bimanual & 4 & Gripper  & 724 & \ding{51} \\
Daimon-Infinity  & 2026 & Bimanual & Diverse & Gripper  & 274.7K & \ding{51} 
\\
\bottomrule
\end{tabular}
\end{table}

\subsection{Cross-Embodiment Deployment}
Cross-embodiment deployment aims to transfer policies learned from robot-free UMI demonstrations to robots with different kinematic structures and sensing configurations. The key idea is to separate an embodiment-agnostic task representation from embodiment-specific execution. Rather than predicting joint-space commands tied to a particular robot, UMI-style policies operate in an end-effector-centric action space, typically producing relative six-degree-of-freedom trajectories and gripper commands. During deployment, the predicted relative trajectory is composed with the robot's current end-effector pose and transformed into the robot base frame. The resulting target poses are then converted into executable joint commands through inverse kinematics, motion planning, or a low-level Cartesian controller.

This deployment strategy has been extended to robots with increasingly different morphologies. LEGATO~\cite{seo2024legato} maps demonstrations into a motion-invariant task space and retargets the predicted gripper motion to different fixed-base and mobile manipulators through whole-body inverse kinematics. For dexterous manipulation, however, transferring only the wrist trajectory is insufficient because human and robot hands differ substantially in finger morphology, degrees of freedom, and visual appearance. DexUMI~\cite{xu2025dexumi} addresses these gaps through a wearable exoskeleton that constrains human motions toward robot-feasible hand configurations, together with visual inpainting that replaces the human hand with the target robot hand in demonstration images. Other systems improve deployment consistency by using modular cameras or tactile sensors that can be shared between the collection interface and the robot~\cite{li2025vitamin,wu2026tamen}. These approaches progressively extend UMI data from gripper-level transfer toward cross-embodiment deployment across diverse arms, mobile manipulators, and dexterous hands.

\subsection{Advantages and Limitations}
The primary advantages of UMI data are its scalability, environmental diversity, and reduced dependence on a specific robot embodiment. In particular, its portable and robot-free collection setup enables in-the-wild data acquisition in diverse everyday environments. Human operators can collect smooth and natural manipulation demonstrations across diverse real-world settings, while end-effector-centric action representations facilitate reuse across different robot platforms. Recent extensions incorporating bimanual motion, dexterous hands, and tactile sensing further broaden the range of behaviors captured by UMI-style systems.

However, UMI data does not fully eliminate the embodiment gap. Differences in camera viewpoints, end-effector geometry, kinematic constraints, dynamics, and contact conditions may lead to discrepancies between demonstration and robot execution. Its quality also relies on accurate pose tracking and sensor synchronization, which may degrade under occlusion, rapid motion, or visually challenging environments~\cite{wang2026umi}. Moreover, UMI typically lacks robot-specific proprioception and actuator dynamics. Therefore, it is best regarded as a scalable complement to embodiment-specific real-robot data rather than a complete replacement.

\begin{wbtakeaway}
\bUMI~\textbf{UMI-style data} bridges human demonstration and robot execution through portable, robot-free interfaces, providing explicit end-effector action structure at far lower cost than teleoperation. Yet its reliability is limited by fragile visual tracking and calibration, the difficulty of validating trajectory quality without robot execution, and the lack of force feedback for contact-rich tasks.
\end{wbtakeaway}

\section{Egocentric and Exocentric Data}
\subsection{Data Overview}

Egocentric and Exocentric data occupy the middle layer of the embodied data pyramid, offering more authentic real-world observations and physical interactions than simulation data, but less direct alignment with robot execution than UMI data. Most egocentric datasets rely on a head- or body-worn camera that follows the wearer and keeps the ongoing activity in view. However, this moving viewpoint can leave critical aspects of the interaction unobserved: the hands may occlude contact regions, while much of the body and surrounding scene may remain outside the frame. Synchronized exocentric cameras are therefore often added to provide complementary views and a more complete record. Such data is readily scalable and offers high diversity, reusability, and physical fidelity, but its quality is often limited by the constraints of single-camera capture.

Beyond RGB video, egocentric datasets can be enriched through three complementary routes: multimodal sensing during capture, post-processing of the recorded streams, and conversion into robot-oriented representations. During capture, additional hardware provides complementary signals: H2O~\cite{kwon2021h2o} records synchronized RGB-D views. HoloAssist~\cite{wang2023holoassist} combines RGB and depth with head tracking, hand tracking, gaze, and audio. EgoEMG~\cite{egoemg2026} records muscle activity, while FEEL~\cite{feel2026} measures applied force and EgoTouch~\cite{egotouch2026} provides tactile pressure. ACE-Data-0~\cite{cao2026acedata0} pushes this route toward joint coverage of the full perception-action loop by turning real homes into spatially calibrated and temporally synchronized capture spaces at two scales, a table-scale configuration for hand-object manipulation and a room-scale configuration for whole-body motion and locomotion, recording egocentric and multi-view exocentric video together with full-body and articulated hand motion, object geometry and 6-DoF trajectories, audio, and tactile signals.

Post-processing further adds semantic and geometric structure. In EPIC-KITCHENS~\cite{Damen2018EPICKITCHENS,damen2020rescaling}, narration and manual annotation produce action segments and verb-noun labels. HOT3D~\cite{banerjee2024hot3d} uses calibrated motion capture to obtain hand and object poses. EgoDex~\cite{hoque2026egodex} combines headset tracking with SLAM to recover camera and hand trajectories, and DreamHand~\cite{liu2026dreamhand} recovers continuous metric bimanual trajectories from clips in which the hands are occluded or leave the frame entirely. EgoHand~\cite{chen2026egohand} adds a monocular reconstruction pipeline that predicts parametric hand geometry in camera coordinates; calibrated camera motion is still needed to obtain world-frame annotations, and reconstruction does not establish physically executable robot actions. EgoAIM, introduced with Gaze2HOI~\cite{ma2026gaze2hoi}, instead derives phase-specific intention and hand--object forecasting annotations from HOT3D-Aria recordings, adding annotation structure rather than independent capture hours; the checked release provides demo samples while the full benchmark remains pending acceptance.

\begin{figure}[t]
    \centering
    \includegraphics[width=\textwidth]{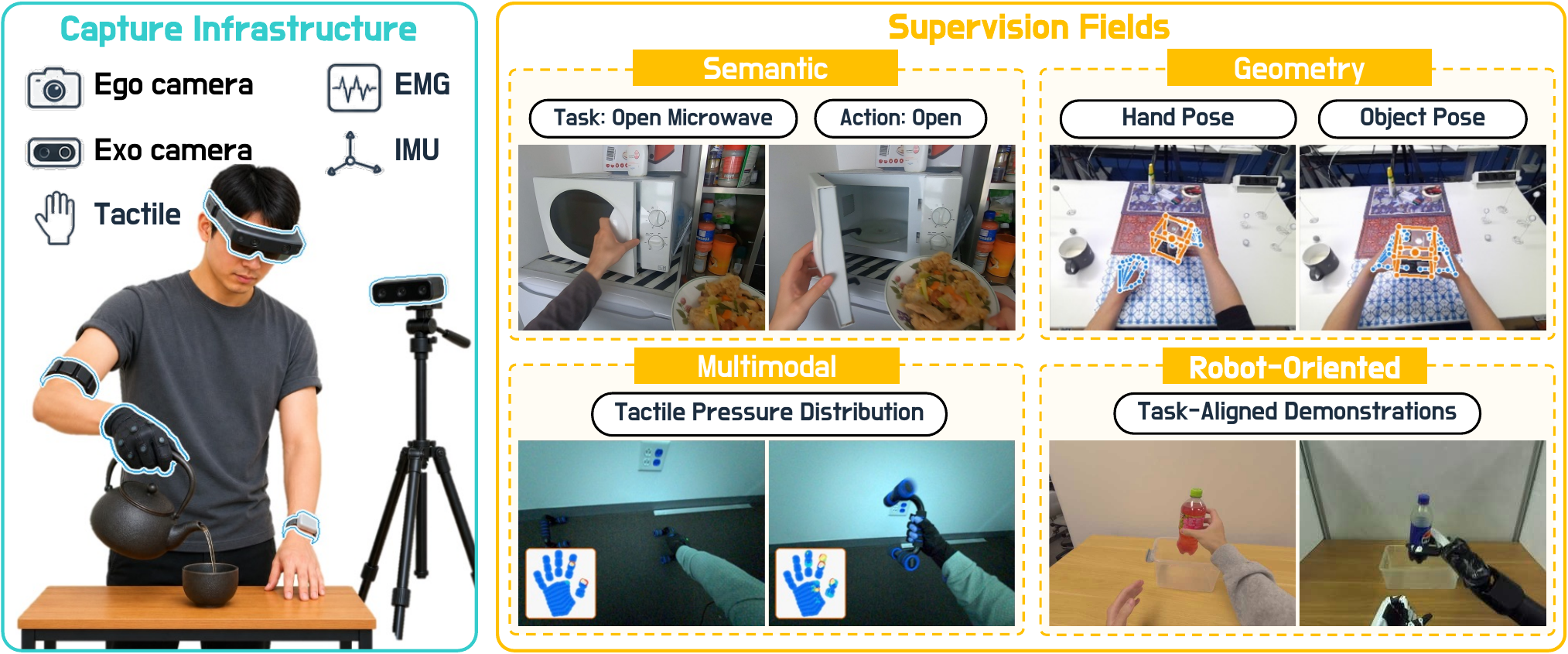}
    \vspace{-0.55cm}
    \caption{\textbf{Overview of capture infrastructure and supervision representations in egocentric and exocentric data.} Wearable and external cameras, motion trackers, and auxiliary sensors record complementary raw signals, which are subsequently annotated, reconstructed, or aligned to form semantic, geometric, multimodal, and robot-oriented supervision.}
    \label{fig:ego-data-overview}
\end{figure}

Finally, obtaining robot-oriented representations requires mapping human motion into robot-compatible action spaces: EgoMimic~\cite{kareer2025egomimic} aligns tracked human hand trajectories with robot demonstrations, whereas EgoVLA~\cite{yang2025egovla} uses inverse kinematics and retargeting to map human wrist and hand poses to robot actions. A parallel line of work instead extracts embodiment-agnostic interaction structure: VLAff~\cite{oh2026vlaff} derives visual, grasp, and trajectory affordances from egocentric video using structure-from-motion and hand-mesh reconstruction, and Pegasus~\cite{luo2026passive} converts passive human video into a graph-based intermediate representation that is transformed into robot-conditioned generations and filtered by a physics verifier. Figure~\ref{fig:ego-data-overview} illustrates the capture infrastructure and supervision types. Table~\ref{tab:ego-exo-datasets} summarizes representative egocentric and exocentric datasets and their key characteristics.

\begin{table}[!t]
\centering
\scriptsize
\setlength{\tabcolsep}{1.7pt}
\renewcommand{\arraystretch}{1.15}
\caption{\textbf{Statistics and key properties of egocentric and exocentric datasets.} Modalities include only Semantic, Gaze, and Interaction annotations reported as part of each dataset; Interaction includes contact, force, tactile, and hand-object annotations. ``Cam. Calib.'', ``Hand Pose'', and ``3D'' denote provided camera calibration parameters, articulated frame-level hand poses, and metric 3D supervision, respectively.}
\vspace{-0.2cm}
\label{tab:ego-exo-datasets}
\rowcolors{2}{white}{w_blue!7}
\begin{tabular}{@{}
>{\raggedright\arraybackslash}m{0.20\textwidth}
>{\centering\arraybackslash}m{0.06\textwidth}
>{\centering\arraybackslash}m{0.08\textwidth}
>{\centering\arraybackslash}m{0.12\textwidth}
>{\centering\arraybackslash}m{0.21\textwidth}
>{\centering\arraybackslash}m{0.07\textwidth}
>{\centering\arraybackslash}m{0.07\textwidth}
>{\centering\arraybackslash}m{0.07\textwidth}
@{}}
\toprule
\textbf{Dataset} & \textbf{Year} & \textbf{View} & \textbf{Hours/Clips} & \textbf{Modality} & \textbf{Cam. Calib.} & \textbf{Hand Pose} & \textbf{3D} \\
\midrule\midrule
EgoHands~\cite{egohands2015} & 2015 & Ego & -/48 & Semantic & \ding{55} & \ding{55} & \ding{55} \\
KrishnaCam~\cite{krishnacam2016} & 2016 & Ego & 70.2/460 & - & \ding{55} & \ding{55} & \ding{55} \\
FPHA~\cite{fpha2017} & 2017 & Ego & -/1,175 & Semantic & \ding{51} & \ding{51} & \ding{51} \\
Charades-Ego~\cite{sigurdsson2018charadesego} & 2018 & Ego+Exo & 68.8/7,860 & Semantic & \ding{55} & \ding{55} & \ding{55} \\
EGTEA Gaze+~\cite{egteagaze2018} & 2018 & Ego & 28/86 & Semantic; Gaze & \ding{55} & \ding{55} & \ding{55} \\
EPIC-KITCHENS v1~\cite{Damen2018EPICKITCHENS} & 2018 & Ego & 55/432 & Semantic & \ding{55} & \ding{55} & \ding{55} \\
EPIC-KITCHENS-100~\cite{damen2020rescaling} & 2020 & Ego & 100/700 & Semantic; Interaction & \ding{55} & \ding{55} & \ding{55} \\
Ego4D~\cite{grauman2022ego4d} & 2021 & Ego & 3,670/- & Semantic; Gaze; Interaction & \ding{55} & \ding{55} & \ding{51} \\
H2O~\cite{kwon2021h2o} & 2021 & Ego+Exo & -/- & Semantic; Interaction & \ding{51} & \ding{51} & \ding{51} \\
Assembly101~\cite{sener2022assembly101} & 2022 & Ego+Exo & 513/- & Semantic & \ding{51} & \ding{51} & \ding{51} \\
HOI4D~\cite{liu2022hoi4d} & 2022 & Ego & -/4,000 & Semantic; Interaction & \ding{51} & \ding{51} & \ding{51} \\
ARCTIC~\cite{fan2023arctic} & 2023 & Ego+Exo & -/339 & Interaction & \ding{51} & \ding{51} & \ding{51} \\
AssemblyHands~\cite{ohkawa2023assemblyhands} & 2023 & Ego+Exo & -/- & - & \ding{51} & \ding{51} & \ding{51} \\
CaptainCook4D~\cite{peddi2024captaincook4d} & 2023 & Ego & 94.5/384 & Semantic; Interaction & \ding{55} & \ding{55} & \ding{51} \\
Ego-Exo4D~\cite{grauman2024egoexo4d} & 2023 & Ego+Exo & 1,286/5,035 & Semantic; Gaze & \ding{51} & \ding{51} & \ding{51} \\
ENIGMA-51~\cite{enigma512023} & 2023 & Ego & 22/51 & Semantic; Interaction & \ding{55} & \ding{51} & \ding{55} \\
HoloAssist~\cite{wang2023holoassist} & 2023 & Ego & 166/2,221 & Semantic; Gaze & \ding{51} & \ding{51} & \ding{51} \\
EgoExo-Fitness~\cite{li2024egoexofitness} & 2024 & Ego+Exo & 32/1,276 & Semantic & \ding{55} & \ding{55} & \ding{55} \\
EgoExoLearn~\cite{huang2024egoexolearn} & 2024 & Ego+Exo & 120/747 & Semantic; Gaze & \ding{55} & \ding{55} & \ding{55} \\
EgoSurgery~\cite{egosurgeryphasetoolhts2024} & 2024 & Ego & 15/21 & Semantic; Gaze & \ding{55} & \ding{55} & \ding{55} \\
HOT3D~\cite{banerjee2024hot3d} & 2024 & Ego & 13.9/425 & Gaze; Interaction & \ding{51} & \ding{51} & \ding{51} \\
IndustReal~\cite{industreal2024} & 2024 & Ego & 5.8/84 & Semantic; Gaze & \ding{55} & \ding{51} & \ding{51} \\
EgoDex~\cite{hoque2026egodex} & 2025 & Ego & 829/338,000 & Semantic & \ding{51} & \ding{51} & \ding{51} \\
Egocentric-100K~\cite{buildai100k2025} & 2025 & Ego & 100,405/2.01M & - & \ding{51} & \ding{55} & \ding{55} \\
EgoLife~\cite{egolife2025} & 2025 & Ego+Exo & 266/- & Semantic; Gaze & \ding{55} & \ding{55} & \ding{51} \\
EgoMe~\cite{qiu2025egome} & 2025 & Ego+Exo & 82.8/15,804 & Semantic; Gaze & \ding{55} & \ding{55} & \ding{55} \\
HD-EPIC~\cite{perrett2025hdepic} & 2025 & Ego & 41.3/156 & Semantic; Gaze; Interaction & \ding{51} & \ding{51} & \ding{51} \\
IndEgo~\cite{chavan2025indegod} & 2025 & Ego+Exo & 293.9/- & Semantic; Gaze & \ding{51} & \ding{51} & \ding{51} \\
OpenEgo~\cite{jawaid2025openego} & 2025 & Ego & 1,107/- & Semantic; Interaction & \ding{51} & \ding{51} & \ding{51} \\
TASTE-Rob~\cite{zhao2025tasterob} & 2025 & Ego & -/100,856 & Semantic; Interaction & \ding{55} & \ding{55} & \ding{55} \\
ChildLens~\cite{childlens2026} & 2026 & Ego & 108.58/354 & Semantic & \ding{55} & \ding{55} & \ding{55} \\
Ego-1K~\cite{ego1k2026} & 2026 & Ego & -/956 & - & \ding{51} & \ding{55} & \ding{55} \\
EgoLive~\cite{li2026egolive} & 2026 & Ego & 1,680/- & Semantic; Interaction & \ding{51} & \ding{51} & \ding{51} \\
EgoMAGIC~\cite{egomagic2026} & 2026 & Ego & -/3,355 & Semantic; Interaction & \ding{55} & \ding{55} & \ding{55} \\
EgoVerse~\cite{punamiya2026egoverse} & 2026 & Ego & 1,362/- & Semantic; Interaction & \ding{55} & \ding{51} & \ding{51} \\
HumanNet~\cite{deng2026humannet} & 2026 & Ego+Exo & 967,000/- & Semantic & \ding{55} & \ding{51} & \ding{51} \\
Open-AoE~\cite{li2026openaoeopenegocentricmanipulation} & 2026 & Ego & 2,000/- & Semantic; Interaction & \ding{51} & \ding{51} & \ding{51} \\
Xperience-10M~\cite{xperience_10m} & 2026 & Ego & 10,000/- & Semantic; Interaction & \ding{51} & \ding{51} & \ding{51} \\
\bottomrule
\end{tabular}%
\end{table}

\subsection{Capture Infrastructure}

Egocentric and exocentric capture systems combine wearable and environmental devices to measure appearance, geometry, motion, attention, and physical interaction. We organize the hardware by the physical quantity it observes: visual sensors record image and depth streams, motion-tracking systems measure device or body kinematics, and auxiliary sensors expose attention, muscle activation, and contact. Across these categories, the main trade-offs concern spatial coverage, temporal resolution, mobility, intrusiveness, accuracy, and calibration requirements.

\subsubsection{Visual Sensing}

\noindent\textbf{Wearable Cameras.}
Monocular RGB capture uses a single color camera, typically mounted on the head or chest, to approximate the wearer's visual stream. Such cameras are lightweight, inexpensive, and suitable for long-duration collection, as demonstrated by EPIC-KITCHENS~\cite{Damen2018EPICKITCHENS,damen2020rescaling} and Ego4D~\cite{grauman2022ego4d}; however, they do not measure metric depth directly, and rapid head motion can introduce blur while the hands can occlude manipulated objects.

Stereo rigs instead use two synchronized cameras with a known baseline to estimate depth from binocular disparity. They avoid active illumination but require accurate calibration and can fail in textureless or occluded regions. HOT3D~\cite{banerjee2024hot3d}, for example, uses the onboard multi-camera configurations of Project Aria and Quest~3 to observe hands and objects from slightly different viewpoints. RGB-D cameras pair color images with per-pixel depth, providing more direct 3D measurements but with device-dependent range and sensitivity to reflective surfaces or strong ambient light. H2O~\cite{kwon2021h2o} synchronizes egocentric and exocentric RGB-D views, while HoloAssist~\cite{wang2023holoassist} records RGB and depth with a HoloLens~2.

Event cameras take a different approach: rather than producing conventional frames at a fixed rate, each pixel asynchronously reports changes in brightness. Their high temporal resolution reduces motion blur during fast hand motion, but produces sparse data that requires specialized processing; EgoEV-HandPose~\cite{egoevhands2026} explores this setting with stereo event cameras.

\noindent\textbf{Exocentric Camera Systems.}
Exocentric systems place one or more cameras in the environment to observe the wearer from fixed third-person viewpoints. They reveal body, hand, and scene context that may be occluded or outside the wearable camera's field of view, and synchronized views support triangulation and multi-view reconstruction. Assembly101~\cite{sener2022assembly101} combines wearable and external cameras for fine-grained assembly activities, while Ego-Exo4D~\cite{grauman2024egoexo4d} pairs Project Aria glasses with multiple surrounding GoPro cameras. EgoExOR~\cite{ozsoy2025egoexor} extends synchronized ego-exo capture to two emulated spine procedures, combining wearable RGB, gaze, hand tracking, and audio with exocentric RGB-D and ultrasound for surgical scene-graph understanding. ToolDex~\cite{elmoghany2026tooldex} records 10.8K human tool-use trials from a synchronized egocentric view and eight exocentric cameras, with separate annotations for the tool and the target object. ACE-Data-0~\cite{cao2026acedata0} instead calibrates entire real homes as recording studios, using a table-scale rig for hand-object manipulation and a room-scale rig for locomotion and whole-body activity, which extends exocentric coverage from a single workspace to a furnished multi-room environment. The additional coverage comes at the cost of environmental setup, cross-camera calibration and synchronization, higher storage demand, and reduced collection mobility.

\subsubsection{Motion and Pose Tracking}

\noindent\textbf{Inertial Body Tracking.}
An inertial measurement unit (IMU) combines accelerometers and gyroscopes, and sometimes a magnetometer, to measure linear acceleration and angular velocity. IMUs are compact and remain available under visual occlusion, but orientation and especially position estimates accumulate drift when the measurements are integrated over time. Ego-Exo4D~\cite{grauman2024egoexo4d} records the IMUs built into Project Aria, while WEAR~\cite{wear2023} distributes body-worn IMUs to capture outdoor activity dynamics. Full-body inertial motion-capture systems extend this idea by placing multiple synchronized IMUs across the body and fitting their measurements to an articulated skeleton. Nymeria~\cite{ma2024nymeria} combines Project Aria and wrist-worn miniAria devices with an Xsens suit to capture head, hand, and body motion without an external camera array, although the suit increases fitting and body-specific calibration effort.

\noindent\textbf{Hand-Worn Tracking Devices.}
Hand-worn trackers place magnetic, inertial, flex, or optical sensors on the fingers, hand, or wrist to obtain pose-related measurements such as sensor orientation, finger bending, fingertip position, and wrist motion. These measurements are converted into articulated hand pose through device calibration, sensor fusion, or kinematic fitting; they should therefore be regarded as sensor-based pose capture rather than unprocessed pose labels. FPHA~\cite{fpha2017} attaches six magnetic sensors to the fingertips and recovers articulated hand motion even when the fingers are visually occluded, although nearby metal and electromagnetic interference can distort the measurements. DexGloveHOI~\cite{dexglovehoi2026} instead synchronizes egocentric images with glove-mounted 6-DoF IMUs and motion-capture reference poses for vision-inertial 3D hand tracking. Compared with image-only estimation, instrumented wearables generally provide more continuous and accurate 3D hand motion under severe hand-object occlusion. Their scalability is nevertheless limited by hand-shape calibration, sensor displacement, magnetic interference or inertial drift, and possible interference with natural grasping.

\noindent\textbf{Optical and Headset-Based Tracking.}
Optical motion capture uses multiple external cameras to triangulate visible markers, yielding accurate 3D trajectories in a calibrated workspace. HOT3D~\cite{banerjee2024hot3d} places small markers on the hands and objects to obtain pose ground truth, trading accuracy for controlled setup and possible marker occlusion. RoboMoCap~\cite{deshmukh2026robomocap} pairs egocentric vision with 120~Hz optical whole-body motion capture to supervise locomotion and loco-manipulation, including movements beyond the wearable cameras' field of view. AR/VR headsets provide a more portable alternative through device-native inside-out tracking, which fuses onboard cameras and inertial sensing to estimate the headset and hands. EgoDex~\cite{hoque2026egodex} uses Apple Vision Pro tracking, and HoloAssist~\cite{wang2023holoassist} uses HoloLens~2. These systems reduce external infrastructure but depend on proprietary tracking pipelines and can fail under fast motion, occlusion, or limited field of view.

\subsubsection{Auxiliary and Interaction Sensing}

\noindent\textbf{Gaze and Audio.}
Eye trackers use inward-facing cameras and illumination to estimate gaze direction, providing a cue to visual attention but requiring wearer-specific calibration and stable headset placement. Microphones record speech, environmental sounds, and acoustic contact events; multi-channel arrays additionally capture spatial information and support more robust speech recording, but remain sensitive to background noise and privacy constraints. Ego-Exo4D~\cite{grauman2024egoexo4d} records eye gaze and seven-channel audio with Project Aria, while HoloAssist~\cite{wang2023holoassist} synchronizes gaze and audio with its visual and tracking streams.

\noindent\textbf{Electromyography.}
Electromyography (EMG) uses skin-contact electrodes to measure the electrical activity produced by muscle activation rather than hand pose itself. EgoEMG~\cite{egoemg2026} records bilateral wrist EMG and IMU at higher rates than video, exposing motor activity even when the hands are visually occluded. The signal is informative about muscle recruitment, but depends strongly on electrode placement, skin contact, participant physiology, and session-specific calibration.

\noindent\textbf{Force and Tactile Sensing.}
Force and tactile devices measure physical interaction more directly. A force sensor reports the magnitude or direction of load at one or several locations, whereas a tactile or pressure array resolves how contact is distributed over an area. FEEL~\cite{feel2026} places piezoresistive force sensors on the fingertips and palm, Touch and Go~\cite{touchandgo2022} records visual probing with a tactile sensor, and EgoTouch~\cite{egotouch2026} combines continuous wearable pressure measurements with head- and wrist-mounted RGB and bimanual hand pose. These wearables provide contact signals unavailable from vision alone, but limited sensing area, sensor fitting, per-user calibration, cabling, and interference with natural grasping can restrict collection scale and fidelity.

\subsection{Supervision Construction}

Egocentric supervision may be recorded by sensors, added by human annotators, or recovered through post-processing. We group it into four complementary categories according to the information exposed to learning: semantic supervision describes task content and structure; geometric supervision describes spatial state and motion; multimodal supervision supplies attention, dynamics, and contact cues beyond RGB; and robot-oriented supervision expresses human demonstrations in forms intended for robot policy learning. Their construction combines annotation, calibration, synchronization, reconstruction, and cross-embodiment alignment, with different costs and sources of error.

\subsubsection{Semantic Supervision}

\noindent\textbf{Action and Language Annotations.}
Action and language annotations identify what activity occurs and when it occurs. Common forms include narrations, object references, verb-noun actions, action classes, and temporal segments. Participant narrations are transcribed and aligned with video, while action classes and boundaries are assigned manually or with model assistance followed by human verification. EPIC-KITCHENS~\cite{Damen2018EPICKITCHENS,damen2020rescaling}, for example, turns free-form participant narrations into temporally localized verb-noun actions. Ego4D~\cite{grauman2022ego4d} combines large-scale narrations with annotations tailored to individual benchmarks, and Assembly101~\cite{sener2022assembly101} provides fine-grained labels for assembly actions. These annotations support recognition, retrieval, and temporal grounding, but do not specify the continuous motion or contact needed to perform an action.

\noindent\textbf{Procedural and State Annotations.}
Procedural annotations describe how individual actions compose into a task. They include ordered steps, task progress, execution errors, and object or environment state changes, and are typically constructed by aligning videos with task scripts and verifying deviations from the expected procedure. Assembly101~\cite{sener2022assembly101} organizes fine-grained actions within assembly procedures, whereas CaptainCook4D~\cite{peddi2024captaincook4d} labels cooking steps together with errors and object states. This supervision exposes long-horizon dependencies and task outcomes that isolated action labels miss, although its vocabulary is often specific to a particular procedure or domain.

\subsubsection{Geometric Supervision}

\noindent\textbf{Camera and Scene Geometry.}
Camera and scene supervision establishes the spatial reference in which an interaction occurs. It includes camera intrinsics and extrinsics, depth maps, point clouds, camera poses and trajectories, scene reconstructions, and point tracks. Depth may be measured directly, as in the RGB-D observations of H2O~\cite{kwon2021h2o}; other quantities require calibration, visual-inertial SLAM, multi-view reconstruction, or tracking. HOT3D~\cite{banerjee2024hot3d} provides calibrated camera poses, while EgoDex~\cite{hoque2026egodex} combines headset tracking with SLAM to recover the moving egocentric camera trajectory. Synchronized egocentric and external views, such as those in Ego-Exo4D~\cite{grauman2024egoexo4d}, further support reconstruction beyond the wearable camera's field of view. The resulting geometry is sensitive to calibration drift, motion blur, and the choice of coordinate frame.

\noindent\textbf{Hand and Body Pose.}
Human pose supervision is released at several levels of detail. Hands may be encoded as 2D or 3D keypoints, wrist and fingertip poses, joint rotations and trajectories, or MANO pose and shape parameters, while body supervision ranges from head and upper-body trajectories to full articulated skeletons or parametric body models. Because these representations describe the output rather than how it was obtained, pose labels can be organized by three main construction routes.

\emph{(1) Annotation-assisted Geometric Reconstruction.}
Annotators mark 2D hand or body keypoints in one or more RGB views. With registered depth, visible keypoints can be back-projected into 3D; with synchronized and calibrated cameras, corresponding keypoints can instead be triangulated across views. AssemblyHands~\cite{ohkawa2023assemblyhands} illustrates a scalable variant: it starts from a manually annotated subset and trains a multi-view annotation model to extend accurate 3D hand poses to a much larger collection. This route retains direct human quality control, but manual labeling is costly and its 3D accuracy depends on depth quality, cross-view correspondence, and camera calibration.

\emph{(2) Model-based Pose Prediction and Parametric Fitting.}
Learned vision models generate pose labels directly from recorded RGB or RGB-D observations. Keypoint detectors predict 2D or 3D joints, while model-recovery networks may regress MANO pose and shape parameters or analogous body-model parameters. These initial predictions can be used as pseudo-labels or further refined with depth, silhouettes, multi-view consistency, temporal constraints, and parametric-model fitting. H2O~\cite{kwon2021h2o}, for example, uses OpenPose predictions to initialize its hand annotations and then optimizes a MANO hand model with synchronized multi-view RGB-D observations. This route scales more readily to unlabeled recordings, but the resulting supervision inherits model errors and biases caused by occlusion, motion blur, and domain shift. Occlusion and out-of-frame gaps are the dominant failure mode for egocentric hands, and recent work addresses them by widening the temporal context rather than improving per-frame regression: DreamHand~\cite{liu2026dreamhand} repurposes a video diffusion model as a deterministic clip-level geometry encoder, reading occluded and out-of-sight hands from a single forward pass over the clean latent and decoding continuous bimanual trajectories with metric placement and no external detector.

Ego-Exo4D-HM~\cite{maddukuri2026egoexo4dhm} enriches synchronized ego-exo recordings with dense body and hand meshes, global motion, calibrated cameras, and validity masks. Multi-view triangulation and metric-frame optimization yield 104.6 hours of reconstructed motion across 2,649 takes; the larger video-hour count reflects multiple camera streams. The released annotations provide geometric supervision, although they remain reconstruction-derived pseudo-labels and the pipeline requires calibrated external views. The paper does not evaluate downstream robot-policy learning.

\emph{(3) Sensor-based Pose Capture.}
The aforementioned hand-worn, optical, headset-based, and inertial systems provide pose-related measurements independently of image-only inference. Their signals are converted into articulated pose through calibration, sensor fusion, or kinematic fitting. FPHA~\cite{fpha2017} uses magnetic fingertip sensors, HOT3D~\cite{banerjee2024hot3d} converts professional optical motion capture into UmeTrack and MANO hand annotations, EgoDex~\cite{hoque2026egodex} obtains upper-body and hand trajectories from headset tracking, and Nymeria~\cite{ma2024nymeria} combines wearable devices with an inertial suit for full-body motion. Instrumented capture generally yields more continuous and accurate 3D motion under visual occlusion, but increases calibration effort and participant burden and remains dependent on device-specific tracking models. Across all three routes, pose supervision differs in joint definitions, coordinate frames, temporal stability, and confidence, which should be reported explicitly when datasets are compared or combined.

\noindent\textbf{Object and Interaction Geometry.}
Object-centered supervision describes how manipulated objects move relative to the hands and scene. Typical forms include object 6-DoF pose, object trajectory, hand-object relative pose, and coupled hand-object trajectories. They are obtained through CAD-model registration, multi-view reconstruction, marker-based motion capture, pose estimation, or temporal tracking. H2O~\cite{kwon2021h2o} jointly provides hand and object poses in RGB-D sequences, HOI4D~\cite{liu2022hoi4d} annotates the spatial evolution of hand-object interactions, and HOT3D~\cite{banerjee2024hot3d} releases calibrated hand and object poses. A recurring gap is linguistic rather than geometric: object-motion supervision is typically paired with coarse verb-noun labels, which underdetermine how an object should move. The MOVE dataset introduced with DreamTraj~\cite{ding2026dreamtraj} addresses this by releasing 5,038 object-centric egocentric trajectories, each paired with a fine-grained natural-language instruction rather than a category label, so that 6-DoF object motion can be supervised at instruction-level granularity. Unlike semantic object labels, these annotations expose manipulation geometry directly, although occlusion and object symmetry can make pose recovery ambiguous.

\subsubsection{Multimodal Supervision}

\noindent\textbf{Gaze and Audio.}
Gaze and audio associate visual observations with attention and acoustic context. Raw gaze measurements must be calibrated and projected into fixation points, attended regions, or gaze trajectories, whereas audio must be synchronized with video and optionally aligned with actions or state changes. EGTEA Gaze+~\cite{egteagaze2018} provides gaze aligned with kitchen activities, and Ego-Exo4D~\cite{grauman2024egoexo4d} records synchronized gaze and multi-channel audio. HoloAssist~\cite{wang2023holoassist} similarly pairs gaze and audio with instructional activities. These cues help localize relevant objects and events, but gaze calibration errors and ambient sound can weaken their semantic correspondence.

\noindent\textbf{Inertial and Physiological Signals.}
Inertial and physiological supervision captures motion dynamics and muscle activity that are not explicit in images. IMU streams provide acceleration and angular velocity, while EMG measures muscle activation; their use requires synchronization, filtering, normalization, and segmentation at a sampling rate often much higher than video. WEAR~\cite{wear2023} combines egocentric video with body-worn IMUs for outdoor activities, Ego-Exo4D~\cite{grauman2024egoexo4d} includes Project Aria inertial measurements, and EgoEMG~\cite{egoemg2026} pairs bilateral EMG and IMU streams with hand motion. These signals remain available under visual occlusion, but are sensitive to sensor placement, drift, and participant physiology.

Whole-body capture extends these signals to coordination between locomotion and manipulation. EgoExplorer~\cite{zhou2026egoexplorer} aligns video, tactile observations, hand articulation, and reconstructed body motion over a reported 100-hour collection; sparse-sensor reconstruction is not full-body motion-capture ground truth, and real-world policy-learning benefits remain unverified. HumanVerse~\cite{lambdateam2026lambda0} reports 500 hours of synchronized egocentric video and body/hand motion, with offline viewpoint and controller alignment for humanoid learning. Release scope remains distinct: EgoExplorer currently offers a one-hour gated preview, while the full HumanVerse mid-training corpus is absent from the checked code release.

\noindent\textbf{Force and Tactile Signals.}
Force and tactile supervision provides direct evidence of physical contact. Depending on the sensor, it may be represented as scalar or vector force, binary contact events, fingertip and palm pressure, or spatial tactile maps. FEEL~\cite{feel2026} synchronizes fingertip and palm force measurements with egocentric observations, Touch and Go~\cite{touchandgo2022} records visual exploration together with tactile sensing, and EgoTouch~\cite{egotouch2026} pairs wearable tactile pressure with head- and wrist-mounted RGB and bimanual hand pose. Such measurements distinguish visually similar motions with different contact outcomes, although wearable sensing does not necessarily reproduce a robot's contact geometry or force response. Because tactile instrumentation is precisely what limits collection scale, a complementary strategy is to predict contact from vision instead of measuring it. EgoTac~\cite{zhang2026egotac} trains on a unified corpus of over 5.7M image-tactile pairs spanning continuous force measurements and binary contacts, and infers tactile signals directly from in-the-wild egocentric video, reporting in-domain average force error below 0.06\,N. Such predicted tactile channels can retrofit contact supervision onto existing uninstrumented egocentric corpora, though they remain estimates rather than measurements and inherit the sensor conventions of their training data.

React, introduced with ACTS~\cite{ma2026acts}, records 4.9 hours of human-operated bimanual interaction with three objects using handheld tactile sensors, wrist and third-person video, tracked sensor motion, and tactile-derived normal force. Sensor-centered motion and force change provide a shared action representation for human collection and robot teleoperation. The force channel remains an estimate rather than measured force ground truth throughout the human recordings. Robot transfer tests concern future-observation prediction, with mixed visual transfer, rather than autonomous manipulation success; code and data are announced for release upon acceptance.

\subsubsection{Robot-Oriented Supervision}

\noindent\textbf{Robot-Compatible Action Supervision.}
Robot-compatible supervision maps human motion into a target action space, such as an end-effector pose, robot joint configuration, gripper command, or retargeted dexterous-hand action. It is constructed from tracked wrist, fingertip, or MANO motion through coordinate transformation, inverse kinematics, morphology-aware retargeting, and feasibility filtering. EgoVLA~\cite{yang2025egovla}, for example, predicts wrist pose and MANO parameters and converts them to robot commands through inverse kinematics and hand retargeting. EgoScale~\cite{zheng2026egoscale} supervises human pretraining with relative wrist motion and retargeted high-DoF hand actions, while MotionTrans~\cite{yuan2025motiontrans} transforms human VR motion for robot policy learning. These targets are often generated by a method rather than released as native annotations of the source dataset, and geometric plausibility does not guarantee dynamically feasible or contact-consistent execution. Two further strategies loosen the requirement that supervision be a robot action at all. The first targets an embodiment-agnostic intermediate: VLAff~\cite{oh2026vlaff} builds EgoAffordance, comprising 204K episodes with 5.6M visual affordances and 11.6M grasp and trajectory affordances, encoding where to interact, how to grasp, and how to move rather than which joint values to command. The second generates the demonstration before retargeting it: RoboReact~\cite{he2026roboreact} synthesizes human manipulation video from a single egocentric RGB-D observation, extracts geometry-preserving interaction keyframes through depth-aware 3D reconstruction, and retargets them to high-DoF humanoid platforms while preserving hand-object interaction geometry, while Pegasus~\cite{luo2026passive} routes passive video through task, affordance, and constraint graphs into a robot planning graph and rejects generations that fail kinematic-feasibility and collision checks. Because both introduce a generative step, the resulting supervision is only as physically meaningful as the verifier that filters it.

HuRo~\cite{jeong2026huro} jointly aligns visual observations and action trajectories from heterogeneous human videos with a target robot embodiment by estimating camera and hand geometry, segmenting manipulation, retargeting motion, and rendering the robot into the video. Its processed corpus spans about 630,000 episodes from five sources, and downstream experiments evaluate stage-wise completion on one robot embodiment. Such estimated supervision is not automatically physically executable: contact, joint-limit, and occlusion errors remain. The conversion code is available, while the full derived corpus is still pending.

EgoMatrix~\cite{wang2026egomatrix} adds contact-aware hand-object reconstruction, simulation-based retargeting and policy refinement, and rendering of robot observations synchronized with state and control trajectories. Its reconstructed geometry prioritizes manipulation-relative contact over exact metric scene recovery. The reported end-to-end success evaluation is in simulation, supplemented by real-robot trajectory replay; this does not establish closed-loop policy generalization across arbitrary physical tasks. Background transfer reuses compatible executed trajectories rather than re-simulating their contact dynamics in each target scene, and monocular depth and occlusion ambiguities remain.

\noindent\textbf{Human-Robot Alignment Supervision.}
Human-robot alignment supervision relates demonstrations across embodiments without requiring identical raw action spaces. It may take the form of task- or scene-matched human and robot demonstrations, temporally aligned trajectories, or a shared representation to which both embodiments are mapped. EgoMimic~\cite{kareer2025egomimic} normalizes human hand and robot demonstrations for joint policy training, while EgoScale~\cite{zheng2026egoscale} uses aligned human-robot play in matched tasks and scenes as an intermediate training stage. H-RDT~\cite{bi2025hrdt} instead combines human hand-pose pretraining with robot-specific action adapters, and METIS~\cite{fu2025metis} integrates multiple human and robot sources under a consistent action space. These examples provide different strengths of correspondence rather than uniform frame-level pairs, so the type and precision of alignment should be stated explicitly. Two orthogonal questions follow from this. The first is which human data to align at all: SiMDex~\cite{lin2026simdex} casts human-data selection for dexterous VLA post-training as a recommendation problem, mining task-relevant subsets from a pool of roughly 32M egocentric samples through a recall-ranking-re-ranking pipeline in a morphology-agnostic action space, and reports that about 1.49M mined samples raise success from 47.7\% to 61.1\% relative to an equal-size random draw, indicating that indiscriminate mixing wastes much of the pool. The second is how to close the residual appearance gap: HandEdit~\cite{yang2026handedit} provides an embodiment-aware image-editing dataset and benchmark of over 200M editing instances derived from five source datasets and covering 26 URDFs, for transforming human hands and arms into specific robotic embodiments within egocentric frames. Selection and visual translation are complementary to retargeting rather than substitutes for it, since neither supplies contact dynamics.

EgoWild~\cite{lin2026egowild2dex} collects 538.9 hours of unscripted egocentric video, tracked bimanual motion, and structured language annotations across diverse everyday sites. EgoWild2Dex combines projective view alignment and motion retargeting with human pretraining, task-specific human--robot co-training, and robot-domain refinement. Physical tests cover three dexterous tasks and still use task-specific robot demonstrations and corrective data. The result supports a human-data engineering pipeline, while camera field-of-view limits and pending code and data release constrain its current reuse.

\subsection{Advantages and Limitations}

Egocentric and exocentric data capture natural human demonstrations at scale without requiring execution on a target robot. Wearable systems can be deployed across different participants, environments, objects, and tasks, while first-person observations emphasize hand-object interaction and long-horizon activity structure. Unlike fixed-base robot collection, human operators can move freely through the environment and manipulate objects directly with dexterous hands. This mobility enables broader spatial coverage and more diverse hand trajectories, including variations in approach direction, motion path, and interaction height. The EgoVerse example in Figure~\ref{fig:diversity} illustrates this potential: its human-hand interaction keyframes span a broad three-dimensional workspace and exhibit substantial spatial variation.

However, egocentric data are not equivalent to robot experience. First-person recordings suffer from partial observability, hand-object occlusion, rapid camera motion, motion blur, and device-dependent viewpoints. Dense geometric and multimodal supervision requires additional hardware, calibration, synchronization, and reconstruction, and is not uniformly available at scale. More fundamentally, human recordings lack robot proprioception, actuator dynamics, and directly executable actions. Differences in morphology, kinematics, sensing, and contact response further prevent direct trajectory transfer. Their use in robot learning therefore requires reliable state reconstruction, embodiment-aware representations, and human-to-robot alignment or retargeting.

\begin{wbtakeaway}
\bEgo~\textbf{Egocentric and exocentric data} occupy the middle layer of the embodied data pyramid, providing a scalable route to collecting physically grounded interaction trajectories in diverse in-the-wild environments. By recording human agents interacting with the physical world, these data capture natural manipulation behaviors and high-degree-of-freedom hand motions with substantial potential for reuse across robotic embodiments. Their collection, however, involves a trade-off between scalability and annotation fidelity. Device-free recordings are unobtrusive and easy to scale, but typically lack accurate hand-pose and interaction labels, requiring downstream pipelines to rely on model-based estimation or learned visual retargeting. Wearable sensing systems can provide more direct and precise measurements, but remain constrained by hardware cost, durability, user burden, and portability, particularly for large-scale outdoor collection. Beyond this acquisition trade-off, their utility for robot learning is further limited by unresolved alignment between human and robot embodiments.
\end{wbtakeaway}

\section{Simulation Data}
\begin{table}[!t]
\centering
\scriptsize
\setlength{\tabcolsep}{1.8pt}
\renewcommand{\arraystretch}{1.15}
\caption{\textbf{Simulation Benchmarks.} Simulation benchmarks, summarized by embodiment configuration, tactile sensing, object deformability, dexterous manipulation, and mobile manipulation. In the ``Embod.'' column, `S', `D', and `H' denote single-arm, dual-arm, and humanoid embodiments, respectively.}
\vspace{-0.2cm}
\label{tab:simulation-benchmarks}
\rowcolors{2}{white}{w_blue!7}
\begin{tabular}{@{}
>{\raggedright\arraybackslash}m{0.21\textwidth}
>{\centering\arraybackslash}m{0.06\textwidth}
>{\centering\arraybackslash}m{0.113\textwidth}
>{\centering\arraybackslash}m{0.05\textwidth}
>{\centering\arraybackslash}m{0.113\textwidth}
>{\centering\arraybackslash}m{0.05\textwidth}
>{\centering\arraybackslash}m{0.05\textwidth}
@{}}
\toprule
\textbf{Benchmark} & \textbf{Time} & \textbf{Embod.} & \textbf{Tactile} &
\textbf{Deform.} & \textbf{Dex.} & \textbf{Mobile} 
\\
\midrule\midrule
Adroit~\cite{adroitdapg2018}
& 2017 & S & \ding{55} & Rigid & \ding{51} & \ding{55} \\
Franka Kitchen~\cite{frankakitchenrpl2019}
& 2019 & S & \ding{55} & Rigid & \ding{55} & \ding{55} \\
Meta-World~\cite{metaworld2021}
& 2019 & S & \ding{55} & Rigid & \ding{55} & \ding{55} \\
RLBench~\cite{rlbench2019}
& 2019 & S & \ding{55} & Rigid & \ding{55} & \ding{55} \\
PlasticineLab~\cite{plasticinelab2021}
& 2021 & S & \ding{55} & Soft & \ding{55} & \ding{55} \\
RoboMimic~\cite{robomimic2021}
& 2021 & S/D & \ding{55} & Rigid & \ding{55} & \ding{55} \\
CALVIN~\cite{calvin2022}
& 2021 & S & \ding{55} & Rigid & \ding{55} & \ding{55} \\
Tactile Gym 2.0~\cite{tactilegym202022}
& 2022 & S & \ding{51} & Rigid & \ding{55} & \ding{55} \\
Bi-DexHands~\cite{bidexhands2022}
& 2022 & D & \ding{55} & Rigid & \ding{51} & \ding{55} \\
BEHAVIOR-1K~\cite{behavior1k2024}
& 2022 & S & \ding{55} & Soft & \ding{55} & \ding{51} \\
VIMA-Bench~\cite{vimabench2023}
& 2022 & S & \ding{55} & Rigid & \ding{55} & \ding{55} \\
DaXBench~\cite{daxbench2023}
& 2023 & S & \ding{55} & Soft & \ding{55} & \ding{55} \\
DexArt~\cite{dexart2023}
& 2023 & S & \ding{55} & Rigid & \ding{51} & \ding{55} \\
RoboHive~\cite{robohive2023}
& 2023 & S & \ding{55} & Rigid & \ding{51} & \ding{55} \\
ARNOLD~\cite{arnold2023}
& 2023 & S & \ding{55} & Rigid & \ding{55} & \ding{55} \\
FurnitureBench~\cite{heo2025furniturebench}
& 2023 & S & \ding{55} & Rigid & \ding{55} & \ding{55} \\
ManiSkill2~\cite{maniskill22023}
& 2023 & S/D & \ding{55} & Soft & \ding{55} & \ding{51} \\
LIBERO~\cite{libero2023}
& 2023 & S & \ding{55} & Rigid & \ding{55} & \ding{55} \\
ManiSkill-HAB~\cite{maniskillhab2025}
& 2024 & S & \ding{55} & Rigid & \ding{55} & \ding{51} \\
TacSL~\cite{tacsl2025}
& 2024 & S & \ding{51} & Soft & \ding{55} & \ding{55} \\
ManiSkill-ViTac~\cite{maniskillvitac20252024}
& 2024 & S & \ding{51} & Soft & \ding{55} & \ding{55} \\
GarmentLab~\cite{garmentlab2024}
& 2024 & S/D & \ding{55} & Soft & \ding{51} & \ding{51} \\
HumanoidBench~\cite{humanoidbench2024}
& 2024 & H & \ding{55} & Rigid & \ding{51} & \ding{51} \\
BiGym~\cite{bigym2024}
& 2024 & D & \ding{55} & Rigid & \ding{55} & \ding{51} \\
The Colosseum~\cite{thecolosseum2024}
& 2024 & S & \ding{55} & Rigid & \ding{55} & \ding{55} \\
VLABench~\cite{vlabench2024}
& 2024 & S & \ding{55} & Rigid & \ding{55} & \ding{55} \\
SIMPLER-Env~\cite{simplerenv2024}
& 2024 & S & \ding{55} & Rigid & \ding{55} & \ding{55} \\
ManiSkill3~\cite{maniskill32025}
& 2024 & S/H & \ding{55} & Soft & \ding{51} & \ding{51} \\
RoboTwin~\cite{mu2024robotwin}
& 2024 & S/D & \ding{55} & Rigid & \ding{55} & \ding{55} \\
RoboCasa~\cite{robocasa2024}
& 2024 & S/H & \ding{55} & Rigid & \ding{55} & \ding{51} \\
MemoryBench~\cite{memorybench2025}
& 2025 & S & \ding{55} & Rigid & \ding{55} & \ding{55} \\
RoboCerebra~\cite{robocerebra2025}
& 2025 & S & \ding{55} & Rigid & \ding{55} & \ding{55} \\
GenManip~\cite{genmanip2025}
& 2025 & S & \ding{55} & Rigid & \ding{55} & \ding{55} \\
RoboTwin 2.0~\cite{chen2025robotwin}
& 2025 & D & \ding{55} & Rigid & \ding{55} & \ding{55} \\
RoboEval~\cite{roboeval2025}
& 2025 & D & \ding{55} & Rigid & \ding{55} & \ding{55} \\
REALM~\cite{realm2025}
& 2025 & S & \ding{55} & Rigid & \ding{55} & \ding{55} \\
RobotArena$\infty$~\cite{robotarena2026}
& 2025 & S & \ding{55} & Rigid & \ding{55} & \ding{55} \\
DuoBench~\cite{duobench2026}
& 2026 & D & \ding{55} & Rigid & \ding{55} & \ding{55} \\
BiCoord~\cite{bicoord2026}
& 2026 & D & \ding{55} & Rigid & \ding{55} & \ding{55} \\
MIKASA-Robo-VLA~\cite{mikasarobo2026}
& 2026 & S & \ding{55} & Rigid & \ding{55} & \ding{55} \\
RoboMemArena~\cite{robomemarena2026}
& 2026 & S & \ding{55} & Rigid & \ding{55} & \ding{55} \\
RoboMME~\cite{robomme2026}
& 2026 & S & \ding{55} & Rigid & \ding{55} & \ding{55} \\
RMBench~\cite{chen2026rmbench}
& 2026 & D & \ding{55} & Rigid & \ding{55} & \ding{55} \\
UniVTAC~\cite{chen2026univtac}
& 2026 & S & \ding{51} & Soft & \ding{55} & \ding{55} \\
RoboCasa365~\cite{robocasa3652026}
& 2026 & S/H & \ding{55} & Rigid & \ding{55} & \ding{51} \\
RoboDojo~\cite{chen2026robodojo}
& 2026 & D & \ding{55} & Rigid & \ding{55} & \ding{55} \\
\bottomrule
\end{tabular}%
\end{table}

\begin{figure}[t]
\centering
\includegraphics[width=\textwidth]{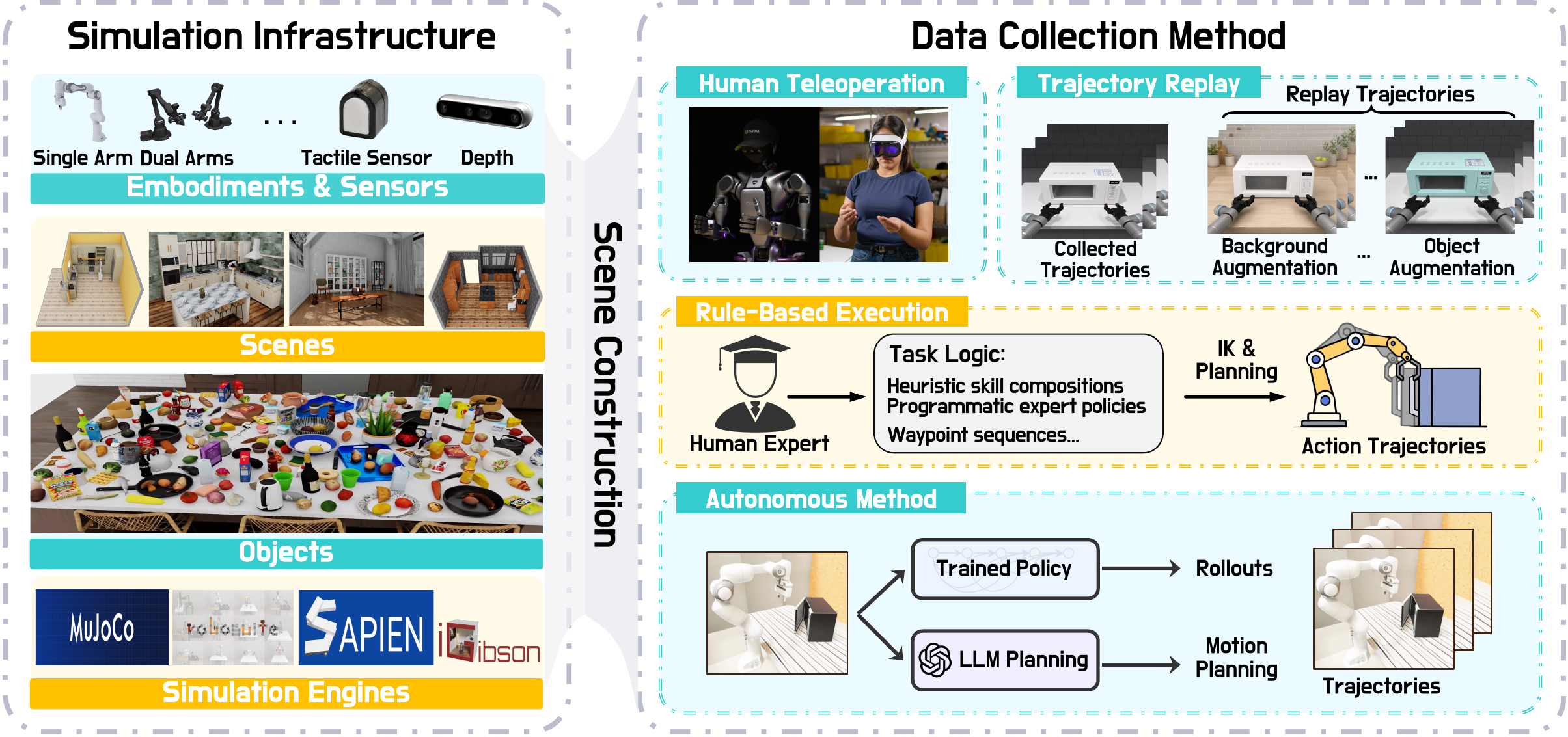}
\vspace{-0.6cm}
\caption{\textbf{Overview of simulation infrastructure and data collection pipeline.} The left panel summarizes the core components required for constructing simulation environments, including robotic embodiments and sensors, scenes, object assets, and simulation engines. The right panel categorizes representative trajectory collection methods into human teleoperation, trajectory replay with scene and object augmentation, rule-based execution based on expert-defined task logic and motion planning, and autonomous generation through trained policies or LLM-based planning.}
\label{fig:simulation_data}
\end{figure}

\subsection{Data Overview}
Within the embodied data pyramid, simulation data serves as a scalable and controllable complement to real-world robot data. While real-robot data is necessary for capturing embodiment-specific dynamics, sensor noise, contact effects, and deployment constraints, its collection is often limited by cost, safety, hardware availability, and annotation difficulty. Simulation alleviates part of this bottleneck by enabling repeatable interaction conditions, automatic access to privileged states and labels, and low-cost trajectory generation. Recent datasets further attempt to reduce the sim-to-real gap through digital twins~\cite{robomind2025,manitwin2026}, domain randomization~\cite{chen2025robotwin,mu2024robotwin,chen2026robodojo}, and paired sim-to-real evaluation~\cite{simplerenv2024}. As a result, simulation has become not only a tool for benchmark construction and evaluation, but also an important data layer for training general-purpose robot policies. Nevertheless, compared with the three higher layers of the embodied data pyramid, simulation data generally remains more limited in diversity and physical fidelity.

As summarized in Tables~\ref{tab:simulation-benchmarks} and~\ref{tab:sim-data}, simulated robot data has expanded in both scale and scope. Early benchmarks mainly focused on constrained rigid-body manipulation, such as table-top control~\cite{metaworld2021}, language-conditioned manipulation~\cite{rlbench2019}, and large-scale grasping~\cite{acronym2020}. Later datasets introduced longer-horizon interaction~\cite{calvin2022}, compositional task settings~\cite{libero2023}, and household manipulation~\cite{robocasa2024}. More recent resources further extend simulation toward bimanual manipulation~\cite{chen2025robotwin,mu2024robotwin,chen2026robodojo}, humanoid manipulation~\cite{bu2025agibot}, articulated-object interaction~\cite{maniskill22023}, deformable-object and fluid manipulation~\cite{interndataa12025,garmentlab2024,wang2025dexgarmentlab}, and contact-rich manipulation with tactile sensing~\cite{chen2026univtac}. In parallel, data generation has become increasingly automated, moving from scripted tasks and teleoperation toward pipelines based on motion planning and RL experts~\cite{maniskill32025}, demonstration augmentation~\cite{mimicgen2023}, digital-twin reconstruction~\cite{robomind2025,manitwin2026}, and LLM-assisted task or environment synthesis~\cite{interndataa12025,internvlam12025,gensim22024,robogen2024}.

\subsection{Simulation Infrastructure and Asset Ecosystems}

As shown in Figure~\ref{fig:simulation_data}, a simulated robot-learning ecosystem can be viewed as a composition of three coupled infrastructure components: embodiment-sensor systems, object and scene assets, and physics and rendering backends. The embodiment-sensor system defines not only the robot morphology, kinematics, actuation model, and controller interface, but also the sensing channels physically associated with the robot body, such as proprioception, wrist or head-mounted vision, force sensing, and tactile arrays. Object and scene assets provide the geometric, semantic, and affordance structure of the environment, while physics and rendering backends determine the contact dynamics, visual appearance, and execution fidelity of the generated data. 

\subsubsection{Robotic Embodiments and Sensors}
Constructing simulated manipulation data first requires importing an embodiment-sensor system into the simulator, where the robot asset is instantiated as a controllable articulated model and its sensors are configured to acquire observations during interaction. The simulator parses robot descriptions such as URDF, MJCF, SDF, or USD to specify kinematics, joint limits, inertial properties, collision and visual geometries, actuators, and sensor attachments~\cite{zhao2024aloha,internvlam12025,interndataa12025}, before placing the system in a scene with defined initial pose, joint state, objects, lighting, gravity, and solver settings.

Across simulated robotics datasets, stationary single-arm manipulators are the most common embodiment, as in Franka Kitchen~\cite{frankakitchenrpl2019}, Meta-World~\cite{metaworld2021}, RLBench~\cite{rlbench2019}, RoboMimic~\cite{robomimic2021}, CALVIN~\cite{calvin2022}, Simpler-Env~\cite{simplerenv2024}, RMBench~\cite{chen2026rmbench}, UniVTAC~\cite{chen2026univtac} and LIBERO~\cite{libero2023}. Other benchmarks cover dexterous hands, including Adroit~\cite{adroitdapg2018}, Bi-DexHands~\cite{bidexhands2022}, and DexArt~\cite{dexart2023}; mobile manipulation, including BEHAVIOR-1K~\cite{behavior1k2024}, ManiSkill2~\cite{maniskill22023}, ManiSkill-HAB~\cite{maniskillhab2025}, and RoboCasa~\cite{robocasa2024}; and dual-arm or bimanual systems, including RoboTwin~\cite{mu2024robotwin}, BiGym~\cite{bigym2024}, RMBench~\cite{chen2026rmbench}, DuoBench~\cite{duobench2026}, BiCoord~\cite{bicoord2026}, and RoboDojo~\cite{chen2026robodojo}.  ManiSkill3~\cite{maniskill32025}, RoboTwin 2.0~\cite{chen2025robotwin}, RoboMIND-Sim~\cite{robomind2025} and RoboMME~\cite{robomme2026} further support multiple embodiment classes, while HumanoidBench~\cite{humanoidbench2024} targets humanoid control.

\subsubsection{Object and Scene Assets}
The asset ecosystem not only includes robot models but also manipulable objects, materials, and scene layouts. Object-level assets determine the category coverage, geometric variation, texture realism, and affordance structure of simulated manipulation data. CAD-based and web-scale resources such as ShapeNet~\cite{shapenet2015}, Objaverse~\cite{objaverse2023}, and Objaverse-XL~\cite{objaversexl2023} provide broad object-shape diversity. Scanned or product-grounded resources such as Google Scanned Objects~\cite{googlescannedobjects2022}, Amazon Berkeley Objects~\cite{abo2022}, and OmniObject3D~\cite{omniobject3d2023} improve geometric and texture realism by grounding assets in real objects or product data.

However, robot-learning assets must be more than visually plausible meshes: they also require simulator-compatible collision geometry, consistent scale, stable poses, mass and inertia parameters, material properties, and, when applicable, grasp affordances or articulation annotations. PartNet-Mobility~\cite{partnetmobility2020} provides articulated objects with mobility annotations for part-level interaction. ObjectFolder 2.0~\cite{objectfolder202022} introduces multisensory object representations that connect visual, acoustic, and tactile properties. PhysXNet~\cite{physx} further introduces the first physics-aware 3D dataset, extending 3D representations beyond geometry and appearance to include physical properties. PhysX-Mobility~\cite{physxanything} then provides plug-and-play simulation-ready articulated assets, while PhysXVerse~\cite{physxomni} generalizes this paradigm to a broader and more diverse collection of simulation-ready 3D assets.

Beyond curated or reconstructed collections, generative modeling offers a scalable route to expanding the asset pool: ManiTwin~\cite{manitwin2026} contributes a large-scale library of over 100K AIGC-generated rigid-body objects, each equipped with manipulation-oriented annotations for automated data synthesis. This coupling is essential, because asset scaling alone is insufficient---without operable annotations, even abundant assets cannot be converted into usable manipulation data---which positions AIGC-driven generation as a promising direction for jointly scaling asset diversity and downstream data synthesis. At the scene level, ScanNet~\cite{scannet2017}, Replica~\cite{replica2019}, Matterport3D~\cite{matterport3d2017}, HM3D~\cite{hm3d2021}, HSSD~\cite{hssd2024}, 3D-FRONT~\cite{threedfront2021}, and ProcTHOR~\cite{procthor2022} provide reconstructed, synthetic, or procedurally generated indoor environments for navigation, rearrangement, and household interaction. Thus, the asset ecosystem defines not only what the agent observes, but also what it can physically contact, manipulate, and reason about.

\subsubsection{Physics and Rendering Backends}
The physics and rendering backend determines contact dynamics, joint constraints, numerical stability, parallelization, and visual realism. MuJoCo~\cite{todorov2012mujoco} remains widely used for rigid-body control, contact-rich manipulation, and locomotion because of its efficient dynamics and stable contact modeling. MuJoCo Playground~\cite{zakka2025mujocoplayground} further extends this line toward accelerator-based parallel training. Isaac Sim~\cite{nvidia2025isaacsim}, Isaac Gym~\cite{makoviychuk2021isaacgym}, and Isaac Lab/Orbit~\cite{mittal2023orbit} emphasize GPU-parallel simulation, photorealistic rendering, sensor modeling, and synthetic data generation. SAPIEN/PartNet-Mobility~\cite{partnetmobility2020} focus on articulated-object interaction and scalable manipulation under PhysX-based simulation. PyBullet~\cite{coumans2021pybullet} remains useful for lightweight prototyping and baseline environments. CoppeliaSim~\cite{rohmer2013coppeliasim} and PyRep~\cite{james2019pyrep} support flexible task construction, as exemplified by RLBench~\cite{rlbench2019}. Habitat 2.0~\cite{szot2021habitat2} and Habitat 3.0~\cite{puig2024habitat3} target navigation, rearrangement, and human-robot co-habitation.

Specialized engines further extend the physical regime: SoftGym~\cite{lin2021softgym} supports deformable-object manipulation, TDW~\cite{gan2021threedworld} supports multimodal physical interaction, Brax~\cite{freeman2021brax} provides JAX-native accelerator-friendly simulation, Genesis~\cite{genesis2024} targets unified multi-physics simulation, and Newton~\cite{nvidia2025newton} aims at differentiable physical simulation. In this sense, the simulator backend constrains the feasible data distribution: rigid-body engines favor contact-rich control and manipulation, rendering-oriented engines support visually diverse synthetic data, and specialized engines enable deformable, tactile, or differentiable robot learning.


\begin{table}[!t]
\centering
\scriptsize
\setlength{\tabcolsep}{2.0pt}
\renewcommand{\arraystretch}{1.15}
\caption{\textbf{Large-scale simulation datasets for robot learning}, summarized by embodiment configuration, data scale, task coverage, object deformability, dexterous manipulation, and mobile manipulation. In the ``Arm'' column, `S' and `D' denote single-arm and dual-arm, respectively.}
\vspace{-0.2cm}
\label{tab:sim-data}
\rowcolors{2}{white}{w_blue!7}

\begin{tabular}{@{}
>{\raggedright\arraybackslash}m{0.22\textwidth}
>{\centering\arraybackslash}m{0.04\textwidth}
>{\centering\arraybackslash}m{0.070\textwidth}
>{\centering\arraybackslash}m{0.050\textwidth}
>{\centering\arraybackslash}m{0.15\textwidth}
>{\centering\arraybackslash}m{0.058\textwidth}
>{\centering\arraybackslash}m{0.072\textwidth}
>{\centering\arraybackslash}m{0.05\textwidth}
>{\centering\arraybackslash}m{0.05\textwidth}
@{}}
\toprule
\textbf{Dataset} & \textbf{Time} & \textbf{Arm} & \textbf{Embod.} &
\textbf{Traj./Hours} & \textbf{Tasks} & \textbf{Deform.} &
\textbf{Dex.} & \textbf{Mobile} \\
\midrule\midrule

DexGraspNet~\cite{dexgraspnet2023}
& 2022 & S & 1 & 1.32M / - & Grasp & Rigid & \ding{51} & \ding{55} \\

MimicGen~\cite{mimicgen2023}
& 2023 & S & 3 & 50K / - & 16 & Rigid & \ding{55} & \ding{55} \\

DexMimicGen~\cite{dexmimicgen2025}
& 2024 & D & 2 & 21K / - & 9 & Rigid & \ding{51} & \ding{55} \\

DexGraspNet 2.0~\cite{dexgraspnet202024}
& 2024 & S & 1 & 427M / - & Grasp & Rigid & \ding{51} & \ding{55} \\

InternData-A1~\cite{interndataa12025}
& 2025 & S/D & 4 & 637K / 7.4K h & 70 & Soft & \ding{55} & \ding{55} \\

InternData-M1~\cite{internvlam12025}
& 2025 & S & 1 & 244K / - & 200 & Rigid & \ding{55} & \ding{55} \\

NVIDIA GR00T-X Sim~\cite{nvidia2025gr00txsim}
& 2025 & S/D & 6+ & 345K / - & 58+ & Rigid & \ding{51} & \ding{51} \\

SynGrasp-1B~\cite{graspvlasyngrasp1b2025}
& 2025 & S & 1 & 10M / - & Grasp & Rigid & \ding{55} & \ding{55} \\

Dex1B~\cite{dex1b2025}
& 2025 & S & 3 & 1B / - & 2 & Rigid & \ding{51} & \ding{55} \\

MolmoB0T~\cite{deshpande2026molmob0t}
& 2026 & S & 2 & 1.7M / 5.8K h & 8 & Artic. & \ding{55} & \ding{51} \\

\bottomrule
\end{tabular}%

\vspace{0.3em}

\end{table}

\subsection{Benchmarks and Simulation Datasets}
Table~\ref{tab:simulation-benchmarks} and Table~\ref{tab:sim-data} summarize two complementary components of the simulation-data ecosystem. Table~\ref{tab:simulation-benchmarks} focuses on benchmarks and evaluation environments, characterizing their supported embodiments and interaction settings. In this table, `S', `D', and `H' denote single-arm, dual-arm, and humanoid embodiments, respectively; dexterous-hand and mobile capabilities are reported separately in the ``Dex.'' and ``Mobile'' columns. Existing benchmarks cover a broad range of configurations, from single-arm tabletop manipulation~\cite{libero2023,rlbench2019,metaworld2021} to bimanual coordination~\cite{chen2026robodojo,chen2025robotwin}, dexterous manipulation, mobile manipulation~\cite{robocasa2024,robocasa3652026}, and humanoid control~\cite{humanoidbench2024}. Nevertheless, rigid-object manipulation remains dominant, while tactile sensing and deformable-object interaction are supported by a comparatively limited subset of environments. Targeted benchmarks are beginning to fill specific gaps in this coverage: FlatLab~\cite{zhu2026flatlab} addresses flat objects, whose ungraspable configurations and strong geometry and material variation make them poorly served by grasp-centric suites, and provides a simulation benchmark for systematic evaluation rather than the closed settings in which such pre-manipulation strategies are usually tested. DexArena~\cite{ping2026dexarena} evaluates long-horizon dexterous manipulation with 30 stage tasks and 3,000 replay-validated demonstrations. HARP-Bench~\cite{zhou2026harpbench} provides paired evaluation settings and scripted demonstration generation for manipulation around active humans.

Table~\ref{tab:sim-data} summarizes large-scale synthetic and simulation-based data resources in terms of data volume, task coverage, embodiment diversity, and supported manipulation settings. These resources range from task-level manipulation trajectories to large-scale grasp and dexterous-hand samples. Recent datasets increasingly incorporate multiple embodiments and substantially larger data volumes, including hundreds of thousands or millions of trajectories and, for grasp-centric resources, substantially larger collections of generated samples. However, most datasets remain concentrated on rigid or articulated-object manipulation, with comparatively limited coverage of deformable interaction and mobile manipulation. CuraEval~\cite{xu2026curaeval} tests demonstration selection on existing mixed-skill robomimic data and shows that conclusions vary with the evaluation protocol, motivating size-matched random subsets and full-data controls.

\subsection{Synthetic Demonstration Generation}

\textbf{Human-Executed Demonstrations.}
The most direct form of simulation data collection records trajectories produced through human control. This category includes teleoperation, VR-based control, motion-capture retargeting, and play-style interaction in simulated environments~\cite{calvin2022,libero2023,bigym2024}. In this setting, the simulator mainly serves as an instrumented recording environment that can provide synchronized observations, robot states, actions, language annotations, object states, and success labels. Compared with fully scripted generation, human-executed demonstrations can better capture adaptive behaviors, recovery strategies, and temporally coherent manipulation patterns, especially in contact-rich or long-horizon tasks where suitable scripted experts are difficult to specify. However, scalability remains limited by episode-level human labor, operator expertise, and the fidelity of the teleoperation interface. For this reason, human demonstrations in simulation are often used either as direct imitation-learning data or as seed trajectories for later replay, augmentation, or synthetic expansion.

AXIS~\cite{zhao2026axis} lowers the entry cost of human-executed simulation data through browser-based MuJoCo teleoperation, followed by validation, trajectory refinement, and Isaac Sim augmentation. Its evaluated snapshot contains 50,129 episodes across 207 tasks. Matched-volume pretraining experiments support downstream robustness gains in simulation, while physical deployment is illustrated qualitatively. This community-driven collection paradigm separates contributor access from compute-intensive processing; the released dataset remains subject to approval and non-commercial terms.

\textbf{Rule-Based Execution.}
Rule-based execution generates simulation trajectories from manually specified or privileged task-level experts, including heuristic skill compositions~\cite{vlabench2024}, programmatic expert policies~\cite{vimabench2023}, and waypoint sequences~\cite{rlbench2019}. A task is typically decomposed into stages such as selecting the target object, computing a grasp or contact pose, approaching the target, executing the interaction, and checking task completion. Inverse kinematics, task-and-motion planning, model-predictive control, collision-aware motion planning, trajectory optimization, or task-specific controllers may then convert these stage-level goals into dense joint-space or end-effector actions~\cite{maniskill22023,maniskill32025}. This paradigm has been widely used in simulation benchmarks and data-generation environments where privileged states, object poses, articulation states, and collision models are available, including RLBench~\cite{rlbench2019}, ManiSkill2~\cite{maniskill22023}, ManiSkill3~\cite{maniskill32025}, VIMA-Bench~\cite{vimabench2023}, and VLABench~\cite{vlabench2024}. DemoStream~\cite{garrett2026demostream} further optimizes batched task-and-motion-planning solutions for demonstration quality, evaluating how action-space and planning choices affect downstream behavior cloning; its automated physical collection remains bounded by the specified planning models and task setup.

Rule-based execution improves reproducibility and reduces the marginal cost of collecting additional episodes. It is particularly effective for tasks with clearly specified geometric goals, structured workspaces, reliable resets, and well-defined success conditions. Nevertheless, the resulting behavior distribution is bounded by the expressiveness of the programmatically specified expert. Conventional motion planning primarily addresses geometric execution and does not, by itself, determine task semantics, contact affordances, or recovery strategies. As a result, rule-based datasets may underrepresent ambiguous semantic decisions, non-prehensile interactions, contact-rich adaptation, and the diversity of suboptimal but recoverable behaviors encountered during real deployment.

\textbf{Trajectory Playback and Demonstration Expansion.}
Trajectory playback and demonstration expansion start from existing trajectories and reuse them to synthesize additional simulation data. The source trajectories may come from teleoperation, kinesthetic teaching, human video retargeting, motion planning, or previous policy rollouts. In simulation, these trajectories can be replayed, spatially transformed, segmented, stitched, or retargeted under new object configurations, scene layouts, embodiments, and camera conditions. Object-centric augmentation methods such as MimicGen reuse the temporal structure of a small number of demonstrations while adapting them to new object poses and task instances~\cite{mimicgen2023}. This paradigm has been extended to bimanual and dexterous manipulation in DexMimicGen~\cite{dexmimicgen2025}. RoboCasa integrates related automated trajectory-generation methods as one component of a broader framework for household and kitchen manipulation~\cite{robocasa2024}. ScheduleGen~\cite{jawale2026schedulegen} adapts object-relative skills and jointly schedules the connecting motions across arms. DexSeed~\cite{pi2026dexseed} expands human hand--object interactions through grasp-preserving retargeting and physics-based rollout validation, while MOMASPLAT~\cite{kumar2026momasplat} re-renders manipulated objects and retargets actions from new robot base poses.

This class occupies an intermediate position between human-executed data and fully scripted generation. Compared with teleoperation, it substantially reduces repeated human effort; compared with hand-designed rule-based experts, it can preserve the temporal structure, motion style, and task strategies present in human demonstrations. However, its coverage remains conditioned on the diversity and quality of the seed trajectories. When object geometry, contact conditions, or task constraints differ substantially from the original demonstrations, open-loop playback may fail, and closed-loop correction, replanning, or filtering becomes necessary. Therefore, demonstration expansion is most effective when the task admits object-centric transformations or reusable skill segments, but remains limited for tasks requiring qualitatively new strategies or long-horizon semantic decisions.

\textbf{Autonomous Policy and Generative Rollouts.}
A more automated class generates simulation trajectories through learned policies~\cite{SEIL}, reinforcement-learning experts, LLM- or VLM-generated task programs, scene-graph solvers, compositional skill libraries, world models, or generative simulation pipelines. In autonomous policy rollouts, actions are produced by policies trained through reinforcement learning, imitation learning, curriculum learning, or previous rounds of data generation. Such rollouts are commonly collected in simulation environments for multi-task manipulation, dexterous manipulation, articulated-object manipulation, deformable-object manipulation, and whole-body control. Representative environments include Meta-World~\cite{metaworld2021}, DexArt~\cite{dexart2023}, Bi-DexHands~\cite{bidexhands2022}, SoftGym~\cite{lin2021softgym}, PlasticineLab~\cite{plasticinelab2021}, DaXBench~\cite{daxbench2023}, and HumanoidBench~\cite{humanoidbench2024}. Relative to fixed rule-based experts, learned policies can react to observations and may visit policy-induced states, including failures and recoveries. That said, the resulting data distribution is strongly shaped by reward design, exploration strategy, policy competence, simulator fidelity, and data-retention criteria. Low-quality policies may therefore produce large quantities of redundant or unsuccessful data, making validation and filtering essential. DiscoDemo~\cite{park2026discodemo} makes successful-solution diversity an explicit data-generation objective: a privileged latent-conditioned reinforcement-learning policy produces trajectories that are filtered for success and task-specific safety before imitation training. Its four-task evaluation relies on simulator state, resets, and success predicates, so the mechanism does not establish task-independent generation or physical safety guarantees. ClothDojo~\cite{huang2026clothdojo} trains a simulator policy with canonical garment coordinates, then filters successful flattening and folding rollouts and re-renders them as RGB demonstrations for downstream policies.

Two recent findings sharpen how such rollouts should be used. First, the volume of synthetic demonstrations is not the binding constraint: in visual humanoid loco-manipulation, behavior cloning on synthetic demonstrations saturates at a low performance ceiling regardless of how much data is added, whereas refining the cloned policy with reinforcement learning on a single sparse reward breaks through it~\cite{rayyan2026fetchman}. Second, rollouts from an existing generalist policy can themselves serve as training data: generating online interaction rollouts from a zero-shot VLA and fine-tuning on them yields a multi-task policy on the target robot that retains prior tasks and instruction-following priors while acquiring the new embodiment, in contrast to fine-tuning on in-domain expert data alone, which improves the expert task but erodes those priors~\cite{garg2026finetuning}. Both results suggest that self-generated experience is most valuable when paired with a mechanism, whether a reward or the retained prior itself, that constrains what the added data is allowed to overwrite.

Recent generative data factories further shift human effort from writing per-task experts toward designing scalable task, asset, skill, and validation pipelines. LLM-assisted systems such as GenSim~\cite{gensim2024} and GenSim2~\cite{gensim22024} generate simulation tasks, task code, or articulated-object manipulation settings and subsequently use planning or reinforcement-learning solvers to produce demonstrations. RoboGen~\cite{robogen2024} similarly explores generative simulation for task, scene, and supervision generation. RoboTwin~\cite{mu2024robotwin} and RoboTwin2.0~\cite{chen2025robotwin} emphasize scalable bimanual manipulation data generation through generative digital twins, domain randomization, and embodiment-aware execution. InternData-M1~\cite{internvlam12025}, GenManip~\cite{genmanip2025}, and InternData-A1~\cite{interndataa12025} instantiate a more modular form of this paradigm, in which scene graphs, object states, skill specifications, waypoints, motion planning, physics validation, and rendering are decoupled to generate data for instruction following, spatial grounding, and generalist-policy training. Larger simulation platforms and data engines, including RoboVerse~\cite{roboverse2025}, Genie Sim 3.0~\cite{geniesim302026}, ManiTwin~\cite{manitwin2026}, and RoboCasa365~\cite{robocasa3652026}, further extend this direction by integrating asset construction, scenario generation, multi-embodiment execution, rendering, and benchmarking into unified synthetic-data pipelines.

ARSTAG~\cite{li2026arstag} converts a real image and task instruction into a task-scoped simulated scene, robot-feasible expert demonstrations, and policy-learning data. Coordinated agents use feasibility checks and task-consistent randomization to generate recovery-aware trajectories. Transfer experiments across seven tabletop tasks support this workflow on one dual-arm platform, while precision manipulation and long-horizon composition remain difficult. It is therefore a task-specific data-generation method rather than evidence that arbitrary photographs can be converted into reliable robot experience.

AutoRSR~\cite{zhang2026autorsr} uses robot-grounded multi-view workspace scans to reconstruct object assets and register a scene to the robot base frame at metric scale. Reference motions, trajectory interpolation, and domain randomization then produce executable synthetic demonstrations. Physical evaluation covers three tasks on one arm, so the method supports a bounded real-to-sim-to-real workflow rather than unrestricted autonomous scene conversion. Automated registration also depends on synchronized camera poses and robot states; it does not remove the initial scanning and semantic-query requirements.

This category also includes recent efforts to expand synthetic data beyond conventional RGB and proprioceptive trajectories. GraspVLA/SynGrasp-1B~\cite{graspvlasyngrasp1b2025} and Dex1B~\cite{dex1b2025} generate large-scale synthetic action data for grasping and dexterous manipulation. Visuotactile simulation frameworks extend policy rollouts to contact-rich settings by synthesizing tactile observations, elastomer deformation, marker displacement fields, or other multimodal tactile signals. Representative systems include Taxim~\cite{taxim2021}, TACTO~\cite{tacto2022}, Tactile Gym~\cite{tactilegym202022}, Tacchi~\cite{tacchi2023}, DiffTactile~\cite{difftactile2024}, TacEx~\cite{tacex2024}, TacSL~\cite{tacsl2025}, TacCel~\cite{taccel2025}, TacFlex~\cite{huang2025tacflex}, UniVTac~\cite{chen2026univtac}, and Tac2Real~\cite{tac2real2026}. Tac2Real~\cite{tac2real2026} is representative of this trend because it uses high-throughput visuotactile simulation to support online reinforcement learning and zero-shot sim-to-real transfer for contact-rich manipulation.

Beyond physics-based simulation, DreamGen~\cite{dreamgen2025} introduces a related world-model-based direction in which video world models generate synthetic robot videos or neural trajectories from initial visual states and language instructions, followed by pseudo-action recovery for policy learning. These approaches reduce the need for task-specific teleoperation and manual scripting, but they also introduce additional risks, including task invalidity, semantic inconsistency, visual-action misalignment, hallucinated dynamics, planner-induced behavioral bias, and distributional mismatch with real-robot experience.

\subsection{World Models as Simulators}
\label{sec:world-model-simulator}

Beyond conventional physics-based simulation, recent work increasingly explores \emph{world models as learned simulators} for robot learning~\cite{hou2026world}. Rather than relying on manually specified assets, dynamics, and rendering pipelines, world models learn predictive environment dynamics from collected observations and interactions. Conditioned on observation histories, task descriptions, and candidate actions, they generate future observations, latent states, rewards, or task outcomes, thereby transforming existing experience into imagined interaction trajectories. This idea builds on model-based reinforcement learning and latent imagination~\cite{ha2018worldmodels,hafner2019planet,hafner2020dreamer,hafner2023dreamerv3}, and has been extended to real-robot learning and interactive visual simulation by systems such as DayDreamer and UniSim~\cite{wu2023daydreamer,yang2024unisim}.

\subsubsection{World Models for Policy Training and Post-Training}
World models can serve as virtual environments in which policies are optimized without repeatedly interacting with physical robots. Imagined transitions and estimated rewards support reinforcement learning, policy adaptation, and VLA post-training. Representative approaches include RehearseVLA (formerly World-Env), VLA-RFT, DiWA, and World4RL~\cite{rehearse,vlarft,diwa,world4rl}, while recent systems further explore autonomous play, progress estimation, and large-scale world-model-based policy refinement~\cite{sharma2026world,playworld,wmpo,prophrl,rise,gigabrain}. These methods can reduce real-world interaction costs and generate experience in parallel, but their benefits depend on whether the learned simulator preserves the causal effects of robot actions.

A recent extension is to give the learned simulator a contact channel, on the argument that tactile signals are grounded directly in physical contact and may therefore exhibit a smaller simulation-to-real gap than purely visual observations. ViTacWorld~\cite{huang2026vitacworld} builds an action-conditioned visuo-tactile world model by pretraining on large-scale real and simulated visuo-tactile trajectories assembled from public tactile datasets and a constructed simulation environment, then fine-tuning on real-world data. $N_0$-TWAM~\cite{neoteai2026n0twam} scales this idea into a tactile-native world-action model that predicts both future vision and future contact, pretrained with visuo-tactile joint training over tactile-rich demonstrations spanning six embodiments and 450 tasks, and uses a unified force-based tactile representation together with tactile contact events for task staging. From a data-centric view, these systems matter because they turn scarce tactile recordings into a generator of contact-conditioned experience rather than consuming them only as policy inputs.

How such predicted futures should be structured for downstream use remains an open question, and end-to-end training conflates several distinct failure sources. To isolate the future-to-action interface, Oracle Visuo-Tactile Foresight~\cite{yao2026disentangling} supplies paired RGB and tactile futures drawn from verified successful trajectories, so that interface design can be studied without simultaneously debugging physically invalid visual futures and cross-modally inconsistent tactile forecasts. This kind of controlled substitution is a useful template for attributing gains in learned-simulator pipelines to the right component.

\subsubsection{World Models for Policy Evaluation}
Learned simulators can also estimate policy performance before physical deployment. WorldGym and WorldEval study whether generated rollouts can support policy evaluation and checkpoint selection, while Veo-based evaluation examines nominal, out-of-distribution, and safety-related behaviors~\cite{worldgym2026,li2025worldeval,veorobotics2025}. Unlike policy training, the goal here is to predict task progress, completion, failure, or unsafe outcomes rather than directly update the policy. Visual realism alone is therefore insufficient: useful evaluators must preserve action controllability, task semantics, temporal consistency, and the relative ranking of candidate policies.

\subsubsection{World Models as Synthetic Data Engines}
World models further provide a scalable mechanism for synthesizing interaction data conditioned on scenes, instructions, actions, viewpoints, or embodiments. DreamGen generates robot videos and recovers pseudo-actions through latent-action or inverse-dynamics models, while GigaWorld-0 explicitly treats generative world models as data engines for embodied learning~\cite{dreamgen2025,team2025gigaworld}. Compared with conventional augmentation, these methods model temporal evolution and action consequences; compared with physics-based simulation, they reduce dependence on manually constructed assets. However, generated videos do not necessarily contain accurate executable actions, and action recovery may introduce visual-action misalignment.

Overall, world models are evolving from auxiliary predictive components into learned environments for policy optimization, evaluation, and data generation. Their scalability and visual coverage make them a promising complement to physics-based simulators, but prediction errors, model exploitation, action ambiguity, and distribution shift can produce plausible yet physically invalid trajectories. Consequently, learned simulation requires filtering, uncertainty estimation, and calibration against real-robot or physics-based interaction data.

\subsection{Sim-to-Real Gap}
Sim-to-real transfer has long been a central problem in simulation-based robot data, as synthetic trajectories are expected to support real-robot deployment rather than only in-simulator learning~\cite{mu2024robotwin,robocasa3652026,chen2026univtac}. From a data-centric perspective, this gap mainly arises from observation mismatch and interaction mismatch. 

The first concerns the discrepancy between rendered sensory observations and real-robot perception. Although recent simulation datasets improve realism through 3D assets, lighting, camera models, articulated objects, and multimodal sensors~\cite{interndataa12025,deshpande2026molmob0t,internvlam12025}, their observations can still differ from real data in texture, illumination, material appearance, depth noise, camera calibration, occlusion, tactile response, and force signals. 

The second concerns the mismatch between simulated and real robot-environment interactions, which can be further separated into kinematic and dynamic gaps. Kinematic gaps arise from differences in robot morphology, joint limits, coordinate frames, action interfaces, controller frequencies, and calibration, and can often be reduced through accurate robot modeling, real-to-sim alignment, and hardware-consistent control interfaces. Dynamic gaps are harder to eliminate, because real execution involves actuator latency, compliance, torque saturation, backlash, friction, contact, deformation, object inertia, and multi-body interaction effects that are only partially captured by current simulation engines. For efficiency, many simulation pipelines further simplify contact dynamics, use coarse time steps, or approximate robot-environment state updates, producing trajectories that may be visually plausible but dynamically inconsistent with real execution~\cite{da2025surveysimtorealmethodsrl,wang2024simtoreal,beattie2016deepmindlab}. As a result, the core question is not only how much simulation data can be generated, but whether the generated data captures the sensory variations, kinematic consistency, and irreducible dynamic uncertainty that determine real-robot deployment.

\subsection{Advantages and Limitations}
Simulation data provide scalable supervision by generating robot observations, actions, and task outcomes in controllable virtual environments. Compared with real-robot data, they can be collected with lower marginal cost, extensive parallelization, limited safety risk, and no hardware wear. Simulators also provide privileged labels that are difficult to obtain from real systems, such as object poses, segmentation masks, contact states, physical parameters, success signals, and dense rewards. These properties make simulation useful for large-scale policy pretraining, controlled evaluation, scene and task randomization, and the synthesis of diverse interaction trajectories.

A primary limitation of simulation data is limited physical fidelity. Approximations in sensing, actuation, contact, and object dynamics can produce substantial sim-to-real discrepancies, particularly for contact-rich manipulation, deformable objects, reflective materials, liquids, and long-horizon tasks. Moreover, scale does not necessarily imply diversity: as shown by InternData-A1 in Figure~\ref{fig:diversity}, skill-constrained generation can produce many trajectories whose action keypoints remain concentrated within a small region, indicating substantial repetition. Consequently, simulated policies often require domain randomization, system identification, digital-twin reconstruction, or real-world fine-tuning.

\begin{wbtakeaway}
\bSim~\textbf{Simulation data} forms the most scalable and cost-efficient robot-aligned layer of the pyramid. Automated generation enables repeated collection of finely segmented atomic behaviors, providing dense and controllable supervision for learning specific manipulation skills. However, real-world deployment requires policies to internalize physical regularities that cannot be fully recovered from imperfect simulator dynamics—particularly for contact, materials, and object interactions. Scene diversity is likewise hard to guarantee, as generated data remain constrained by available assets. Moreover, scaling skill and task diversity in simulation still hinges on annotation, skill scripting, or human demonstrations; automating skill discovery from video and using it to drive data synthesis will be a critical next step.
\end{wbtakeaway}

\section{General Data}

\begin{figure}
    \centering
    \includegraphics[width=\linewidth]{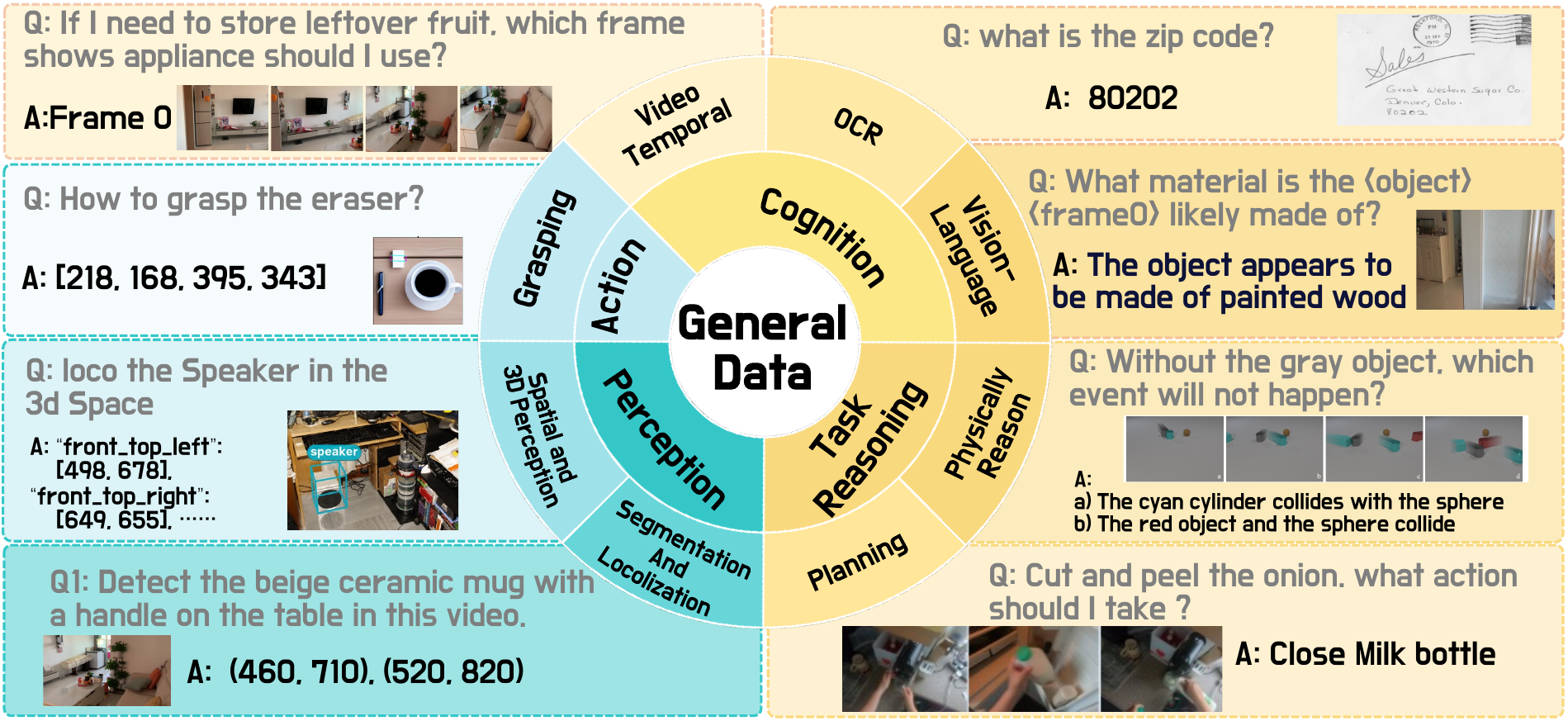}
    \vspace{-0.55cm}
    \caption{\textbf{Overview of General Data Categories.} Overview of general data categories and representative tasks, spanning cognition, task reasoning, perception, and action, with examples of vision-language QA, OCR, video temporal reasoning, physical reasoning, planning, segmentation and localization, spatial and 3D perception, and grasping.}
    \label{fig:general-data}
\end{figure}

\subsection{Data Overview}
Robot learning relies on more than data collected directly from physical robots. Although robot trajectories provide the most direct supervision for action generation and control, they are expensive to collect and usually
cover only a limited range of objects, environments, and tasks. Modern robot foundation models therefore incorporate large amounts of general image, video, language, spatial, and reasoning data. The primary purpose of such data is not to teach a particular robot how to move, but to establish the perceptual, semantic, spatial, temporal, and planning capabilities required before physical actions can be learned efficiently. General data is characterized by massive scale, broad coverage, and high diversity, but remains weakly aligned with executable robot actions and lacks direct robot-environment interaction.

General data should be understood according to the capability they contribute to a robot model. As shown in Figure~\ref{fig:general-data} and Table~\ref{tab:general_datasets}, general vision-language data provide broad semantic knowledge and commonsense reasoning; segmentation and
localization data ground this knowledge in spatial regions; video data introduce temporal and procedural structure; spatial and 3D data provide geometric awareness; and planning-oriented data teach the model to organize observations and goals into executable steps. These capabilities form the cognitive foundation on top of which robot-specific trajectories and control signals are learned.

\begin{table}[!t]
\centering
\caption{\textbf{Representative General-Data Datasets.} Representative datasets grouped by their primary supervision type, with their data modality, scale, supervision targets, and annotation pipeline. VL denotes vision-language, Seg./Loc. denotes segmentation and localization, and EE denotes end effector. Annotation types are categorized as Manual for human-provided labels, Automatic for labels generated from metadata, geometry, simulation, computer-vision methods, or programmatic procedures, and Semi-auto for model-assisted annotation followed by filtering or human verification.}
\vspace{-0.2cm}
\label{tab:general_datasets}
\scriptsize
\setlength{\tabcolsep}{1.6pt}
\renewcommand{\arraystretch}{0.96}
\rowcolors{2}{white}{w_blue!7}
\begin{tabularx}{\linewidth}{@{}
>{\raggedright\arraybackslash}p{0.105\linewidth}
>{\raggedright\arraybackslash}p{0.160\linewidth}
>{\raggedright\arraybackslash}p{0.155\linewidth}
>{\raggedright\arraybackslash}p{0.170\linewidth}
>{\raggedright\arraybackslash}X
>{\raggedright\arraybackslash}p{0.090\linewidth}
@{}}
\toprule
\textbf{Task} &
\textbf{Dataset} &
\textbf{Modality} &
\textbf{Scale} &
\textbf{Labels / Targets} &
\textbf{Anno.} \\
\midrule\midrule

VL (8) &
LLaVA-Instruct-150K~\cite{llava2023visual} &
Image-Text &
158K &
Caption; VQA; instruction; dialogue &
Semi-auto \\

&
LLaVA-OneVision-SI~\cite{llavaonevision2024} &
Image-Text &
3.2M &
VQA; caption; OCR; knowledge; instruction &
Semi-auto \\

&
ShareGPT4V~\cite{sharegpt4v2024} &
Image-Text &
1.2M captions &
Dense caption; image-text alignment &
Semi-auto \\

&
PixMo~\cite{deitke2025molmo} &
Image-Text &
1M img.; 3.3M ann. &
Caption; QA; count; point; document &
Manual \\

&
A-OKVQA~\cite{schwenk2022aokvqa} &
Image-Text &
24.9K QA &
Knowledge VQA; rationale &
Manual \\

&
LLaVA-Video-178K~\cite{zhang2024llavavideo} &
Video-Text &
178K vid.; 1.34M ann. &
Caption; open/MC QA &
Semi-auto \\

&
RoboVQA~\cite{sermanet2024robovqa} &
Robot/Human Video-Text &
238 h; 829.5K pairs &
Plan; success; affordance; past/future QA &
Semi-auto \\

&
EgoTaskQA~\cite{jia2022egotaskqa} &
Video-Text &
40K QA &
State; causal; predictive QA &
Automatic \\

\midrule

Robot Video (4) &
RoVid-X~\cite{deng2026rethinking} &
Video / Depth / Flow &
4M clips; 10K+ h &
Verb; segment; captions; depth; flow &
Automatic \\

&
RoboNet~\cite{dasari2020robonet} &
Robot Video &
162K vid.; 15M frames &
Raw robot trajectories &
- \\

&
RH20T-P~\cite{chen2025rh20tp} &
Robot Video &
38K clips; 67 tasks &
Primitive labels; temporal segments &
Manual \\

&
RobAVA~\cite{sun2025robava} &
Robot Video &
40K videos &
Action class; normal/anomaly attributes &
Manual \\

\midrule

OCR/UI (5) &
OCR-VQA~\cite{mishra2019ocrvqa} &
Image-Text &
207K img.; 1.0M QA &
OCR-based VQA &
Automatic \\

&
TextOCR~\cite{singh2021textocr} &
Image-Text &
28K img.; 900K words &
Text boxes; transcription &
Manual \\

&
DocVQA~\cite{mathew2021docvqa} &
Document-Text &
12.8K img.; 50K Qs &
Document VQA &
Manual \\

&
ChartQA~\cite{masry2022chartqa} &
Chart-Text &
32.7K QA &
Chart QA; numeric reasoning &
Semi-auto \\

&
MultiUI~\cite{liu2025multiui} &
UI-Text &
7.3M &
UI grounding; elements; actions &
Automatic \\

\midrule

Seg./Loc. (8) &
Objects365~\cite{shao2019objects365} &
Image-Box &
638K img.; 10M boxes &
Object boxes; categories &
Manual \\

&
SA-1B~\cite{kirillov2023segment} &
Image-Mask &
11M img.; 1.1B masks &
Promptable masks &
Semi-auto \\

&
ADE20K~\cite{zhou2019semantic} &
Image-Mask &
25K img.; 150 cls. &
Semantic masks; scene labels &
Manual \\

&
PACO-LVIS~\cite{ramanathan2023paco} &
Image-Mask &
641K part masks &
Object/part masks; attributes &
Manual \\

&
RefCOCO~\cite{yu2016modeling} &
Image-Text-Box &
142K expr. &
Referring-expression grounding &
Manual \\

&
PixMo-Points~\cite{deitke2025molmo} &
Image-Text-Point &
428k img.; 2.3M expr. &
Pointing; counting &
Manual \\

&
RoboPoint~\cite{yuan2024robopoint} &
Image-Text-Point &
1.43M QA &
Object/affordance/free-space points &
Automatic \\

&
RoboAfford++~\cite{hao2025roboafford} &
Image-Text-Point &
870K img.; 2.0M QA &
Part/affordance localization &
Semi-auto \\

\midrule

Temporal (6) &
Ego4D NLQ~\cite{grauman2022ego4d} &
Video-Text-Time &
227 h; 19K queries &
Query-to-moment localization &
Manual \\

&
Charades-STA~\cite{gao2017tall} &
Video-Text-Time &
16K moment pairs &
Sentence-to-moment grounding &
Semi-auto \\

&
DiDeMo~\cite{hendricks2017localizing} &
Video-Text-Time &
10K vid.; 40K desc. &
Text-to-moment localization &
Manual \\

&
HiREST~\cite{zala2023hirest} &
Video-Text-Time &
3.4K pairs; 8.6K steps &
Moment; step segments; captions &
Manual \\

&
Moment-10M~\cite{qian2024momentor} &
Video-Text-Time &
10M moments &
Segment-level grounding &
Semi-auto \\

&
COIN~\cite{tang2019coin} &
Video-Step-Time &
11.8K vid.; 46.3K seg. &
Procedure steps; temporal segments &
Manual \\

\midrule

Spatial/3D (7) &
ScanNet~\cite{scannet2017} &
RGB-D / Mesh &
1.5K scenes; 2.5M views &
Depth; pose; mesh; 3D semantics &
Semi-auto \\

&
ScanNet++~\cite{yeshwanth2023scannetpp} &
RGB-D / Mesh &
460 scenes &
Hi-res geometry; views; semantics &
Semi-auto \\

&
ARKitScenes~\cite{baruch2021arkitscenes} &
RGB-D / LiDAR &
5K cap.; 1.7K scenes &
Depth; reconstruction; 3D boxes &
Semi-auto \\

&
WildDet3D~\cite{huang2026wilddet3d} &
Image-3D Box &
1.00M img.; 3.91M boxes &
Open-vocab monocular 3D boxes &
Semi-auto \\

&
MMScan~\cite{lyu2024mmscan} &
3D-Text &
1.4M cap.; 3.04M QA &
3D captions; grounding; QA &
Semi-auto \\

&
ScanQA~\cite{azuma2022scanqa} &
3D-Text &
41K Qs; 800 scenes &
Free-form 3D QA &
Manual \\

&
SQA3D~\cite{ma2023sqa3d} &
3D-Text &
33.4K Qs; 6.8K sit. &
Situated agent-centric 3D QA &
Manual \\

\midrule

Planning (4) &
ALFRED~\cite{shridhar2020alfred} &
Obs.-Lang.-Action &
25K dir.; 8K demos &
Goal/subgoal instructions; actions &
Manual \\

&
EgoPlan-IT~\cite{chen2026egoplan} &
Video-Text-Plan &
50K planning QA &
Goal; subgoal; next action &
Semi-auto \\

&
ShareRobot~\cite{robobrain2025} &
Robot Video/Image-Text-Point &
51.4K eps.; 1.028M QA &
Planning; affordance; EE trajectory &
Semi-auto \\

&
LLaRP~\cite{szot2024llarp} &
3D-Lang.-Action &
114 tasks &
Language task; high-level actions &
Automatic \\

\midrule

Physics/Failure (4) &
CLEVRER~\cite{yi2020clevrer} &
Video-QA &
20K vid.; 300K+ Qs &
Descriptive/causal/predictive/counterfactual QA &
Automatic \\

&
IntPhys~\cite{riochet2018intphys} &
Video-Label &
15K videos &
Possible vs.\ impossible &
Automatic \\

&
InfLevel~\cite{weihs2022inflevel} &
Video-Label &
5.8K real; 75.3K sim. &
Continuity; solidity; gravity; plausibility &
Automatic \\

&
RoboFail~\cite{liu2023reflect} &
Video-Text &
100 sim.; 30 real &
Failure type/cause; recovery &
Manual \\

\midrule

Grasping (4) &
RealVLG-11B~\cite{li2026realvlgr1} &
Image-Text-Mask-Grasp &
165K img.; 1.3M ann.; 11B grasps &
Grounding; box; mask; 4-DoF grasp; contact &
Semi-auto \\

&
GraspNet-1Billion~\cite{fang2020graspnet} &
RGB-D-Point-Grasp &
97K img.; 1.1B grasps &
Scene-level 6-DoF grasps &
Semi-auto \\

&
SuctionNet-1Billion~\cite{cao2021suctionnet} &
RGB-D-Point-Grasp &
97K img.; 1.1B labels &
Contact; normal; suction quality &
Automatic \\

&
DexGraspNet 2.0~\cite{dexgraspnet202024} &
Mesh-Hand-Grasp &
8.3K scenes; 427M grasps &
Hand pose; contact; grasp quality &
Automatic \\

\bottomrule
\end{tabularx}

\end{table}

\subsection{Visual-Language Data}

Vision-language data provide the semantic and reasoning foundation of modern robot models. Whereas robot trajectories usually cover a limited set of environments, objects, and tasks, large-scale image-text and video-text corpora expose models to diverse scenes, activities, tools, materials, and linguistic expressions. Beyond visual recognition and instruction understanding, captioning, visual question answering, and multimodal instruction data provide commonsense knowledge about object attributes, spatial relations, human intentions, functions, and likely event outcomes, which is rarely annotated explicitly in robot demonstrations but is important for task interpretation and planning.

Image-based instruction mixtures, including LLaVA-style data~\cite{llava2023visual,llavaonevision2024}, ShareGPT4V~\cite{sharegpt4v2024}, and PixMo~\cite{deitke2025molmo}, combine captions, questions, detailed descriptions, and reasoning responses from benchmarks such as COCO, VQA, Visual Genome, and A-OKVQA~\cite{lin2014microsoft,antol2015vqa,krishna2017visualgenome,schwenk2022aokvqa}. They teach models to identify scene content, understand relations, compare regions or views, and answer natural-language questions. Video-language data further introduce human actions, object-state changes, temporal order, and procedural structure. Egocentric or robot-view videos annotated with captions or question-answer pairs provide first-person semantic and temporal understanding, although they do not directly supervise robot actions~\cite{grauman2022ego4d,jia2022egotaskqa,zhang2024llavavideo,hou2025fire}.

These datasets are collected from public visual repositories, web pages, video platforms, instructional videos, and existing benchmarks. While earlier corpora relied mainly on human annotations, recent pipelines increasingly use multimodal models to generate captions, questions, answers, rationales, and conversations at scale. Such generation improves coverage and diversity but may introduce hallucinated objects, incorrect relations, or generic language. Consequently, automatic annotations are commonly filtered through confidence estimation, consistency checks, answer verification, model agreement, or human review, and are often combined with smaller high-quality human-annotated datasets.

In robot foundation models, vision-language data are primarily used for multimodal alignment and instruction tuning, establishing broad scene understanding, commonsense reasoning, task comprehension, and question answering, including under ambiguous, insufficient, or noisy instructions~\cite{wu2026abstaineqa,jiang2026reibench,wu2026noisyeqa}. During subsequent robot-specific training, general vision-language data may be retained to preserve semantic and linguistic capabilities, while spatial grounding, planning, grasping, and trajectory supervision connect this broad knowledge to executable behavior.

\subsection{Segmentation and Localization Data}

Object detection, segmentation, and localization remain central to robot perception because manipulation is typically organized around objects and their functional parts. Before grasping a cup, opening a drawer, pressing a button, or placing an item, a robot must identify the relevant object, estimate its location, and determine where interaction should occur. These data therefore complement the semantic knowledge learned from vision-language corpora with explicit spatial grounding.

Classical datasets supervise object categories through bounding boxes, semantic or instance masks, and part-level regions. COCO~\cite{lin2014microsoft}, LVIS~\cite{gupta2019lvis}, Objects365~\cite{shao2019objects365}, ADE20K~\cite{zhou2019semantic}, and SA-1B~\cite{kirillov2023segment} provide broad coverage of everyday objects and scenes, while part-centric resources such as PACO~\cite{ramanathan2023paco} annotate finer-grained functional components. This distinction is particularly important for manipulation: recognizing a drawer is insufficient if the policy cannot localize its handle, and using a tool may require separating its graspable region from its functional end.

Language-conditioned localization further connects visual perception with natural-language instructions. Referring-expression datasets such as RefCOCO~\cite{yu2016modeling} associate descriptions with target boxes or masks, while PixMo-Points~\cite{deitke2025molmo} supervises coordinate-based pointing. Robot-oriented datasets, including RoboRefIt~\cite{lu2023vlgrasp} and RoboPoint~\cite{yuan2024robopoint}, extend these tasks to cluttered indoor and manipulation scenes. Affordance datasets add action relevance by identifying graspable parts, handles, buttons, placement regions, or navigable areas, shifting the prediction target from \emph{what an object is} to \emph{where an interaction should occur}.

These datasets are collected from web images, photographic repositories, recognition benchmarks, tabletop scenes, egocentric recordings, robot-mounted cameras, and synthetic environments. Their annotations may take the form of manually drawn boxes and masks, interactive or model-assisted segmentation, referring expressions, or point clicks. Simulation offers exact object identities, masks, coordinates, and visibility information without manual labeling, while recent pipelines increasingly use multimodal models to generate region descriptions or referring expressions, followed by geometric checks, confidence filtering, or human verification.

For robot learning, the main value of such data is that they transform semantic concepts into spatially actionable targets. A model may understand the instruction ``pick up the red cup,'' but a manipulation system still requires a box, mask, point, or part-level region identifying the cup in the current observation. This representation can then condition grasp generation, navigation, motion planning, or low-level control. Segmentation, grounding, pointing, and affordance localization thus provide a direct interface between general visual understanding and physical execution.

However, annotations from general images do not fully capture the difficulty of robot perception. Robotic scenes often involve heavy occlusion, repeated object categories, unusual viewpoints, motion blur, and partially visible functional parts, while automatically generated expressions may be ambiguous or refer to the wrong instance. Robust systems therefore commonly combine large-scale general-purpose segmentation data with smaller, task-relevant robot datasets and apply relation-aware annotation, filtering, or validation in real manipulation scenes.

\subsection{3D Data}
3D data provide the geometric foundation required for embodied perception and physical interaction. General vision-language data may allow a model to recognize objects and understand scene semantics, but a robot must additionally estimate where objects are located, how far away they are, how they are oriented, and how they are arranged relative to the agent and the surrounding environment. Such data therefore support reasoning about depth, distance, direction, containment, support relations, visibility, free space, and viewpoint changes.

This category includes RGB-D images, point clouds, reconstructed meshes, multi-view observations, camera trajectories, 3D object annotations, and spatial question-answer data. Indoor scene datasets such as ScanNet~\cite{scannet2017}, ScanNet++~\cite{yeshwanth2023scannetpp}, ARKitScenes~\cite{baruch2021arkitscenes}, and 3RScan~\cite{wald2019rio} provide RGB-D sequences together with camera poses and reconstructed geometry. Language-augmented resources such as ScanQA~\cite{azuma2022scanqa} and SQA3D~\cite{ma2023sqa3d} further require models to answer questions about object locations, relative directions, and agent-centric observations. Other datasets generate spatial questions from object coordinates, scene graphs, or reconstructed environments, providing scalable supervision for metric and relational reasoning.

These data are commonly collected using RGB-D cameras, mobile-device depth sensors, stereo systems, LiDAR, or multi-view reconstruction pipelines. Depth maps~\cite{silberman2012nyu,baruch2021arkitscenes} and camera poses may be obtained directly from sensors, visual-inertial tracking, SLAM~\cite{mur2015orbslam}, or structure-from-motion~\cite{schonberger2016structure}. Multiple observations are then fused into point clouds or meshes. Object identities~\cite{scannet2017,wald2019rio}, semantic regions~\cite{scannet2017,yeshwanth2023scannetpp}, 3D boxes~\cite{baruch2021arkitscenes}, and scene relations are often annotated manually~\cite{wald2020learning}, while geometric properties such as distance, relative direction, visibility, and containment can be computed automatically from the reconstructed scene. Synthetic environments provide an additional source of supervision because they can directly output exact depth, surface normals, camera parameters, segmentation labels, and object poses.

For robot learning, the main value of these data is to transform visual observations into a physically meaningful representation of the environment. A robot must determine not only which object is relevant, but also whether it is reachable, whether another object blocks the approach path, whether enough free space is available for manipulation, and from which viewpoint the target can be observed most clearly. Three-dimensional scene representations also allow information obtained from different views to be integrated, which is important when objects are partially occluded or temporarily outside the camera view.

Spatial and 3D supervision consequently supports navigation, collision avoidance, viewpoint selection, object search, grasp planning, and manipulation. It provides the geometric context needed to evaluate whether a semantically reasonable action is physically feasible. In a robot foundation model, these data complement visual-language understanding by grounding object and task concepts in depth, scale, orientation, and scene structure, thereby providing the geometric basis on which planning and control modules operate.

\subsection{Planning Data for Task Decomposition}

Planning data supervise the transformation of a task goal and current observation into an ordered sequence of steps, subgoals, or next actions. Unlike low-level control data, their targets describe abstract task structure, such as locating an object, preparing the workspace, executing a manipulation step, and verifying the resulting state. They therefore provide an intermediate reasoning layer between semantic understanding and embodiment-specific execution.

Such data can be derived from human-written instructions, manuals, instructional or egocentric videos, robot demonstrations, and simulation. Video-based datasets segment demonstrations into meaningful stages and associate observations with next-step predictions, subgoal sequences, or reasoning traces, as exemplified by EgoPlan-IT and EgoCOT~\cite{chen2026egoplan,mu2023embodiedgpt}. Robot trajectories provide more physically grounded supervision through action sequences and object-state transitions, while simulation enables scalable generation and verification using explicit states, action preconditions, and success conditions.

Language and multimodal models are also increasingly used to generate plans or next-action annotations, although such outputs may contain plausible but physically invalid steps. A complementary use is to let a pretrained model supply the plan directly at deployment time rather than as offline supervision: ASCENT performs zero-shot object-goal navigation in multi-floor buildings by pairing frontier ranking with LLM-based contextual reasoning, generalizing to unseen object categories without task-specific retraining~\cite{gong2025ascent}. Such training-free pipelines illustrate how far general-data priors alone can carry high-level embodied decision making, while also exposing their limits, since the resulting plans remain ungrounded in the embodiment's own dynamics. These annotations therefore require filtering through demonstrated action order, state changes, simulator execution, success conditions, or human review. SCOPE~\cite{li2026scope} operationalizes planner feedback by repairing scene-graph inconsistencies when generated task plans fail, coupling scenario diversity with executable planning-data construction. This validation remains tied to the planner and represented scene constraints, rather than certifying open-world physical execution. Combined with robot trajectories and control data, planning supervision helps embodied models decompose long-horizon goals, track task progress, and select executable subgoals.

\subsection{Video and Temporal Data}
Video-temporal data provide embodied models with a form of perceptual memory. Rather than observing a task as a single image, a robot receives a continuous stream of partial observations in which objects may move, change state, become occluded, or disappear from view. The model must therefore connect the current observation with past events and determine when a task-relevant action or state transition occurred. This category mainly includes temporal grounding, event localization, step segmentation, and long-video memory supervision, rather than generic action recognition.

Temporal grounding aligns a language query or task description with a specific video interval. Given a query such as when was the drawer opened?'' or locate the step in which the object was placed into the container,'' the model predicts the corresponding start and end timestamps. This requires searching over the observation history, retrieving the relevant event, and distinguishing it from similar actions at other times. Long-video and episodic-memory data further train models to answer what happened, when it occurred, and where an object was previously observed. Egocentric recordings are especially useful because they provide long, moving, and partially observed streams resembling the perceptual history of an embodied agent.

These datasets are typically collected from instructional, daily-activity, or egocentric videos. Human annotators may segment videos into meaningful steps, write descriptions for each interval, or select temporal boundaries in response to queries. Narration timestamps provide weaker event-language alignment, while larger-scale pipelines may use multimodal models to generate descriptions and candidate timestamps, followed by transcript alignment, consistency filtering, or human verification. Such supervision helps an embodied model retrieve past events, track object-state changes, and recognize task progress over extended interactions.

\subsection{Physical, Causal, and Failure Reasoning}
Physical reasoning data teach models to predict how scenes may change under different actions. Datasets such as CLEVRER~\cite{yi2020clevrer} and IntPhys~\cite{riochet2018intphys} provide supervision for collision, object permanence, stability, future-state prediction, and counterfactual reasoning. They help models distinguish a physically plausible outcome from an impossible one.

Failure-oriented data~\cite{liu2023reflect} such as RoboFail~\cite{liu2023reflect} focus more directly on robot execution. They may label whether a task failed, identify the stage at which the failure occurred, explain its cause, or describe a possible recovery action. Such supervision can be used to train critics, progress estimators, or planning-time verifiers.

Synthetic physics datasets are generated in simulation, where object states, collisions, and counterfactual outcomes can be obtained automatically. Real-world failure data require repeated task execution under varied conditions. Labels may be provided manually, derived from task-success
detectors, or generated by comparing observed state transitions with the intended goal.

\subsection{Grasp Data}
Grasp data specify where and how a robot end effector should contact an object. The appropriate representation depends on the task. Cornell~\cite{lenz2015deep}, Jacquard~\cite{depierre2018jacquard}, and Grasp-Anything~\cite{vuong2023graspanything} provide image-plane grasp rectangles, whereas GraspNet-1Billion~\cite{fang2020graspnet}, ACRONYM~\cite{acronym2020}, and Grasp-Anything-6D~\cite{nguyen2024language6d} provide three-dimensional parallel-jaw grasp candidates. SuctionNet~\cite{cao2021suctionnet} provides surface contact and suction quality, while DexGraspNet-style datasets~\cite{dexgraspnet2023,dexgraspnet202024} represent wrist pose and
multi-finger joint configurations.

Two-dimensional grasp data are efficient to construct and support top-down grasp prediction, but their outputs normally require depth and camera calibration before robot execution. In contrast, 6-DoF datasets directly specify the gripper pose in three-dimensional coordinates and may also include gripper width, collision status, and grasp quality. Language-conditioned datasets such as RefGraspNet~\cite{bhat2026maplegrasp} additionally connect a referring expression to a set of executable grasp poses.

Grasp annotations are generated using several methods. Small 2D datasets may use manually drawn oriented rectangles. Larger synthetic datasets sample candidate poses around object meshes and evaluate them using geometric metrics, collision checking, or physics simulation. Scene-level datasets render or capture RGB-D observations and transform valid object-level grasps into scene coordinates. Real-robot trials may then be used to validate simulation-derived quality scores.

It is important to distinguish direct grasp-pose data from auxiliary grasp resources. Contact maps, human hand-object interaction data, object pose annotations, and 3D object meshes provide useful priors, but they do not
directly specify an executable robot gripper pose. They normally require a
separate grasp generator, retargeting method, or optimization stage.




\subsection{Advantages and Limitations}
General data offer a scale, diversity, and accessibility that no robot-collected source can match. Requiring neither hardware, teleoperation, nor environment resets, they can be assembled from existing web, video, and benchmark resources at low marginal cost, and they remain embodiment-agnostic, so the same corpora are reusable across platforms and tasks. Their contributions are also complementary rather than redundant: semantic and commonsense knowledge from vision-language corpora, spatial grounding from segmentation and localization data, geometric structure from 3D data, perceptual memory from video, task structure from planning data, and physical plausibility from reasoning benchmarks jointly form the cognitive foundation that makes subsequent robot-specific training more efficient.

The corresponding limitation is weak action grounding. General data capture neither proprioception, actuator dynamics, contact forces, nor the physical consequences of executed actions, and therefore cannot by themselves specify what a particular robot should do. Their visual distributions also differ from robot observations, which more often involve occlusion, unusual viewpoints, motion blur, and repeated object instances. Increasingly automated annotation pipelines add a further risk, since generated labels, referring expressions, and plans may be hallucinated, ambiguous, or physically invalid, and require verification before use. General data are consequently best treated as a broad prior over perception, semantics, and reasoning that the more robot-aligned layers must ground in physical execution.

\begin{wbtakeaway}
\bGen~\textbf{General data} supplies web-scale semantic, perceptual, and reasoning priors far beyond what physical-interaction data can provide, but with weak action and contact grounding. It is most valuable for building perception, reasoning, and planning capabilities that higher, more robot-aligned layers later ground in physical execution.
\end{wbtakeaway}

\section{Data Applications in Embodied Foundation Models}

\subsection{Overview}
The preceding sections introduced the collection paradigms and major forms of embodied data. We now shift from data construction to data application, examining how different types of data are used to train embodied foundation models.
We first summarize the data recipes and action representations adopted by representative model architectures. We then discuss three major model families, \ie, \textbf{embodied brain models}, \textbf{vision-language-action models (VLAs)}, and \textbf{world action models (WAMs)}, focusing on how each family utilizes data with and without robot action labels. This perspective highlights the distinct roles of different data types in supporting embodied understanding, control, and dynamics modeling.

\subsection{Data Composition and Action Representation}

\begin{table}[!t]
\centering
\caption{\textbf{Representative VLA and WAM approaches.} The models are grouped and sorted by release time, model type, institution, and data source icons. Left Half: Models from 2023 to 2025; Right Half: Models from 2026. Data sources are indicated by
\DataLegendItem{\DataReal}{real-robot data},
\DataLegendItem{\DataEgo}{egocentric data},
\DataLegendItem{\DataUMI}{UMI data},
\DataLegendItem{\DataSim}{simulation data}, and
\DataLegendItem{\DataGen}{general data}.}
\vspace{-0.2cm}
\label{tab:vla-wam-methods}
\begingroup
\fontsize{7.3pt}{8.0pt}
\selectfont
\setlength{\tabcolsep}{2pt}
\renewcommand{\arraystretch}{1.15}
\begin{minipage}{0.98\linewidth}
\begin{minipage}[t]{0.492\linewidth}
\vspace{0pt}
\centering
\rowcolors{2}{white}{w_blue!7}
\begin{tabularx}{\linewidth}{@{}C{0.105\linewidth}C{0.300\linewidth}ZC{0.08\linewidth}L{0.18\linewidth}@{}}
\toprule
\textbf{Time} & \textbf{Method} & \textbf{Institution} & \textbf{Model} & \textbf{Data} \\
\midrule\midrule
2023.3 & PaLM-E~\cite{driess2023palmeembodiedmultimodallanguage} & Google & VLA & \DataIcons{\DataReal\hspace{0.12em}\DataSim\hspace{0.12em}\DataGen} \\
2023.7 & RT-2~\cite{brohan2023rt2visionlanguageactionmodelstransfer} & DeepMind & VLA & \DataIcons{\DataReal\hspace{0.12em}\DataGen} \\
\midrule
2024.5 & Octo~\cite{team2024octo} & UC Berkeley & VLA & \DataIcons{\DataReal} \\
2024.6 & OpenVLA~\cite{kim2024openvla} & Stanford University & VLA & \DataIcons{\DataReal} \\
2024.10 & GR-2~\cite{cheang2024gr2} & ByteDance & VLA & \DataIcons{\DataReal\hspace{0.12em}\DataEgo} \\
2024.10 & $\pi_0$~\cite{black2026pi0visionlanguageactionflowmodel} & Physical Intelligence & VLA & \DataIcons{\DataReal} \\
2024.10 & RDT-1B~\cite{liu2025rdt} & THU & VLA & \DataIcons{\DataReal} \\
2024.11 & CogACT~\cite{li2024cogactfoundationalvisionlanguageactionmodel} & THU & VLA & \DataIcons{\DataReal} \\
\midrule
2025.1 & SpatialVLA~\cite{qu2025spatialvlaexploringspatialrepresentations} & Shanghai AI Lab & VLA & \DataIcons{\DataReal} \\
2025.1 & UP-VLA~\cite{zhang2025up} & THU & VLA & \DataIcons{\DataReal\hspace{0.12em}\DataGen} \\
2025.3 & HybridVLA~\cite{liu2025hybridvlacollaborativediffusionautoregression} & PKU & VLA & \DataIcons{\DataReal} \\
2025.3 & GR00T N1~\cite{bjorck2025gr00t} & NVIDIA & VLA & \DataIcons{\DataReal\hspace{0.12em}\DataEgo\hspace{0.12em}\DataSim} \\
2025.3 & GO-1~\cite{bu2025agibot} & Shanghai AI Lab \& AgiBot & VLA & \DataIcons{\DataReal\hspace{0.12em}\DataEgo\hspace{0.12em}\DataGen} \\
2025.3 & CoT-VLA~\cite{zhao2025cot} & NVIDIA & VLA & \DataIcons{\DataReal\hspace{0.12em}\DataEgo}\hspace{0.12em}\DataGen \\
2025.4 & UWM~\cite{zhu2025unifiedworldmodelscoupling} & University of Washington & WAM & \DataIcons{\DataReal} \\
2025.4 & $\pi_{0.5}$~\cite{intelligence2025pi_} & Physical Intelligence & VLA & \DataIcons{\DataReal\hspace{0.12em}\DataGen} \\
2025.5 & UniVLA~\cite{bu2025univla} & HKU & VLA & \DataIcons{\DataReal\hspace{0.12em}\DataEgo} \\
2025.6 & SmolVLA~\cite{shukor2025smolvla} & Hugging Face & VLA & \DataIcons{\DataReal} \\
2025.6 & GR00T N1.5~\cite{bjorck2025gr00t} & NVIDIA & VLA & \DataIcons{\DataReal\hspace{0.12em}\DataSim} \\
2025.7 & DreamVLA~\cite{zhang2025dreamvlavisionlanguageactionmodeldreamed} & SJTU & VLA & \DataIcons{\DataReal}\hspace{0.12em}\DataSim \\
2025.7 & EgoVLA~\cite{yang2025egovla} & UC San Diego & VLA & \DataIcons{\DataEgo} \\
2025.7 & Being-H0~\cite{luo2025beingh0} & PKU & VLA & \DataIcons{\DataEgo} \\
2025.7 & GR-3~\cite{cheang2025gr3} & ByteDance Seed & VLA & \DataIcons{\DataReal\hspace{0.12em}\DataEgo\hspace{0.12em}\DataGen} \\
2025.7 & H-RDT~\cite{bi2025hrdt} & THU & VLA & \DataIcons{\DataEgo} \\
2025.8 & GalaxeaVLA(G0)~\cite{jiang2025galaxea} & Galaxea & VLA & \DataIcons{\DataReal} \\
2025.8 & ReconVLA~\cite{song2026reconvla} & HKUST & VLA & \DataIcons{\DataReal\hspace{0.12em}\DataSim} \\
2025.8 & MolmoAct~\cite{lee2025molmoact} & AI2 \& UW & VLA & \DataIcons{\DataReal\hspace{0.12em}\DataGen} \\
2025.8 & EO-1~\cite{qu2025eo1} & Shanghai AI Lab & VLA & \DataIcons{\DataReal\hspace{0.12em}\DataGen} \\
2025.9 & RynnVLA-001~\cite{jiang2025rynnvla001usinghumandemonstrations} & DAMO & VLA & \DataIcons{\DataReal\hspace{0.12em}\DataEgo} \\
2025.10 & X-VLA~\cite{zheng2025xvlasoftpromptedtransformerscalable} & THU & VLA & \DataIcons{\DataReal} \\
2025.10 & InternVLA-M1~\cite{internvlam12025} & Shanghai AI Lab & VLA & \DataIcons{\DataReal\hspace{0.12em}\DataSim\hspace{0.12em}\DataGen} \\
2025.10 & VITRA~\cite{li2025scalablevisionlanguageactionmodelpretraining} & THU & VLA & \DataIcons{\DataEgo} \\
2025.11 & METIS~\cite{fu2025metis} & PKU & VLA & \DataIcons{\DataReal\hspace{0.12em}\DataEgo} \\
2025.11 & iFlyBot-VLA~\cite{zhang2025iflybotvla} & iFLYTEK \& LindenBot & VLA & \DataIcons{\DataReal\hspace{0.12em}\DataEgo\hspace{0.12em}\DataGen} \\
2025.12 & VideoVLA~\cite{shen2025videovlavideogeneratorsgeneralizable} & XJTU & VLA & \DataIcons{\DataReal} \\
2025.12 & Motus~\cite{bi2025motusunifiedlatentaction} & THU & WAM & \DataIcons{\DataReal\hspace{0.12em}\DataEgo\hspace{0.12em}\DataSim\hspace{0.12em}\DataGen} \\
2025.12 & GR00T N1.6~\cite{bjorck2025gr00t} & NVIDIA & VLA & \DataIcons{\DataReal\hspace{0.12em}\DataSim} \\
\bottomrule
\end{tabularx}
\end{minipage}\hfill
\begin{minipage}[t]{0.492\linewidth}
\vspace{0pt}
\centering
\rowcolors{2}{white}{w_blue!7}
\begin{tabularx}{\linewidth}{@{}C{0.095\linewidth}C{0.310\linewidth}ZC{0.08\linewidth}L{0.18\linewidth}@{}}
\toprule
\textbf{Time} & \textbf{Method} & \textbf{Institution} & \textbf{Model} & \textbf{Data} \\
\midrule\midrule
2026.1 & InternVLA-A1~\cite{cai2026internvla} & Shanghai AI Lab & VLA & \DataIcons{\DataReal\hspace{0.12em}\DataEgo\hspace{0.12em}\DataSim} \\
2026.1 & LaST$_0$~\cite{liu2026last0latentspatiotemporalchainofthought} & PKU & VLA & \DataIcons{\DataReal} \\
2026.1 & Being-H0.5~\cite{luo2026beingh05scalinghumancentricrobot} & BeingBeyond & VLA & \DataIcons{\DataReal\hspace{0.12em}\DataEgo\hspace{0.12em}\DataSim\hspace{0.12em}\DataGen} \\
2026.1 & LingbotVLA~\cite{wu2026pragmaticvlafoundationmodel} & Ant Group & VLA & \DataIcons{\DataReal} \\
2026.1 & LingbotVA~\cite{li2026causalworldmodelingrobot} & Ant Group & WAM & \DataIcons{\DataReal\hspace{0.12em}\DataUMI\hspace{0.12em}\DataSim} \\
2026.2 & RDT2~\cite{liu2026rdt2exploringscalinglimit} & THU & VLA & \DataIcons{\DataUMI\hspace{0.12em}\DataGen} \\
2026.2 & Xiaomi-\newline Robotics-0~\cite{cai2026xiaomirobotics0opensourcedvisionlanguageactionmodel} & Xiaomi & VLA & \DataIcons{\DataReal\hspace{0.12em}\DataGen} \\
2026.2 & ABot-M0~\cite{yang2026abotm0vlafoundationmodel} & AMAP CV Lab & VLA & \DataIcons{\DataReal} \\
2026.2 & DM0~\cite{yu2026dm0embodiednativevisionlanguageactionmodel} & Dexmal & VLA & \DataIcons{\DataReal\hspace{0.12em}\DataSim\hspace{0.12em}\DataGen} \\
2026.2 & LDA-1B~\cite{lyu2026lda1bscalinglatentdynamics} & PKU & WAM & \DataIcons{\DataReal\hspace{0.12em}\DataEgo\hspace{0.12em}\DataSim} \\
2026.2 & DreamZero~\cite{ye2026worldactionmodelszeroshot} & NVIDIA & WAM & \DataIcons{\DataReal} \\
2026.2 & EgoScale~\cite{zheng2026egoscale} & NVIDIA & VLA & \DataIcons{\DataReal\hspace{0.12em}\DataEgo} \\
2026.3 & GigaWorld-Policy~\cite{ye2026gigaworldpolicyefficientactioncenteredworldaction} & GigaAI & WAM & \DataIcons{\DataReal\hspace{0.12em}\DataEgo\hspace{0.12em}\DataGen} \\
2026.3 & MolmoB0T~\cite{deshpande2026molmob0t} & AI2 & VLA & \DataIcons{\DataSim} \\
2026.3 & UniDex~\cite{zhang2026unidex} & THU & VLA & \DataIcons{\DataEgo} \\
2026.4 & JoyAI-RA 0.1~\cite{zhang2026joyaira01foundationmodel} & JD & VLA & \DataIcons{\DataReal\hspace{0.12em}\DataEgo\hspace{0.12em}\DataSim\hspace{0.12em}\DataGen} \\
2026.4 & $\pi_{0.7}$~\cite{intelligence2026pi07steerablegeneralistrobotic} & Physical Intelligence & VLA & \DataIcons{\DataReal\hspace{0.12em}\DataEgo\hspace{0.12em}\DataGen} \\
2026.4 & GR00T N1.7~\cite{bjorck2025gr00t} & NVIDIA & VLA & \DataIcons{\DataReal\hspace{0.12em}\DataEgo\hspace{0.12em}\DataSim} \\
2026.4 & Being-H0.7~\cite{luo2026beingh07latentworldactionmodel} & BeingBeyond & WAM & \DataIcons{\DataReal\hspace{0.12em}\DataEgo\hspace{0.12em}\DataSim\hspace{0.12em}\DataGen} \\
2026.4 & MotuBrain~\cite{motubrainteam2026motubrainadvancedworldaction} & Shengshu & WAM & \DataIcons{\DataReal\hspace{0.12em}\DataEgo\hspace{0.12em}\DataGen} \\
2026.5 & Qwen-VLA~\cite{wang2026qwenvlaunifyingvisionlanguageactionmodeling} & Qwen & VLA & \DataIcons{\DataReal\hspace{0.12em}\DataEgo\hspace{0.12em}\DataSim\hspace{0.12em}\DataGen} \\
2026.5 & Wall-OSS-0.5~\cite{yu2026walloss05technicalreport} & X Square Robot & VLA & \DataIcons{\DataReal\hspace{0.12em}\DataGen} \\
2026.5 & MolmoAct2~\cite{fang2026molmoact2} & UW \& AI2 & VLA & \DataIcons{\DataReal\hspace{0.12em}\DataEgo\hspace{0.12em}\DataGen} \\
2026.6 & Wall-WM~\cite{li2026wallwmcarvingworldaction} & X Square Robot & WAM & \DataIcons{\DataReal\hspace{0.12em}\DataEgo\hspace{0.12em}\DataUMI\hspace{0.12em}\DataGen} \\
2026.6 & LaST-HD~\cite{liu2026lasthdlearninglatentphysical} & PKU & VLA & \DataIcons{\DataReal\hspace{0.12em}\DataEgo} \\
2026.6 & Hy-Embodied-0.5-VLA~\cite{zhang2026hyembodied05vlavisionlanguageactionmodelsrealworld} & Tencent Robotics X & VLA & \DataIcons{\DataUMI} \\
2026.6 & Kairos~\cite{kairosteam2026kairos} & ACE Robotics & WAM & \DataIcons{\DataReal\hspace{0.12em}\DataEgo\hspace{0.12em}\DataGen} \\
2026.6 & Qwen-RobotManip~\cite{yuan2026qwen} & Qwen & VLA & \DataIcons{\DataReal\hspace{0.12em}\DataEgo\hspace{0.12em}\DataSim\hspace{0.12em}\DataGen} \\
2026.7 & ABot-M0.5~\cite{chen2026abotm05unifiedmobilityandmanipulationworld} & AMAP CV Lab & WAM & \DataIcons{\DataReal\hspace{0.12em}\DataSim} \\
2026.7 & ACE-Brain-0.5~\cite{brainteam2026acebrain05unifiedembodiedfoundational} & ACE-Robotics & VLA & \DataIcons{\DataReal\hspace{0.12em}\DataGen} \\
2026.7 & InternVLA-A1.5~\cite{ma2026internvlaa15unifyingunderstandinglatent} & Shanghai AI Lab & VLA & \DataIcons{\DataReal\hspace{0.12em}\DataUMI\hspace{0.12em}\DataSim\hspace{0.12em}\DataGen} \\
2026.7 & LingbotVLA 2.0~\cite{wu2026foundationapplicationimprovingvla} & Ant Group & VLA & \DataIcons{\DataReal\hspace{0.12em}\DataEgo} \\
2026.7 & LingbotVA 2.0~\cite{zhang2026nativevideoactionpretraininggeneralizable} & Ant Group & WAM & \DataIcons{\DataReal\hspace{0.12em}\DataEgo\hspace{0.12em}\DataUMI\hspace{0.12em}\DataSim\hspace{0.12em}\DataGen} \\
2026.7 & HumanScale~\cite{ma2026humanscaleegocentrichumanvideo} & PKU & WAM & \DataIcons{\DataEgo} \\
2026.7 & GigaWorld-Policy-0.5~\cite{gigaworldteam2026gigaworldpolicy05fasterstrongerwam} & GigaAI & WAM & \DataIcons{\DataReal} \\
2026.7 & Xiaomi-Robotics-1~\cite{xiaomi2026robotics1} & Xiaomi & VLA & \DataIcons{\DataReal\hspace{0.12em}\DataUMI\hspace{0.12em}\DataGen} \\
2026.7 & Xiaomi-Robotics-U0~\cite{li2026xiaomiroboticsu0} & Xiaomi Robotics & WAM & \DataIcons{\DataReal\hspace{0.12em}\DataEgo\hspace{0.12em}\DataSim\hspace{0.12em}\DataGen} \\

\bottomrule
\end{tabularx}
\end{minipage}
\end{minipage}
\endgroup
\end{table}

\subsubsection{Data Recipes for Embodied Foundation Models}
As summarized in Table~\ref{tab:vla-wam-methods}, the data recipes of embodied foundation models are evolving from single-source robot demonstrations toward increasingly heterogeneous mixtures. This evolution can be clearly observed in successive generations of the same model families. For example, the $\pi$ series expands from real-robot data in $\pi_0$~\cite{black2026pi0visionlanguageactionflowmodel}, to real-robot and general-purpose data in $\pi_{0.5}$~\cite{intelligence2025pi_}, and further incorporates egocentric data in $\pi_{0.7}$~\cite{intelligence2026pi07steerablegeneralistrobotic}. Similarly, LingbotVA~\cite{li2026causalworldmodelingrobot} combines real-robot, UMI, and simulation data, whereas LingbotVA 2.0~\cite{zhang2026nativevideoactionpretraininggeneralizable} extends its recipe to all five levels of the data pyramid.

Despite this growing tendency toward heterogeneous pretraining, the optimal data recipe for embodied foundation models remains an open question. Models pretrained primarily or exclusively on robot data, such as LingbotVLA~\cite{wu2026pragmaticvlafoundationmodel} and DreamZero~\cite{ye2026worldactionmodelszeroshot}, can also achieve strong performance, and current evidence does not establish that incorporating more data sources is inherently preferable. Nevertheless, we argue that different data types can provide complementary capabilities: general-purpose data can support broad visual, semantic, and reasoning abilities; egocentric and UMI data can expand the coverage of scenes, objects, viewpoints, and human behaviors; and simulation data can strengthen particular skills through controllable and scalable interaction generation.

A second trend is the rapid growth of pretraining scale. Earlier models such as OpenVLA~\cite{kim2024openvla} and VideoVLA~\cite{shen2025videovlavideogeneratorsgeneralizable} were pretrained predominantly on robot demonstration corpora from the Open X-Embodiments ecosystem, while GR-2~\cite{cheang2024gr2} represented an early attempt to jointly train on a comparatively limited mixture of robot and egocentric data. By 2026, the reported scale had increased substantially. Qwen-RobotManip~\cite{yuan2026qwen} constructs an approximately 38,100-hour multi-source corpus, including roughly 11.4K hours of open-source robot data and 24,808 hours of robot-compatible trajectories synthesized from 1,933 hours of egocentric video. Xiaomi-Robotics-1~\cite{xiaomi2026robotics1} further reports pretraining on more than 100,000 hours of real-world UMI trajectories, followed by approximately 10,000 hours of cross-embodiment post-training data, including more than 7,200 hours of in-house robot trajectories and over 1,000 hours of instruction-labeled UMI data. Although these quantities are not directly comparable because of differences in modality, processing, filtering, and training stage, they collectively illustrate the progression of embodied pretraining from conventional robot datasets toward corpora containing tens or even hundreds of thousands of hours of interaction data.

A third and closely related trend is the growing importance of egocentric human data. As shown in Table~\ref{tab:vla-wam-methods}, egocentric data has been incorporated into the heterogeneous pretraining recipes of an increasing number of models, including GR-2~\cite{cheang2024gr2}, GR00T N1~\cite{bjorck2025gr00t}, GR-3~\cite{cheang2025gr3}, Motus~\cite{bi2025motusunifiedlatentaction}, $\pi_{0.7}$~\cite{intelligence2026pi07steerablegeneralistrobotic}, Qwen-VLA~\cite{wang2026qwenvlaunifyingvisionlanguageactionmodeling}, Qwen-RobotManip~\cite{yuan2026qwen}, and LingbotVA 2.0~\cite{zhang2026nativevideoactionpretraininggeneralizable}.

Other studies go further by using egocentric data as the primary or exclusive source during pretraining, including EgoVLA~\cite{yang2025egovla}, Being-H0~\cite{luo2025beingh0}, H-RDT~\cite{bi2025hrdt}, VITRA~\cite{li2025scalablevisionlanguageactionmodelpretraining}, UniDex~\cite{zhang2026unidex}, and HumanScale~\cite{ma2026humanscaleegocentrichumanvideo}. HumanScale performs controlled pretraining with up to 5,000 hours of curated egocentric video, while EgoScale~\cite{zheng2026egoscale} scales action-labeled egocentric pretraining to 20,854 hours and reports consistent improvements across subsets ranging from 1,000 to 20,000 hours. Together, these studies suggest that egocentric data is becoming more than an auxiliary source: its scalability and rich coverage of human-object interactions make it an increasingly important pretraining substrate between general web data and embodiment-specific robot trajectories.

Beyond these major trends, recent work is also beginning to reconsider strict data-quality requirements. Whereas earlier training pipelines largely retained successful and carefully curated demonstrations, $\pi_{0.7}$~\cite{intelligence2026pi07steerablegeneralistrobotic} suggests that lower-quality trajectories may still provide useful supervision when paired with sufficiently clear prompts. This observation remains preliminary, but indicates a possible shift from aggressive data filtering toward more explicit conditioning on trajectory quality and intent.

\subsubsection{Action-Space Representation in Robot Foundation Models}

Robot foundation models increasingly integrate action trajectories collected from heterogeneous embodiments, yet these actions are not directly comparable across datasets. Differences arise not only from embodiment-specific control dimensionality and actuator semantics, but also from the coordinate frames and geometric conventions used to express end-effector motion. Accordingly, existing approaches address action-space heterogeneity at two complementary levels: cross-embodiment structural alignment, which determines how different control variables are mapped into a shared action interface, and geometric alignment, which determines the reference frame in which spatial actions are represented. The following discussion reviews representative strategies along these two dimensions and examines their implications for multi-embodiment pretraining and transfer.

\noindent\textbf{Cross-Embodiment Action-Space Representations.}
Existing embodied foundation models reconcile heterogeneous robot action spaces through three principal strategies: \emph{embodiment-specific projection}, \emph{fixed-dimensional zero-padded action interfaces}, and \emph{semantic action slots}. These strategies differ in the level at which action-space heterogeneity is resolved. The first preserves embodiment-specific action interfaces and aligns them only within the shared model representation; the second standardizes the action-vector dimensionality without enforcing consistent physical semantics; and the third explicitly organizes heterogeneous robot actions into a common set of semantically defined control dimensions.

\emph{(1) Embodiment-specific projection.} This strategy preserves the native action dimensionality and control semantics of each robot while using embodiment-specific projectors, adapters, or action heads to map heterogeneous state-action representations into and out of a shared policy backbone. It therefore enables representation-level sharing without replacing robot-specific control interfaces, as adopted by GR00T N1~\cite{bjorck2025gr00t}, Octo~\cite{team2024octo}, and SmolVLA~\cite{shukor2025smolvla}.

\emph{(2) Fixed-dimensional zero-padded action interfaces.} This strategy reserves a common action-vector dimensionality across embodiments. Robots with lower-dimensional action spaces fill unused dimensions with zeros, typically together with validity masks that exclude inactive dimensions from the learning objective. It provides a uniform tensor interface for joint training but does not require corresponding dimensions to share the same physical semantics across robots~\cite{black2026pi0visionlanguageactionflowmodel,wang2026qwenvlaunifyingvisionlanguageactionmodeling,yang2026abotm0vlafoundationmodel}.

\emph{(3) Semantic action slots.} This strategy assigns fixed physical meanings to predefined dimensions or blocks in a shared action vector, such that corresponding entries preserve consistent control semantics across embodiments while unavailable components are masked or zero-padded. Qwen-RobotManip\cite{yuan2026qwen} provides a representative implementation by defining an 80-dimensional canonical state-action vector composed of two 29-dimensional arm blocks and 22 reserved dimensions. Each arm block allocates 7 dimensions to arm joints, 9 to the end-effector pose, 1 to the parallel gripper, and 12 to dexterous-hand joints. Different embodiments populate only the relevant subsets, while inactive fields are zero-filled and excluded from the flow-matching objective using per-dimension binary masks. Related semantic-slot designs are adopted by RDT-1B~\cite{liu2025rdt}, Being-H0.5~\cite{luo2026beingh05scalinghumancentricrobot}, Green-VLA~\cite{apanasevich2026greenvlastagedvisionlanguageactionmodel}, UniDex~\cite{zhang2026unidex}, JoyAI-RA 0.1~\cite{zhang2026joyaira01foundationmodel}, Wall-OSS-0.5~\cite{yu2026walloss05technicalreport} and LingbotVLA 2.0~\cite{wu2026foundationapplicationimprovingvla}. Compared with shape-level padding, this formulation provides stronger cross-embodiment alignment by unifying both action dimensionality and physical semantics.

\noindent\textbf{Geometric Action Representations.}




Beyond dimensional and semantic alignment, better cross-dataset alignment may also require considering the coordinate frames used to represent end-effector actions, since equivalent motions can acquire different numerical values under different geometric conventions. Existing approaches can be broadly grouped into three categories.

\emph{(1) Robot-centric representations.} This strategy expresses end-effector poses or motion commands in a world or robot-base frame, providing a direct interface to conventional robot controllers. However, the resulting representations may remain coupled to embodiment-specific base poses, mounting configurations, and workspace conventions, as exemplified by Open X-Embodiment~\cite{o2024open}.

\emph{(2) Camera-centric representations.} This strategy transforms actions into a camera coordinate frame, aiming to assign more consistent numerical representations to visually similar motions across embodiments and scenes. Camera-centric action representations are adopted by Qwen-RobotManip~\cite{yuan2026qwen} and OC-VLA~\cite{zhang2026grounding}.

\emph{(3) Wrist-centric representations.} This strategy expresses fingertip poses, hand keypoints, or local hand motions in a canonical wrist frame, aiming to separate local hand articulation from global wrist translation and rotation. Such representations are used by METIS~\cite{fu2025metis} and LDA-1B~\cite{lyu2026lda1bscalinglatentdynamics}.

Regardless of the selected coordinate frame, its implementation should clearly specify the frame origin and axes, coordinate-system handedness, tool-center point, absolute-versus-delta action mode, rotation parameterization, and physical units. However, existing studies provide limited controlled ablations across these representations, and it remains unclear whether any coordinate convention consistently offers superior transfer or control performance across datasets and embodiments.
\subsection{Applications for Embodied Brain}

Embodied brain models generally refer to vision-language models pretrained on physical-world understanding tasks to support robot perception and high-level decision-making. Unlike VLAs and WAMs, which directly emphasize action generation or dynamics prediction, these models focus on spatiotemporal perception, physical reasoning, grounding, temporal memory, and task planning. Representative models, including RynnBrain~\cite{rynnbrain2026}, HY-Embodied~\cite{hyembodied2026}, and Pelican-VL~\cite{pelicanvl2025}, strengthen these capabilities before or alongside downstream control.

Accordingly, their training data need not always contain low-level robot actions; the key is whether the supervision contributes to physically meaningful representations. General multimodal data can support semantic understanding, visual commonsense~\cite{hou2026interlv}, OCR, memory, and spatial reasoning, whereas interaction data containing actions, states, or motion cues can provide affordance, grasp, trajectory, and planning supervision. Recent embodied brain models therefore combine general multimodal data with embodied grounding data~\cite{rynnbrain2026,robobrain2025}. Based on the available supervision, we organize these data into two categories: action-free data for embodied understanding and action-labeled data for physical grounding.

\subsubsection{Action-free Data for Embodied Understanding}
A substantial portion of the data used to train embodied brains contains no explicit action annotations. This includes general-purpose image-text, video-text, VQA, OCR, spatial and temporal QA, and instruction-following data, as well as egocentric videos re-annotated for captioning, question answering, temporal memory, or task understanding. Although lacking robot states and control commands, such data provide essential supervision for object recognition, language understanding, visual commonsense, spatial-temporal reasoning, object-function inference, and task comprehension~\cite{rynnbrain2026,sensenova_si2025}.

Action-free videos can also support robot-centric video pretraining through temporal prediction, masked video modeling, or future-frame generation. These objectives allow models to learn object permanence, motion patterns, hand-object interactions, state transitions, and coarse physical dynamics without requiring action labels for every segment. Representative examples include GR-1, V-JEPA~2, DreamDojo, and physical-AI video foundation models such as Cosmos Predict~\cite{GR-1,vjepa2,dreamdojo,CosmosPredict}. The resulting visual-temporal representations can serve as backbones for embodied brains or be coupled with planning and action modules during downstream adaptation.

Nevertheless, video-only pretraining provides primarily observational rather than executable knowledge. It does not necessarily reveal which action causes a transition or whether that transition is feasible for a particular embodiment; action-conditioned trajectories, robot states, inverse-dynamics supervision, or policy adaptation are still required for executable control. The primary role of action-free data is therefore to support physical-world understanding rather than embodiment-specific action alignment. General multimodal data provide broad semantic and commonsense priors, while egocentric data expose embodied brains to first-person viewpoints, occlusions, hand-object interactions, object-state changes, and long-horizon task progress~\cite{rynnbrain2026,hyembodied2026}.

\subsubsection{Action-Labeled Data for Physical Grounding}
Action-labeled data contain explicit interaction signals or demonstrations from which such signals can be recovered. They include real-robot trajectories, handheld-gripper demonstrations, simulation rollouts, and interaction videos containing robot states, hand or gripper motion, object-state changes, or action annotations. For embodied brains, these signals are typically not used as raw control targets, but are converted into more transferable supervision, such as affordance locations, grasp poses, trajectory waypoints, action boundaries, subtask labels, and next-action annotations. RoboBrain and ShareRobot exemplify this approach by treating task planning, object affordance, and end-effector trajectories as core supervision signals~\cite{robobrain2025}.

Such supervision supports three closely related capabilities. First, affordance labels identify functionally meaningful interaction regions, such as handles, buttons, contact surfaces, and placement areas. They can be manually annotated or recovered through segmentation, hand or gripper tracking, contact estimation, and object-state change detection~\cite{robobrain2025,pelicanvl2025}. Second, tracking, pose estimation, SLAM, and camera-motion recovery can convert demonstrations into 2D or 3D trajectories, sparse waypoints, and approach directions. These targets provide richer physical information than static affordances while remaining more transferable than embodiment-specific joint commands~\cite{robobrain2025,pelicanunified2026}. Third, demonstrations can be segmented into atomic actions, subgoals, and step boundaries, enabling embodied brains to recognize task progress and predict subsequent actions~\cite{rynnbrain2026,robobrain2025}.

Different data sources provide complementary action supervision. Real-robot data offer embodiment-aligned states and control traces; handheld-gripper data provide a more scalable source of end-effector-like motion; and simulation provides automatically accessible object states, contacts, trajectories, success conditions, and failure labels. Overall, action-labeled data provide a bridge between physical understanding and executable behavior. Rather than requiring embodied brains to directly imitate low-level controls, they ground perception and reasoning in affordances, trajectories, and task structures, which can subsequently support downstream VLA and WAM training~\cite{hyembodied2026,cosmos3_2026,pelicanunified2026}.

\subsection{Applications for Vision-Language-Action Models}
Vision-Language-Action (VLA) models are typically built upon vision-language model backbones and extend their multimodal understanding capabilities toward downstream robot control. While embodied VLMs primarily focus on perception, grounding, spatio-temporal reasoning, and high-level task understanding, VLA models further emphasize how multimodal representations can be translated into executable behaviors. Accordingly, a VLA model is expected not only to recognize task-relevant objects, states, and affordances, but also to plan action sequences, predict low-level action chunks, and adapt its behavior to specific robot embodiments and environments. Recent systems such as OpenVLA~\cite{kim2024openvla}, Octo~\cite{team2024octo}, and $\pi_{0.5}$~\cite{intelligence2025pi_} exemplify this paradigm by integrating large-scale vision-language pretraining with embodied data, enabling generalist manipulation policies that can follow language instructions, ground task-relevant objects, and generate actions across diverse tasks, scenes, and robot platforms.

The action-oriented nature of VLA models involves two complementary levels of capability. On the one hand, a VLA model must understand the scene and environment, including task-relevant object localization, object tracking, spatial relations, object states, affordances, and temporal context. These capabilities build on high-level vision-language understanding and provide the grounding needed to decide what should be done. On the other hand, a VLA model must translate such grounded understanding into embodiment-specific planning and execution, such as action chunk generation, and adaptation to the robot's morphology and action space. This dual objective leads to a distinctive view of embodied training data: action-free data mainly support embodied understanding and physical grounding, while action-labeled data directly supervise policy learning and action generation.

\subsubsection{Action-Labeled Data}
Action-labeled data constitute the primary supervision source for training VLA models, as they explicitly specify how visual-language observations should be mapped to executable robot behaviors. A common strategy is to discretize continuous robot actions into action tokens, allowing actions to be generated autoregressively in the same format as language tokens. This formulation makes it possible to reuse pretrained VLM backbones and preserve the semantic knowledge acquired from large-scale vision-language pretraining, while adapting the model to robot control through trajectory supervision~\cite{brohan2023rt2visionlanguageactionmodelstransfer,kim2024openvla}.

However, discrete action tokens may be insufficient for high-frequency, fine-grained, or dexterous manipulation, where small errors in continuous control can substantially affect execution. To address this limitation, recent VLA systems increasingly adopt continuous action heads. Diffusion-based policies~\cite{liu2025rdt,wen2025diffusionvla,bjorck2025gr00t} formulate action generation as a conditional denoising process over continuous action sequences, enabling expressive modeling of high-dimensional and potentially multimodal action distributions. In parallel, flow matching-based policies~\cite{intelligence2025pi_,shukor2025smolvla,cai2026internvla} learn a continuous vector field that transports noisy action samples toward executable action trajectories, offering a direct and efficient formulation for high-frequency continuous control. These approaches show that action-labeled data not only provide direct control supervision, but also shape the action representation used by VLA models, from language-like discrete tokens to expressive continuous action distributions. The choice of action representation in turn governs which trajectories can be pooled at all. G0.5~\cite{liu2026g05} makes this explicit by emitting reasoning and action tokens from a single autoregressive decoder under one objective, using a learnable cross-embodiment action tokenizer that maps heterogeneous robot actions into a shared vocabulary, so that the pretrained backbone remains a decision-maker rather than a context encoder. Because such a tokenizer is fitted to the corpus it is trained on, the composition and embodiment coverage of the action-labeled pool directly determine how transferable the resulting vocabulary is.

Action-labeled data are also scarce in a second sense: long-horizon demonstrations are rarely annotated at the level of reusable sub-behaviors, which limits compositional generalization to out-of-distribution task sequences. SkillMemo~\cite{wang2026skillmemo} responds by implicitly decomposing long-horizon demonstrations into latent atomic skills through expert-guided trajectory segmentation and storing skill-level features in a dynamic episodic memory bank, recovering compositional structure from existing trajectories rather than collecting new ones. This is the policy-side counterpart of the annotation-enrichment trend noted for real-robot data.

Beyond robot-collected trajectories, egocentric human videos with hand poses provide a scalable source of supervision for dexterous VLA learning. Some approaches reconstruct human hand motions, retarget them to a specific robot hand, and further reduce the visual embodiment gap through hand inpainting or robot-hand rendering~\cite{qin2022dexmv,shaw2024learning,xu2025dexumi,li2025h2r}. Other approaches map human and robot demonstrations into a shared action or interaction space, such as wrist and fingertip poses, MANO parameters, 3D keypoints, or entity-level hand-object representations, enabling large-scale pretraining on human data followed by lightweight adaptation to robot embodiments~\cite{yang2025egovla,luo2026beingh05scalinghumancentricrobot,fu2025metis,zheng2026egoscale,zhang2026unidex}. AdvDex~\cite{zhao2026advdex} combines both moves and adds a tactile channel: it releases OmniShare, a large-scale multimodal dataset of human manipulation demonstrations providing kinematic supervision and tactile measurements, defines a Joint-Aligned Action Space as a canonical representation across embodiments, and uses adversarial learning to prevent task-relevant visual cues from becoming entangled with embodiment-specific appearance, which is a recurring failure mode when heterogeneous human and robot data are pooled. These methods broaden the scope of action-labeled data beyond directly collected robot trajectories and offer a promising route toward scaling dexterous policy learning.

\subsubsection{Action-Free Data}

\noindent\textbf{Action-Proxy Learning from Action-Free Videos.} Large-scale videos provide abundant evidence about how objects move and states change during manipulation, despite lacking the corresponding action labels. To exploit such data for VLA learning, existing methods derive action proxies from temporal observations, either as implicit latent representations or explicit geometric structures such as point trajectories and motion fields. These proxies are not assumed to be executable robot actions. Instead, they encode visual transitions or interaction dynamics that can be learned at video scale and subsequently grounded in a robot-specific action space.

Latent actions are compact representations of action-relevant state transitions learned from observation sequences without explicit action labels. LAPA~\cite{ye2025latent} learns discrete latent actions from unlabeled videos and uses them as pseudo-action labels for VLA pretraining. Moto~\cite{chen2025moto} represents visual motion as latent motion tokens for autoregressive motion-prior learning. UniVLA~\cite{bu2025univla} introduces task-centric latent actions to suppress task-irrelevant visual dynamics. Villa-X~\cite{chen2025villa} further constrains latent actions with visual and proprioceptive future prediction, while CLAP~\cite{zhang2026clap} and ConLA~\cite{dai2026conla} use contrastive objectives to disentangle action-relevant dynamics from nuisance visual changes. These methods are scalable because they avoid explicit hand-object annotation, but the learned latent space still requires robot demonstrations, action decoders, or downstream alignment to become executable on a specific embodiment.

A complementary body of work explicitly reconstructs spatial or geometric representations of manipulation from video. Although most of these methods are not VLA models in the strict sense, they demonstrate how such representations can bridge visual observation and robot control. VideoDex~\cite{shaw2023videodex} transfers human hand motion extracted from video to robot policies, while ViSA-Flow~\cite{chen2025visaflow} learns visual motion representations from human demonstrations before adapting them with robot data. General Flow~\cite{yuan2024generalflow} represents manipulation as object-centric motion fields, and Track2Act~\cite{bharadhwaj2024track2act} converts predicted point trajectories into object transformations and robot actions. VidBot~\cite{chen2025vidbot} and Actron3D~\cite{zhang2025actron3d} further recover three-dimensional trajectories or transferable action representations from human videos. Compared with latent actions, these geometric representations are more interpretable and transferable to robot control, but depend more heavily on reliable perception and cross-embodiment alignment.

\noindent\textbf{Hierarchical and Intermediate Supervision.} Another way to exploit action-free data is to use it for learning hierarchical or intermediate supervision that guides downstream action generation. General visual-language data, human videos, and task demonstrations can provide high-level semantic knowledge, task decomposition, subgoal prediction, and chain-of-thought reasoning, which help VLA models decide what should be done before predicting how to act. For example, $\pi_{0.5}$~\cite{intelligence2025pi_} uses heterogeneous data to improve semantic scene understanding and high-level task structure modeling, while CoT-style VLA models~\cite{zhao2025cot,zhong2026acot,zhong2026dualcot,shou2026halo} introduce explicit intermediate reasoning, such as visual goals, linguistic plans, or action-level thoughts, to support long-horizon manipulation. In addition, action-free or weakly annotated data can provide affordance-level supervision, such as predicting task-relevant interaction regions, graspable areas, or functional object parts, which serves as a spatially grounded bridge between perception and control. Recent affordance-aware VLA models~\cite{li2025coa,yu2026affordancevla} use such intermediate representations to focus the policy on where and how to interact, thereby improving the robustness of action generation in complex scenes.
\subsection{Applications for World Action Models}
World Action Models (WAMs) model the temporal evolution of the physical world and the effect of actions on this evolution. In contrast to embodied VLMs, which emphasize perception and reasoning, and to VLA models, which directly map multimodal observations to executable behaviors, WAMs treat world dynamics as the primary learning objective, so that a trained WAM can function as a learned simulator, a planner, a policy evaluator, or a generator of synthetic interaction data.

This paradigm originates from latent world models in model-based control, where the Dreamer family~\cite{hafner2023dreamerv3,wu2023daydreamer} learns action-conditioned latent dynamics and optimizes policies through imagined rollouts in both simulated and real environments. Generative video modeling subsequently extended world modeling to open-domain visual data, as represented by UniSim~\cite{yang2024unisim}, GAIA-1~\cite{hu2023gaia1}, and the Genie series~\cite{bruce2024genie,deepmind2025genie3}. In robotics, recent systems including GR-2~\cite{cheang2024gr2}, IRASim~\cite{irasim2025}, EnerVerse~\cite{huang2025enerverse}, Genie Envisioner~\cite{liao2025genieenvisioner}, V-JEPA~2~\cite{vjepa2}, WorldVLA~\cite{cen2025worldvla}, and Cosmos~\cite{cosmos3_2026} apply this paradigm to manipulation, coupling action-conditioned future prediction with action generation and policy evaluation.

Recent surveys organize this rapidly expanding design space along complementary axes: by output representation, spanning video-, occupancy-, and LiDAR-based world modeling and generation~\cite{kong2025worldmodeling}, and by capability level, separating models that merely predict from those that act as controllable simulators or self-improving agents, together with the physical and digital regularities each is expected to respect~\cite{chu2026agentic}. These organizing views are useful here because they make explicit which data a given world-model role demands: passive prediction can be learned from observation alone, whereas simulator- and evaluator-style uses additionally require action-conditioned and outcome-labeled interaction.

From a data perspective, WAM training data can be divided according to whether an action condition is available for modeling world transitions. Action-labeled data, including real-robot trajectories, UMI-style demonstrations, and simulation rollouts from the upper layers of the data pyramid, provide executable or action-like signals that specify which action induces a given transition. Action-free data, including general multimodal corpora and egocentric or exocentric videos from the lower layers, support only observation-level future prediction. In practice, most systems follow a two-stage recipe, in which large-scale action-free pretraining establishes general dynamics priors and action-conditioned post-training grounds these priors in interaction data~\cite{GR-1,cheang2024gr2,vjepa2,cosmos3_2026}.

\subsubsection{Action-Labeled Data}
Most WAMs are built upon diffusion-based video generation models and therefore formulate action prediction as a continuous generative problem. Given a ground-truth action sequence, noise is added according to a diffusion schedule, and the model learns to recover the clean actions through a diffusion or flow-matching objective similar to that used for video generation. Architecturally, Genie Envisioner attaches a lightweight flow-matching action expert to its pretrained world-model backbone~\cite{liao2025genieenvisioner,lou2026predicting}. Motus adopts a Mixture-of-Transformers architecture with separate video and action experts, allowing the two modalities to interact while retaining modality-specific parameters~\cite{bi2025motusunifiedlatentaction,li2026efficient}. VideoVLA jointly models video and action streams within a multimodal Diffusion Transformer~\cite{shen2025videovlavideogeneratorsgeneralizable}, whereas Cosmos Policy directly encodes actions as latent frames and processes them using the original video diffusion backbone without introducing a separate action architecture~\cite{kim2026cosmos}.

An important exception is WorldVLA, which is built upon an autoregressive generative model rather than a diffusion model~\cite{cen2025worldvla}. It discretizes both visual observations and continuous robot actions into tokens and jointly predicts them through autoregressive next-token prediction with cross-entropy losses. A related design compresses the prediction target instead of the backbone: in autonomous driving, Xiaomi OneVL routes reasoning through a small set of latent tokens trained with auxiliary decoders that reconstruct textual chain-of-thought and forecast future-frame tokens, so that world dynamics are absorbed into the latent space during training while inference remains a single parallel pass~\cite{lu2026onevl}. Such latent formulations are attractive from a data standpoint, since the future-prediction objective can be supervised by raw future observations rather than by additional action or reasoning annotations. 

The use of action labels in WAMs can therefore be broadly summarized into two paradigms: continuous action denoising with diffusion or flow-matching objectives, and discrete action-token prediction with autoregressive objectives. In both cases, action supervision aligns the model's representation of future world evolution with executable robot control.

A further data demand arises when the action label must describe the whole body rather than an arm. Most WAMs are arm-centric or video-centered, and extending them to legged and humanoid platforms requires trajectories in which locomotion, posture, balance, and manipulation are recorded as one coordinated behavior. $\omega$-0~\cite{li2026omega0} predicts controller-compatible whole-body action latents directly from a language instruction, visual observation, and proprioceptive state, and replaces future-video reconstruction with a lighter objective that predicts compact future observation embeddings. DECOWAM~\cite{ma2026decowam} instead argues that whole-body prediction requires disentangling factors that fixed-base models conflate, separating camera ego-motion from base and arm actions through dedicated conditional interfaces, and releases ARMDOG, a real-robot dataset synchronizing video, whole-body state and action, and language. Both illustrate a general point for this survey: the granularity of the recorded action space, not only the number of trajectories, determines which world-model roles a corpus can support.
\subsubsection{Action-Free Data}
Action-free data establish the basic world priors of WAMs. This category comprises general multimodal data and egocentric or exocentric videos that contain no explicit or recoverable action variables. The corresponding supervision is observation-only future prediction, such as predicting future frames, visual latents, scene states, or language descriptions of state changes. Such objectives teach the model how the world typically evolves but not which action drives a given transition, and action-free data therefore contribute broad temporal and physical priors rather than executable action-conditioned dynamics.

\noindent\textbf{General Video and Multimodal Data.}
Web-scale video corpora constitute the largest source of dynamics supervision. UniPi~\cite{du2023unipi} formulates decision making as text-conditioned video generation, UniSim~\cite{yang2024unisim} aggregates heterogeneous video and interaction data into an interactive simulator of the real world, and Cosmos~\cite{cosmos3_2026} pretrains omnimodal world models on web-scale video prior to physical-AI post-training. Prediction need not occur in pixel space: V-JEPA~2 learns latent video prediction from over one million hours of Internet video~\cite{vjepa2}. These data provide generic but essential priors, including object permanence, scene geometry, physical plausibility, and the typical temporal structure of everyday tasks.

\noindent\textbf{Egocentric and Exocentric Videos.}
Egocentric datasets such as EPIC-KITCHENS~\cite{Damen2018EPICKITCHENS}, Ego4D~\cite{grauman2022ego4d}, and Ego-Exo4D~\cite{grauman2024egoexo4d} are distributionally closer to robot observations than generic web video. They expose the model to first-person interaction, hand-object occlusion, viewpoint changes, object-state transitions, and long-horizon task progress, which are precisely the observation patterns encountered by deployed manipulation systems and are difficult to obtain from other sources.


Viewed through the data pyramid, action-free data at the base provide scalable and diverse observational priors, while higher levels progressively add semantic, latent, and embodiment-specific action grounding. Trained world models in turn feed back into the pyramid by generating synthetic interaction data~\cite{dreamgen2025,cosmos3_2026} and by substituting for physical environments in policy evaluation~\cite{worldgym2026,robowmbench2026}. Unified systems such as Pelican-Unified~\cite{pelicanunified2026} and Cosmos~3~\cite{cosmos3_2026} further demonstrate how heterogeneous supervision—including visual, language, future-generation, and action data—can be incorporated within a shared modeling framework.

\begin{wbtakeaway}
\bApp~ Embodied foundation-model training recipes are shifting toward more heterogeneous data sources and substantially larger scales. Egocentric data are also playing an increasingly important role, alongside growing tolerance for noisy or lower-quality trajectories. Nevertheless, \textbf{the optimal recipe remains unresolved}, as the contribution of each data source has not been systematically isolated and robot-only models can still achieve strong performance. Across embodied brains (VLMs), VLAs, and WAMs, action-unlabeled data primarily preserve general capabilities or support latent-dynamics learning, whereas action-labeled data ground models in physical interaction and executable control.
\end{wbtakeaway}
\section{Challenges and Future Directions}

This section organizes future directions for robot learning data around three questions: what to collect, how to collect it, and how to use it. First, what to collect concerns forms of interaction information that remain insufficiently represented in existing datasets. In particular, failure and recovery trajectories are needed to expose models to unsuccessful behaviors and corrective actions, while tactile data are required to characterize contact states, forces, material properties, and fine-grained interaction dynamics that cannot be fully inferred from visual observations alone.

Second, how to collect focuses on scalable and adaptive data acquisition. Future collection pipelines should reduce dependence on costly manual teleoperation, expand task and scene coverage, and allocate collection effort toward underrepresented, uncertain, or behaviorally informative regions of the data distribution.

Finally, how to use addresses the integration of heterogeneous data sources during model training. This includes aligning action representations across embodiments, determining effective data recipes for combining sources from different levels of the data pyramid, and transferring egocentric human interaction data to dexterous robotic manipulation. Together, these directions shift the focus from merely increasing dataset size toward improving the informational coverage, acquisition efficiency, and effective utilization of embodied data.

\subsection{Tactile Data for Contact-Rich Robot Learning}
Despite the rapid expansion of large-scale robot learning datasets, most existing data resources remain centered on RGB-D observations, proprioceptive states, language instructions, and action trajectories, while tactile sensing is still not systematically incorporated as a standard modality~\cite{o2024open,khazatsky2024droid,dream-tac}. This leaves a missing contact layer in the current data pyramid: visual observations can describe scene geometry and object motion, but they provide only indirect evidence of contact force, slip, local deformation, friction, and grasp stability. Although recent robot datasets such as RoboMIND 2.0~\cite{hou2025robomind} and Humanoid Everyday ~\cite{zhao2025humanoid} have begun to incorporate tactile signals, tactile sensing has not yet become a common or standardized component of robot data collection. Compared with RGB-D observations and action trajectories, tactile data remains less prevalent and is often constrained by sensor-specific hardware, inconsistent signal formats, limited task coverage, and weaker integration with long-horizon manipulation trajectories.

Momentum in this direction has nonetheless accelerated. Vision-based tactile sensors, which convert contact-induced deformation of a soft interface into images, have been surveyed as an integrated sensing-and-learning system spanning hardware, learning methods, simulation, and datasets~\cite{zhou2026visionbased}, and this image-based formulation makes tactile observations directly compatible with the visual encoders already used by embodied foundation models. Building on it, tactile-augmented policies have moved from per-task adaptation toward pretraining at scale: $N_0$-VTLA~\cite{neoteai2026n0vtla} pretrains a vision-tactile-language-action model on a large-scale visuo-tactile robot corpus with staged tactile-pathway integration, $\tau$~\cite{cheng2026tau} learns an action-conditioned spatiotemporal tactile representation from future visual supervision so that high-dimensional tactile signals can be exploited under limited task-specific data, and ReTouch~\cite{zhang2026retouch} predicts future tactile states jointly with actions and refines those predictions online from execution-time feedback. Dataset-side efforts are converging on the same standardization need identified above: SoftVTBench~\cite{jing2026softvtbench} synchronizes multi-view RGB, dual-finger tactile images and marker motion, proprioception, language, and actions with evaluator-only physical ground truth, PRISM~\cite{yu2026prism} records tactile alongside force/torque and multi-view RGB-D for high-precision industrial operations, and EgoTac~\cite{zhang2026egotac} suggests a complementary route in which tactile channels are predicted from egocentric video and thereby retrofitted onto existing uninstrumented human corpora. What remains missing is agreement on a common tactile observation space: these resources still differ in sensor type, signal format, spatial resolution, and calibration, so cross-dataset tactile aggregation is not yet possible in the way it is for RGB and action trajectories.


\subsection{Failure Data and Recovery-Centric Robot Learning}

Most existing robot learning datasets remain strongly biased toward successful expert demonstrations, while failed, near-failure, and suboptimal trajectories are frequently discarded or only coarsely labeled. This success-centric data bias limits the exposure of learned policies to the diverse error states encountered during real-world deployment. A policy trained primarily on successful trajectories may learn how to complete a task under nominal conditions, but receive little supervision on how to recognize an impending failure, diagnose its cause, or recover after execution deviates from the expected trajectory. Failure data should therefore not be regarded merely as low-quality experience, but as an important source of supervision for failure awareness, causal diagnosis, and recovery behavior.

Existing efforts have begun to explore failure-centric data from several perspectives. Some works construct failure-oriented datasets or benchmarks for failure detection, diagnosis, explanation, and correction reasoning~\cite{liu2023reflect,duan2024aha,lu2025robofac,zeng2025vifailback,pacaud2025guardian,grislain2025ifailsense}. Other studies explicitly augment training data with failure and recovery trajectories, enabling policies to learn corrective behaviors beyond nominal expert demonstrations~\cite{dai2025racer,lin2025failsafe}. Meanwhile, imperfect and unsuccessful demonstrations have also been shown to contain reusable trajectory segments that can benefit policy learning when appropriately identified and filtered rather than simply discarded~\cite{wu2025imperfect}.

Nevertheless, failure data remains limited in scale and lacks systematic organization. Failures in robotic manipulation are highly heterogeneous and may arise from perception errors, inaccurate localization, inappropriate task planning, failed grasping, object slippage, excessive contact force, kinematic constraints, or accumulated errors over long-horizon execution. A binary success or failure label is insufficient to represent such temporal and causal structures. Future datasets should provide richer failure-centric annotations, including pre-failure context, failure onset, failure categories and causes, object and robot state changes, recovery actions, and final recovery outcomes. More importantly, future data collection pipelines should systematically preserve and curate failures encountered during robot execution rather than treating them as unusable trajectories. Building large-scale, diverse, and structured failure-recovery datasets may enable embodied agents not only to imitate successful behaviors, but also to understand, anticipate, and recover from their own mistakes.

\subsection{Scalable Data Collection Across Pyramid Layers}


Egocentric data collection is expanding from first-person RGB video to diverse wearable sensors. Head-mounted AR/MR devices can record gaze, head pose, depth, and spatial context, while wrist cameras, hand-mounted IMUs, and instrumented gloves capture hand motion and hand-object interaction. Tactile patches, force sensors, and EMG further provide contact, grasp force, slip, and muscle activation signals that vision cannot directly observe~\cite{ma2024nymeria,kareer2025egomimic,hoque2026egodex,dexglovehoi2026,egoevhands2026,egoemg2026,egotouch2026}. Wearable interfaces such as DexUMI and RealDexUMI also show the potential to collect dexterous demonstrations that are more compatible with robot learning~\cite{xu2025dexumi,realdexumi}.

However, many of these devices are not yet mature for large-scale deployment. Head and hand tracking remain sensitive to occlusion, motion blur, drift, and limited field of view, while gloves, tactile sensors, and EMG require careful fitting, calibration, and synchronization and may interfere with natural operation. Future systems should therefore become lighter, wireless, modular, and less intrusive, while supporting automatic cross-device calibration, on-device hand-object reconstruction, and standardized sensor metadata. Early signs of this consolidation are visible on both the human and robot sides: ACE-Data-0~\cite{cao2026acedata0} calibrates real homes as multi-scale capture spaces so that hand-object manipulation and whole-body activity are recorded in a common spatial and temporal frame, while ModPack~\cite{citron2026modpack} packages computation, power, communication, and storage into a self-contained wearable unit with plug-and-play capability modules that can be reused across robot platforms. Progress also depends on making existing wearables trustworthy rather than only more numerous: soft tactile gloves respond to hand pose as well as contact force, and correcting the resulting pose-related artifacts raises the reliability of low-force contact measurements without changing the hardware~\cite{yu2026poseaware}. These developments could extend the scale of emerging task-oriented ego datasets~\cite{jawaid2025openego,zheng2026egoscale,li2026egolive} without sacrificing the detailed motion and contact signals needed for robot learning.

\subsection{Cross-Embodiment State-Action Alignment}
As real-robot datasets increasingly aggregate trajectories from heterogeneous platforms, a common data format alone does not ensure compatible supervision. To facilitate cross-embodiment aggregation, robot states and actions are often represented in Cartesian end-effector pose space rather than morphology-specific joint space, providing a more transferable interface across platforms. Yet end-effector poses may still be defined relative to the robot base, camera, local end-effector, or world coordinate frame, causing the same physical motion to correspond to different numerical representations across datasets~\cite{o2024open}. Such inconsistencies can introduce conflicting supervision during joint training and substantially affect cross-embodiment transfer and data-scaling behavior~\cite{wang2026rethinking,yuan2026qwen}. Effective state-action alignment therefore requires consistent geometric semantics, rather than merely a shared storage schema.
Future datasets should record coordinate conventions, calibration parameters, and controller modes as first-class metadata. When reliable transformations are available, actions can be canonicalized into a shared reference frame, such as the camera, end-effector, or world frame, to support consistent aggregation across embodiments~\cite{xie2025unify,yuan2026qwen}.

\subsection{Egocentric Priors for Dexterous Hand Policy Learning}

While egocentric human data provides an intuitive source of priors for dexterous manipulation~\cite{yang2025egovla,luo2026beingh05scalinghumancentricrobot,fu2025metis,cai2025n,zheng2026egoscale}, how to optimally leverage such priors for robot policy learning remains an open question. The human-to-robot gap is not merely visual, but also kinematic, morphological, and physical: human and robotic hands differ in joint topology, degrees of freedom, fingertip geometry, compliance, actuation, sensing, friction, and force limits. Consequently, retargeted motions that appear geometrically plausible may still violate robot-specific constraints or fail to produce stable contacts on real hardware. Embodiment-aware inpainting~\cite{xu2025dexumi,li2025h2r,yang2026handedit} can reduce the visual gap by replacing human hands with robot renderings, but the synthesized interactions are often not physically grounded, potentially impairing the model's ability to learn realistic hand-object interaction dynamics, which are crucial for dexterous manipulation~\cite{yuan2025motiontrans,realdexumi}.

A promising direction is to treat egocentric data as a source of structured interaction priors rather than precise robot action labels. Human videos can provide transferable information about task intent, object affordances, grasp choices, contact sequences, tool-use strategies, and long-horizon manipulation structure, while robot data can ground these priors in embodiment-specific actions and contact dynamics. Morphology-conditioned policies, contact-centric representations, and uncertainty-aware retargeting may further help separate broadly transferable interaction knowledge from embodiment-dependent execution details.

FAKTUAL~\cite{sirigiri2026faktual} makes diversity an explicit curation criterion through trajectory signature-kernel entropy, selecting demonstration subsets without training an auxiliary policy or collecting online rollouts. Its simulation and real-robot evaluations show that diverse subsets can retain useful coverage at a smaller data budget. Diversity is not a substitute for quality control: the evaluated datasets contain successful demonstrations, and atypical or harmful trajectories may be selected when entropy alone is optimized.

\subsection{Data Recipes}
As illustrated in Figure~\ref{fig:data-application-evolution}, the data recipes of embodied foundation models have gradually expanded across different layers of the data pyramid. Earlier models mainly combined real-robot data with general-purpose vision-language data, whereas more recent reports increasingly construct mixtures involving real-robot, simulation, egocentric, UMI-style, and general multimodal data. Rather than converging on a single composition, different models select and combine different subsets of these data sources according to their training pipelines. In particular, egocentric data has become an increasingly common component of large-scale pretraining, while general data continues to be widely used together with embodiment-specific and action-aligned data. This trend indicates a shift from relying on a dominant data source toward jointly exploiting multiple levels of the data pyramid.

However, the most effective data composition remains an open problem. The optimal proportions among general, egocentric, simulation, UMI-style, and real-robot data have not been established. Existing reports differ substantially in their source combinations, sampling ratios, and stage-wise allocation strategies. Precise, compute-matched ablations of individual datasets and data categories remain limited. The performance gains produced by the same data source across different model architectures, action representations, and training objectives are also insufficiently characterized. Future work should therefore investigate architecture-aware and stage-dependent data recipes, with systematic comparisons of fixed, curriculum-based, and adaptive mixture strategies. Such studies may help identify when each layer of the data pyramid is most useful and provide more reproducible principles for allocating data under constrained collection and training budgets.

Two complementary results indicate what such studies might find. On the demand side, the number of demonstrations required is not a smooth function of task difficulty: for high-precision closed-world tasks such as assembly, the demonstrations $N$ needed to reach a fixed success rate grow super-exponentially as the target precision $P$ approaches a limit $c$, following $\log(N) \propto 1/(P-c)$, and the limit $c$ is not a physical constant of the task but an emergent property of the whole agent system, including its sensors and expert policy~\cite{xu2026curse}. Improving a system component can therefore be a substitute for collecting more data, and any recipe stated purely as a mixture ratio is incomplete without the precision regime and hardware stack it was measured on. On the supply side, within a single pyramid layer the useful fraction of a pool can be small: mining roughly 5\% of a 32M-sample egocentric pool for task relevance outperforms an equal-size random draw by a wide margin~\cite{lin2026simdex}. Taken together, these suggest that data recipes should be specified over \emph{selected} subsets and reported against a precision target, rather than as fixed proportions of whole corpora.

\section{Conclusion}
This work provides a data-centric synthesis of embodied intelligence and defines an embodied data pyramid organized primarily by scalability and robot alignment, with quality, diversity, reusability, and physical fidelity serving as complementary dimensions. From the apex to the base, the pyramid contains five data categories with distinct strengths and limitations. Real-robot data provides the most direct supervision for executable behavior and physical interaction, but remains costly and difficult to scale.  UMI-style data supports portable real-world collection with explicit end-effector and gripper supervision, but requires calibration, retargeting, and embodiment-specific control. Egocentric and exocentric data captures diverse human activities, hand-object interactions, and long-horizon task structures, while being limited by the human-robot embodiment gap. Simulation data enables controllable and parallel generation of robot-oriented experience with privileged supervision, although its utility is constrained by environment coverage and the sim-to-real gap. General multimodal data offers broad semantic, perceptual, and reasoning priors at large scale, but provides limited grounding in robot actions, contacts, and physical outcomes.

Building on this taxonomy, we analyze recent embodied foundation models from the perspective of data recipes. We summarize action representations and their implications for data compatibility and cross-embodiment transfer, and examine how embodied brain models, vision-language-action models, and world-action models utilize heterogeneous data to support reasoning and planning, executable control, and predictive world modeling, respectively. Finally, we identify open challenges in data quality assessment, multimodal and action-space standardization, cross-embodiment reuse, human-to-robot and sim-to-real transfer, scalable real-world collection, failure and recovery data, and the design and evaluation of heterogeneous data mixtures. We hope this paper provides both a coherent foundation for understanding the embodied data landscape and a forward-looking roadmap for selecting, collecting, combining, and evaluating data for different capabilities, embodiments, and deployment settings.

\bibliographystyle{plainnat}
\bibliography{main}

\end{document}